%% file: main.tex
\documentclass[10pt,twocolumn,letterpaper]{article}

\usepackage[pagenumbers]{cvpr} 

\usepackage{times}
\usepackage{epsfig}
\usepackage{graphicx}
\usepackage{amsmath}
\usepackage{amssymb}
\usepackage{bbm}
\usepackage{booktabs}
\usepackage{multirow}
\usepackage{float}
\usepackage{amsfonts}
\usepackage{comment}
\usepackage{url}
\usepackage{relsize}
\usepackage{dsfont}
\usepackage{arydshln}
\usepackage{graphicx,wrapfig,lipsum}
\usepackage{array}
\usepackage{mynotation}
\usepackage[dvipsnames]{xcolor}
\usepackage{dblfloatfix}  
\usepackage[pagebackref,breaklinks,colorlinks]{hyperref}

\newcommand{\bp}[1]{\textbf{#1}}
\newcommand{\parsection}[1]{\vspace{1mm}\noindent\textbf{#1:}}
\def\etal{\emph{et al}\onedot}
\def\eg{\emph{e.g}\onedot} 
\def\ie{\emph{i.e}\onedot}

\definecolor{dark-gray}{gray}{0.45}

\makeatletter
\def\adl@drawiv#1#2#3{%
        \hskip.5\tabcolsep
        \xleaders#3{#2.5\@tempdimb #1{1}#2.5\@tempdimb}%
                #2\z@ plus1fil minus1fil\relax
        \hskip.5\tabcolsep}
\newcommand{\cdashlinelr}[1]{%
  \noalign{\vskip\aboverulesep
           \global\let\@dashdrawstore\adl@draw
           \global\let\adl@draw\adl@drawiv}
  \cdashline{#1}
  \noalign{\global\let\adl@draw\@dashdrawstore
           \vskip\belowrulesep}}
\makeatother

\usepackage[capitalize]{cleveref}
\crefname{section}{Sec.}{Secs.}
\Crefname{section}{Section}{Sections}
\Crefname{table}{Table}{Tables}
\crefname{table}{Tab.}{Tabs.}

\def\cvprPaperID{1610} 
\def\confName{CVPR}
\def\confYear{2022}

\begin{document}

\title{Probabilistic Warp Consistency for Weakly-Supervised Semantic Correspondences}

\author{Prune Truong \qquad Martin Danelljan \qquad Fisher Yu  \qquad Luc Van Gool \\
Computer Vision Lab, ETH Zurich, Switzerland\\
\small{\texttt{\{prune.truong, martin.danelljan, vangool\}@vision.ee.ethz.ch} \qquad \texttt{i@yf.io}} \\
}

\maketitle

\begin{abstract}

\input{abstract}
\end{abstract}

\input{01_introduction}
\input{02_related_work}
\input{03_method}

\input{04_experimental_work}

\input{05_conclusion}

{\small
\bibliographystyle{ieee_fullname}
\bibliography{biblio}
}

\clearpage
\newpage
\appendix
\begin{center}
	\textbf{\Large Appendix}
\end{center}
\input{supplementary}
\end{document}

%% file: abstract.tex
We propose Probabilistic Warp Consistency, a weakly-supervised learning objective for semantic matching. Our approach directly supervises the dense matching scores predicted by the network, encoded as a conditional probability distribution. We first construct an image triplet by applying a known warp to one of the images in a pair depicting different instances of the same object class. Our probabilistic learning objectives are then derived using the constraints arising from the resulting image triplet. We further account for occlusion and background clutter present in real image pairs by extending our probabilistic output space with a learnable unmatched state. To supervise it, we design an objective between image pairs depicting different object classes. We validate our method by applying it to four recent semantic matching architectures. Our weakly-supervised approach sets a new state-of-the-art on four challenging semantic matching benchmarks. Lastly, we demonstrate that our objective also brings substantial improvements in the strongly-supervised regime, when combined with keypoint annotations. 

%% file: 01_introduction.tex
\section{Introduction}

The semantic matching problem entails finding pixel-wise correspondences between images depicting instances of the same semantic category of object or scene, such as `cat' or `bird'. It has received growing interest, due to its applications in \eg, semantic segmentation~\cite{Taniai2016, RubinsteinJKL13} and image editing~\cite{DaleJSMP09, BarnesSFG09, HaCohenSGL11, LeeKLKCC20}. 
The task nevertheless remains extremely challenging due to the large intra-class appearance and shape variations, view-point changes, and background-clutter.
These issues are further complicated by the inherent difficulty to obtain ground-truth annotations.  

\begin{figure}[t]
\centering%
\vspace{-3mm}
\includegraphics[width=0.47\textwidth, trim=0 0 0 0]{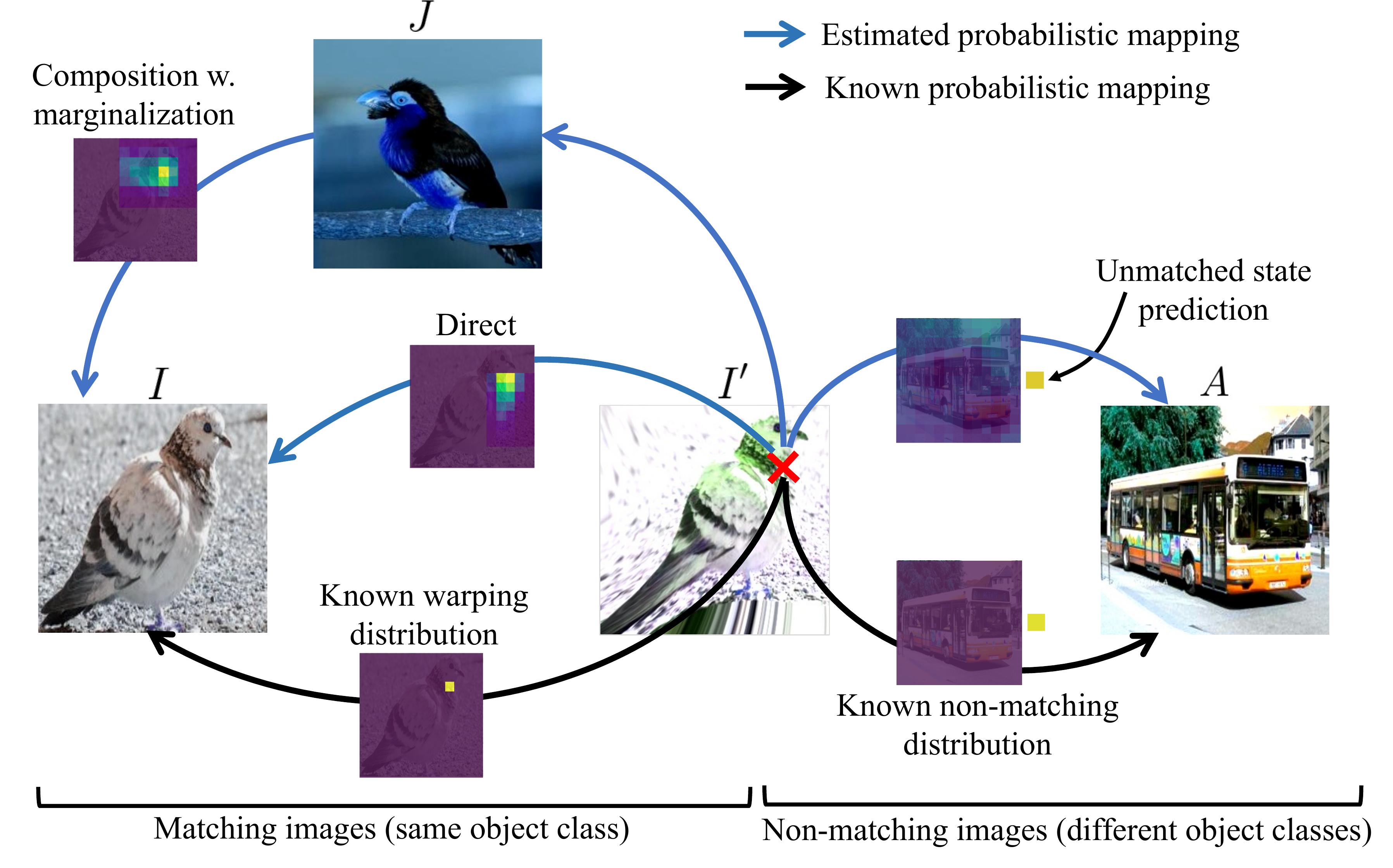}
\vspace{-3mm}
\caption{From a real image pair $(I, J)$ representing the same object class, we generate a new image $I'$ by warping $I$ according to a randomly sampled transformation. We further extend the image triplet with an additional image $A$, that depicts a different object class. 
For each pixel in $I'$, we introduce two consistency objectives by enforcing the conditional probability distributions obtained either from the composition $I' \rightarrow J \rightarrow I$, or directly through $I' \rightarrow I$, to be equal to the known warping distribution. 
We further model occlusion and unmatched regions by introducing a learnable unmatched state. It is trained by enforcing the predicted distribution between the non-matching images $(I', A)$ to be mapped to the unmatched state for all pixels. 
}
\vspace{-4mm}
\label{fig:intro}
\end{figure}

While a few current datasets~\cite{PFPascal, PFWillow, spair} provide manually annotated keypoints matches, these are often ill-defined, ambiguous and scarce. Strongly-supervised approaches relying on such annotations therefore struggle to generalize across datasets, as demonstrated in recent works~\cite{MinLPC20, CATs}. 
As a prominent alternative, unsupervised approaches~\cite{GLUNet, GOCor, Rocco2017a, Melekhov2019, Rocco2018a, SeoLJHC18, MinLPC20} often train the network with synthetically generated dense ground-truth and image data. While benefiting from direct supervision, the lack of real image pairs often leads to poor generalization to real data. 
\emph{Weakly-supervised} methods~\cite{MinLPC20, Rocco2018a, Rocco2018b, Jeon, DCCNet} thus appear as an attractive paradigm, leveraging supervision from real image pairs by only exploiting image-level class labels, which are inexpensive compared to keypoint annotations. 

Previous weakly-supervised alternatives introduce objectives on the predicted dense correspondence volume, which encapsulates the matching confidences for all pairwise matches between the image pair. The most common strategy is to maximize the maximum scores~\cite{Rocco2018b, DCCNet} or negative entropy~\cite{MinLPC20} of the correspondence volume computed between images of the same class, while minimizing the same quantity for images of different classes. However, these strategies only provide very limited supervision due to their weak and indirect learning signal. 
While these approaches act directly on the predicted dense correspondence volume, Truong~\etal~\cite{warpc} recently introduced Warp Consistency, a weakly-supervised learning objective for dense flow regression. The objective is derived from flow constraints obtained when introducing a third image, constructed by randomly warping one of the images in the original pair.
While it achieves impressive results, the warp consistency objective is limited to the learning of flow regression. As such an approach predicts a single match for each pixel without any confidence measure, it struggles to handle occlusions and background clutter, which are prominent in the semantic matching task.

We propose Probabilistic Warp Consistency, a weakly-supervised learning objective for semantic matching. Following~\cite{Rocco2018b, DCCNet, CATs}, we first employ a \emph{probabilistic mapping} representation of the predicted dense correspondences, encoding the transitional probabilities from every pixel in one image to every pixel in the other.  
Starting from a real image pair $( I, J)$, we consider the image triplet introduced in~\cite{warpc}, where the synthetic image $I'$ is related to $I$ by a randomly sampled warp (Fig.~\ref{fig:intro}). We derive our probabilistic consistency objective based on predicting the \emph{known} probabilistic mapping relating $I'$ to $I$ with the composition through the image $J$. The composition is obtained by marginalizing over all the intermediate paths that link pixels in image $I'$ to pixels in $I$ through image $J$.

Since the constraints employed to derive our objective are only valid in mutually visible object regions, we further tackle the problem of identifying pixels that can be matched. This is particularly challenging in the presence of background clutter and occlusions, common in semantic matching. 
We explicitly model occlusion and unmatched regions, by introducing a learnable unmatched state into our probabilistic mapping formulation. 
To train the model to detect unmatched regions, we design an additional probabilistic loss that is applied on pairs of images depicting different object classes, as illustrated in Fig.~\ref{fig:intro}. Further, we also employ a visibility mask, which constrains our introduced consistency loss to visible object regions.

We extensively evaluate and analyze our approach by applying it to four recent semantic matching architectures, across four benchmark datasets.
In particular, we train SF-Net~\cite{SFNet} and NC-Net~\cite{Rocco2018b} with our weakly-supervised Probabilistic Warp Consistency objective. Our approach brings relative gains of $4.3\%$ and  $5.8\%$ on PF-Pascal~\cite{PFPascal} and PF-Willow~\cite{PFWillow} respectively, for SF-Net, and $+22.6\%$ and $+14.8\%$ for NC-Net on SPair-71K~\cite{spair} and TSS~\cite{Taniai2016}, respectively. 
This leads to a new state-of-the-art on all four datasets.
Finally, we extend our approach to the strongly-supervised regime, by combining our probabilistic objectives with keypoint supervision. When integrated in SF-Net, NC-Net, DHPF~\cite{MinLPC20} and CATs~\cite{CATs}, it leads to substantially better generalization properties across datasets, setting a new state-of-the-art on three benchmarks. Code is available at \url{github.com/PruneTruong/DenseMatching}

%% file: 02_related_work.tex
\section{Related Work}

\parsection{Semantic matching architectures} 
Most semantic matching pipelines include 3 main steps, namely feature extraction, cost volume construction, and displacement estimation. 
Multiple works focus on the latter, through either predicting the global geometric transformation parameters~\cite{Rocco2017a, ArbiconNet, SeoLJHC18, Kim2018, Rocco2018a, Jeon}, or directly regressing the flow field~\cite{GLUNet, GOCor, pdcnet, warpc, Kim2019} relating an image pair. Nevertheless, most methods instead predict a cost volume as the final network output, which is further transposed to point-to-point correspondences with argmax or soft-argmax~\cite{SFNet} operations. 
Recent methods thus focus on improving the cost volume aggregation stage, through formulating the semantic matching task as an optimal transport problem~\cite{SCOT} or leveraging multi-resolution features and cost volumes~\cite{HPF, MinLPC20, SFNet, MMNet, CATs}.
Another line of work deals with refining the cost volume, with 4D~\cite{Rocco2018b, ANCNet, DCCNet, PMNC} or 6D~\cite{CHM} convolutions, an online optimization-based module~\cite{GOCor}, an encoder-decoder style architecture~\cite{GSF} or a Transformer module~\cite{CATs}.

\parsection{Unsupervised and weakly-supervised semantic matching}
A common technique for unsupervised learning of semantic correspondences is to rely on synthetically warped versions of images~\cite{Rocco2017a, ArbiconNet, SeoLJHC18, GLUNet, GSF}. 
It nevertheless comes at the cost of poorer generalization abilities to real data. 
Some methods instead use real image pairs, by leveraging additional annotations in the form of 3D CAD models~\cite{Zhou2016, abs-2004-09061}, segmentation masks~\cite{SFNet, ChenL0H21}, or by jointly learning semantic matching with attribute transfer~\cite{Kim2019}. 
Most related to our work are approaches that use proxy losses on the cost volume constructed between real image pairs, with image labels as the only supervision~\cite{Jeon, Rocco2018a, Rocco2018b, DCCNet, Kim2018}. Jeon~\etal~\cite{Jeon} identify correct matches from forward-backward consistency. NC-Net~\cite{Rocco2018b} and DCC-Net~\cite{DCCNet} are trained by maximizing the mean matching scores over all hard assigned matches from the cost volume. Min~\etal~\cite{MinLPC20} instead encourage low and high correlation entropy for image pairs depicting the same or different classes, respectively.  
In this work, we instead construct an image triplet by warping one of the original images with a known warp, from which we derive our probabilistic losses.

\parsection{Unsupervised learning from videos} Our approach is also related to~\cite{randomwalk}, which proposes a self-supervised approach for learning features, by casting matches as predictions of links in a space-time graph constructed from videos. 
Recent works~\cite{DwibediATSZ19,JabriOE20, WangJE19} further leverage the temporal consistency in videos to learn a representation for feature matching. 

%% file: 03_method.tex
\section{Background: Warp Consistency}

We derive our approach based on the warp consistency constraints introduced by~\cite{warpc}. 
They propose a weakly-supervised loss, termed Warp Consistency, for learning correspondence regression networks. We therefore first review relevant background and introduce the notation that we use.

We define the mapping $M_{I \leftarrow  J} : \reals^2 \rightarrow \reals^2$, which encodes the absolute location $M_{I \leftarrow  J}(\mathbf{j}) \in \reals^2$ in $I$ corresponding to the pixel location $\mathbf{j} \in \reals^2$ in image $J$. We consistently use the hat $\, \widehat{\cdot} \, $ to denote an estimated or predicted quantity.

\begin{figure*}[t]
\centering%
\subfloat[Warp Consistency Graph~\cite{warpc} \label{fig:warpc}]{\includegraphics[width=0.30\textwidth]{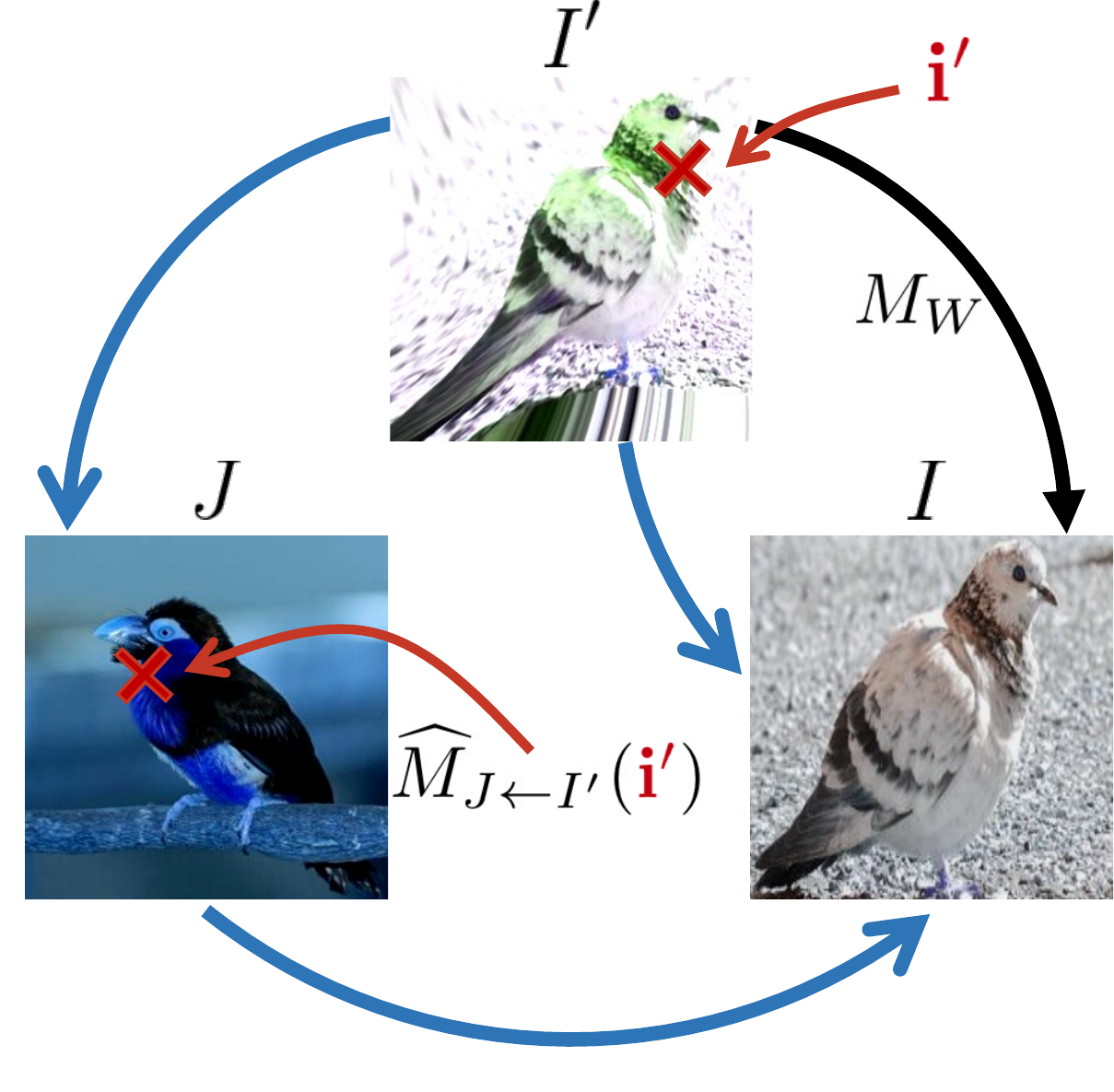}}~ \hspace{2mm}
\subfloat[Our probabilistic PW-bipath~\eqref{eq:compo-w-bipath-proba-matrix} and PWarp-supervision~constraints, with corresponding losses \eqref{eq:ce}-\eqref{eq:ce-warpsup}  \label{fig:proba-warpc}]{\includegraphics[width=0.67\textwidth, trim=0 0 0 5, clip]{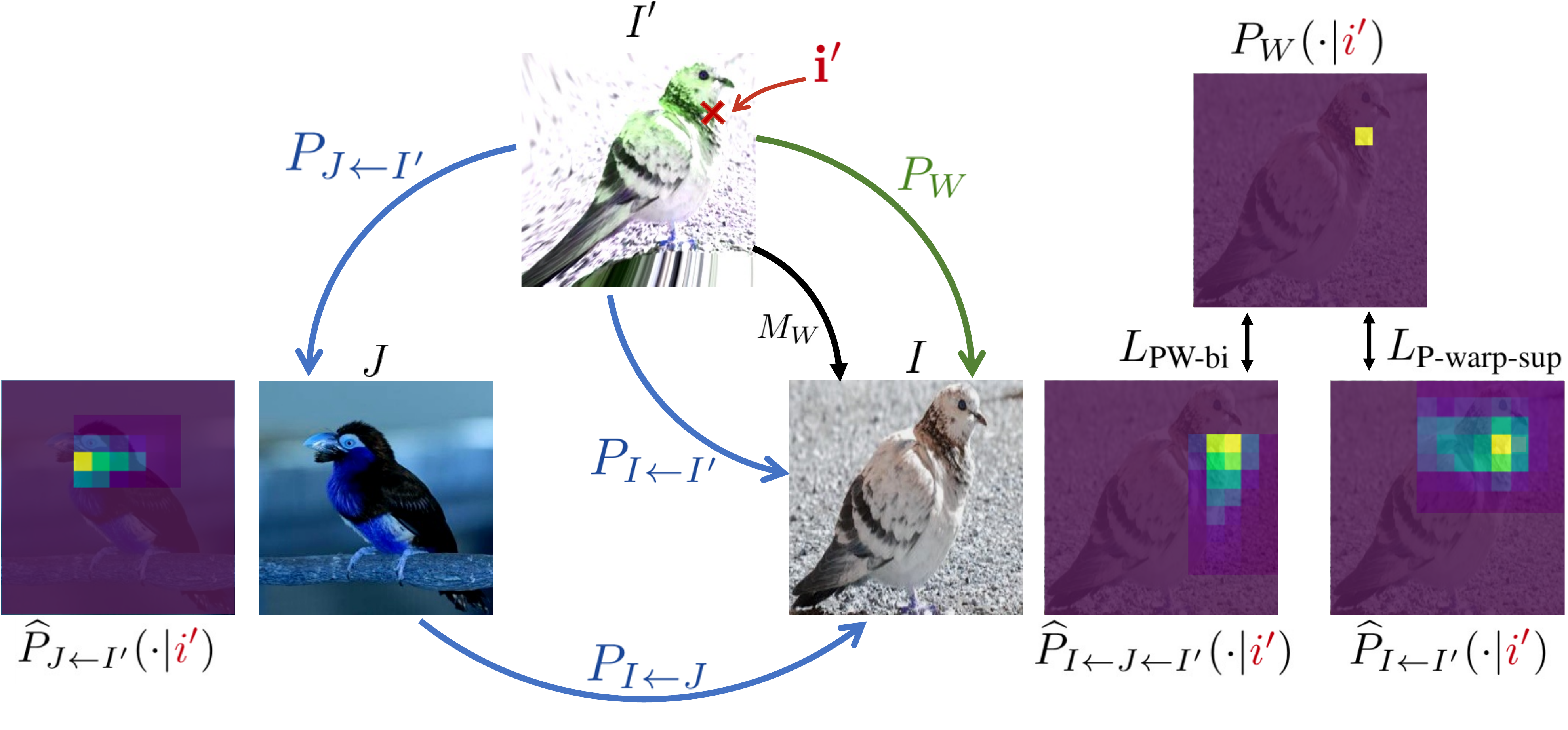}}~%
\vspace{-2mm}
\caption{Mapping and probabilistic mapping constraints derived from the warp consistency graph between the images $( I, I', J)$. $I'$ is generated by warping $I$ according to a randomly sampled mapping $M_W$ (black arrow). (a) The W-bipath~\eqref{eq:bipath-mapping} and warp-supervision~\eqref{eq:warp-sup-mapping} mapping constraints~\cite{warpc} predict $M_W$ by the composition $I' \rightarrow J \rightarrow I$, and directly by $I' \rightarrow I$ respectively. (b) Our probabilistic PW-bipath and PWarp-supervision constraints are derived by enforcing the composition $\widehat{P}_{I \leftarrow J \leftarrow I'}$ of the predicted distributions, and the direct prediction $\widehat{P}_{I \leftarrow I'}$ respectively, to be equal to the known warping distribution $P_W$. 
}\vspace{-4mm}
\label{fig:cycles-method}
\end{figure*}

\parsection{Warp Consistency graph} Truong~\etal~\cite{warpc} first build an image triplet, which is used to derive the constraints. 
From a real image pair $(I, J)$, an image triplet $(I, I', J)$ is constructed, by creating $I'$ through warping of $I$ with a randomly sampled mapping $M_W$, as $I' = I \circ M_W$. Here, $\circ$ denotes function composition. The resulting triplet $(I, I', J)$ gives rise to a warp consistency graph (Fig.~\ref{fig:cycles-method}a), from which a family of mapping-consistency constraints is derived.

\parsection{Mapping-consistency constraints} Truong~\etal~\cite{warpc} analyse the possible mapping-consistency constraints arising from the triplet and identify two of them as most suitable when designing a weakly-supervised learning objective for dense correspondence regression. Particularly, the proposed objective is based on the W-bipath constraint, where the mapping $M_W$ is computed through the composition $I' \rightarrow J \rightarrow I$ via image $J$, formulated as,
\begin{equation}
\label{eq:bipath-mapping}
M_{W} = M_{I\leftarrow J} \circ M_{J\leftarrow I'}  \,.
\end{equation} 
It is further combined with the warp-supervision constraint, 
\begin{equation}
\label{eq:warp-sup-mapping}
M_{W} = M_{I\leftarrow I'} \,,
\end{equation}
derived from the graph by the direct path $I' \rightarrow I$.

In~\cite{warpc}, these constraints were used to derive a weakly-supervised objective for correspondence regression. However, regressing a mapping vector $M_{I \leftarrow J}(\mathbf{j})$ for each position $\mathbf{j}$ only retrieves the position of the match, without any information on its uncertainty or multiple hypotheses. We instead aim at predicting a matching conditional probability distribution for each position $\mathbf{j}$. The distribution encapsulates richer information about the matching ability of this location $\mathbf{j}$, such as confidence, uniqueness, and existence of the correspondence. In this work, we thus generalize the mapping constraints~\eqref{eq:bipath-mapping}-~\eqref{eq:warp-sup-mapping} extracted from the warp consistency graph to conditional probability distributions.

\section{Method}

We address the problem of estimating the pixel-wise correspondences relating an image pair $(J, I)$, depicting semantically similar objects. The dense matches are encapsulated in the form of a conditional probability matrix, referred to as probabilistic mapping. The goal of this work is to design a weakly-supervised learning objective for probabilistic mappings, applied to the semantic matching task.

\subsection{Probabilistic Formulation}

In this section, we first introduce our probabilistic representation and define a typical base predictive architecture. 
We let $\mathbf{j} \in \mathbb{R}^2$ denote the 2D pixel location in a grid of dimension $h_J \times w_J$, corresponding to image $J$. We refer to $j \in \mathbb{R}$ as the index $j=\left\{1, ..., h_Jw_J \right\}$ corresponding to $\mathbf{j}$ when the spatial dimensions  $h_J \times w_J$ are vectorized into one dimension $h_Jw_J$. 
Following~\cite{Rocco2018b, MinLPC20, CATs}, we aim at predicting the probabilistic mapping $P_{I \leftarrow J} \in \mathbb{R}^{h_Iw_I \times h_J w_J}$ relating $J$ to $I$. Given a position $j$ in frame $J$, $P_{I \leftarrow J}(i | j)$ gives the probability that $j$ is mapped to location $i$ in image $I$. $P_{I \leftarrow J}(\cdot | j) \in \mathbb{R}^{h_Iw_I}$  thus encodes the entire discrete conditional probability distribution of where $j$ is mapped in image $I$. 
We can see $P_{I \leftarrow J}$ as a matrix, where each column at index $j$ encapsulates the distribution $P_{I \leftarrow J}(\cdot | j)$. 
Also note that the probabilistic mapping $P_{I \leftarrow J}$ is asymmetric. 

\parsection{Probabilistic mapping prediction} We here describe a standard architecture predicting the probabilistic mapping $P$ relating an image pair. 
We let $D^I \in \mathbb{R}^{h_Iw_I \times d}$ and $D^J \in \mathbb{R}^{h_Jw_J \times d}$ denote the $d$-channel feature maps extracted from the images $I$ and $J$, respectively.

A cost volume $C_{I \leftarrow J} \in \mathbb{R}^{h_Iw_I \times h_Jw_J}$ is then constructed, which encodes the pairwise deep feature similarities between all locations in the two feature maps, as,
\begin{equation}
\label{eq:global-corr}
C_{I \leftarrow J}(i, j)=  D^I(i) ^{T} D^J(j)   \,.
\end{equation}

The cost volume is finally converted to a probabilistic mapping $P_{I \leftarrow J} \in \mathbb{R}^{h_Iw_I \times h_Jw_J}$ by simply applying the SoftMax operation over the first dimension, 
\begin{equation}
\label{eq:softmax}
P_{I \leftarrow J}(i | j) = \frac{\text{exp}(C_{I \leftarrow J}(i , j))}{\sum_{k} \text{exp}( C_{I \leftarrow J}(k, j) )}
\end{equation}
Note that extensions of this basic approach can also be considered, by \eg adding post-processing convolutional layers~\cite{Rocco2018b, DCCNet} or a Transformer module~\cite{CATs}.  The goal of this work is to design a weakly-supervised learning objective to train a neural network $f_\theta$, with parameters $\theta$, that predicts the probabilistic mapping $\widehat{P}_{I \leftarrow J} = f_{\theta} ( J, I )$ relating $J$ to $I$. 

\subsection{Probabilistic Warp Consistency Constraints}

We set out to design a weakly-supervised loss for probabilistic mappings.
To this end, we consider the consistency graph introduced in~\cite{warpc} and generalize the mapping constraints~\eqref{eq:bipath-mapping}-~\eqref{eq:warp-sup-mapping} to their corresponding probabilistic form. 

\parsection{Probabilistic W-bipath constraint} We start from the W-bipath constraint~\eqref{eq:bipath-mapping} extracted from the Warp Consistency graph Fig.~\ref{fig:cycles-method}a and extend it to its probabilistic matrix counterpart, which we denote as PW-bipath. 
It states that we obtain the same conditional probability distribution by proceeding through the path $I' \rightarrow I$, which is determined by the randomly sampled warp $M_W$, or by taking the detour through image $J$. In the latter case, the resulting probability distribution is derived by marginalizing over the intermediate paths that link pixels in $I'$ to pixels in $I$ through $J$ as,
\begin{equation}
\label{eq:compo-w-bipath-proba}
P_{W}(i | i' \,) = \sum_{j}  P_{I\leftarrow J}(i|j) \cdot  P_{J\leftarrow I'}(j|i' \,)\,.
\end{equation}
The above equality is expressed in matrix form as,
\begin{equation}
\label{eq:compo-w-bipath-proba-matrix}
P_{W} = P_{I \leftarrow J} \otimes  P_{J \leftarrow I'}  \,.
\end{equation}
where $\otimes $ represents matrix multiplication. This constraint is schematically represented in Fig.~\ref{fig:cycles-method}b.  

\parsection{PW-bipath training objective} We aim at formulating an objective based on the PW-bipath constraint~\eqref{eq:compo-w-bipath-proba-matrix}. 
Crucially, in our setting, the mapping $M_{I \leftarrow I'} = M_W$ 
is known by construction, from which we can derive the ground-truth probabilistic mapping $P_{I \leftarrow I'} = P_W \in \mathbb{R}^{h_Iw_I \times h_{I'}w_{I'}}$. 
To measure the distance between the right and the left side of~\eqref{eq:compo-w-bipath-proba-matrix}, the KL divergence appears as a natural choice. Since $P_W$ is a constant, it simplifies to the familiar cross-entropy, 
\begin{equation}
\label{eq:ce}
    L_{\text{PW-bi}} = \sum_{i'}  \mathcal{H} \left ( [ \widehat{P}_{I \leftarrow J} \otimes \widehat{P}_{J \leftarrow I'}](\cdot | i') \,\, , \,\, P_W(\cdot | i') \right )
\end{equation}
Here, $\mathcal{H}$ is the cross-entropy loss. To simplify notations, we sometimes refer to the marginalization as $\widehat{P}_{I \leftarrow J \leftarrow I'} = \widehat{P}_{I \leftarrow J} \otimes \widehat{P}_{J \leftarrow I'}$. 
Supervising $\widehat{P}_{I \leftarrow J \leftarrow I'}$ with the label $P_W$ provides an implicit learning signal for the predicted intermediate distributions $\widehat{P}_{J\leftarrow I'}$ and $\widehat{P}_{I\leftarrow J}$.

\parsection{PWarp-supervision constraint and objective} 
Similarly, we generalize the warp-supervision constraint~\eqref{eq:warp-sup-mapping} to its probabilistic matrix form, as $P_{W} = P_{I \leftarrow I'}$. 
As previously, by exploiting the fact that $P_W$ is known, we derive the corresponding training objective, 
\begin{equation}
\label{eq:ce-warpsup}
    L_{\text{P-warp-sup}} = \sum_{i'}  \mathcal{H} \left ( \widehat{P}_{I \leftarrow I'} (\cdot | i') \,\, , \,\, P_W(\cdot | i') \right )
\end{equation}

The PW-bipath constraint~\eqref{eq:compo-w-bipath-proba-matrix} and its loss~\eqref{eq:ce} assume that all pixels of image $I'$ have a match in both $I$ and $J$. 
However, due to the occlusions introduced by the triplet creation and the non-matching backgrounds of the images in the semantic matching task, this assumption is partly invalidated.

\subsection{Modelling Unmatched Regions}
\label{sec:occ}

The semantic matching task aims to estimate correspondences between different image instances of the same object class. However, even in that case, the backgrounds of each image do not match. 
As a result, the \emph{common visible regions} only represent a fraction of the images (see the birds in Fig.~\ref{fig:cycles-method}). Nevertheless, the distribution $P_{I \leftarrow J}(\cdot|j)$ is unable to model the no-match case for pixel $j$. 

Moreover, the construction of our image triplet $(I, I', J)$ introduces occluded areas, for which the constraint~\eqref{eq:compo-w-bipath-proba-matrix} is undefined. In fact, it is only valid in \emph{non-occluded object regions}. However, in our setting, the locations of the objects in the real image pairs $(I, J)$ are unknown. In this section, we derive our visibility-aware learning objective. We additionally introduce explicit modelling of occlusion and unmatchable regions into our probabilistic formulation.

\begin{figure}[t]
\centering%
\includegraphics[width=0.47\textwidth, trim=0 0 0 4, clip]{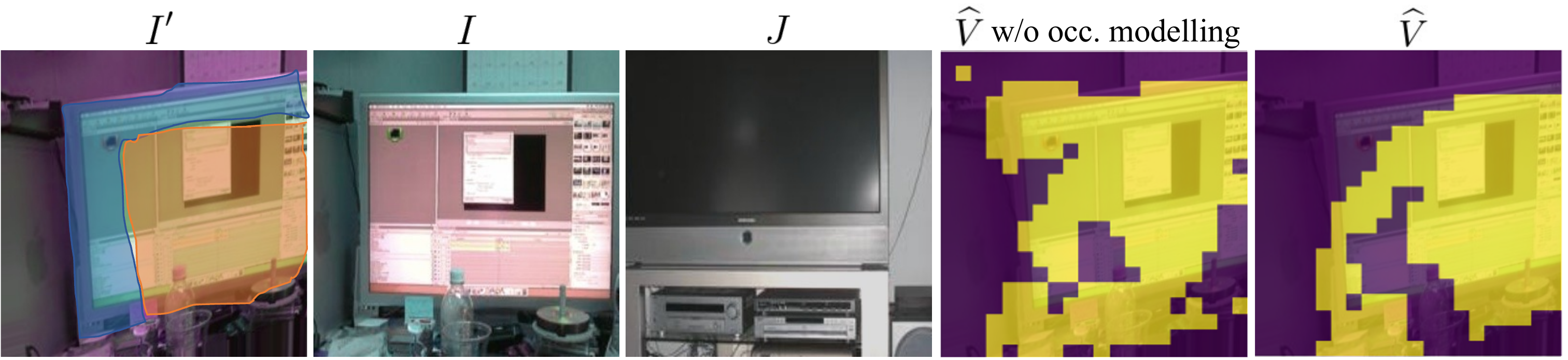}
\vspace{-3mm}\caption{Triplet of images for training, and the visibility mask $\widehat{V}$ (yellow is $\widehat{V}=1$). The shaded blue region on $I'$ represent object pixels visible in both $I'$ and $I$, but occluded in $J$, for which our PW-bipath loss~\eqref{eq:ce} is not valid. It is only valid in object regions visible in all three images, \ie the orange shaded region. Explicitly modelling occlusions further helps to identify them. 
}\vspace{-4mm}
\label{fig:valid-w-bipath}
\end{figure}

\parsection{Visibility-aware training objective} In general, the PW-bipath constraint~\eqref{eq:compo-w-bipath-proba-matrix} is only valid in regions of $I'$ that are visible in both images $J$ and $I$. That is, only in \emph{non-occluded object regions}, as illustrated in Fig.~\ref{fig:valid-w-bipath}. 
Applying the loss~\eqref{eq:ce} in non-matching regions, such as in background areas, or in occluded objects regions (blue area in Fig.~\ref{fig:valid-w-bipath}), bares the risk to confuse the network by enforcing matches in non-matching areas. 
As a result, we extend the introduced loss~\eqref{eq:ce} by further integrating a visibility mask $V \in [0,1]^{w_{I'} h_{I'}}$. The mask $V$ takes a value $V(i' \, ) = 1$ for any pixel $i'$ belonging to the non-occluded common object (roughly the orange area in Fig.~\ref{fig:valid-w-bipath}) and $V(i' \, ) = 0$ otherwise. The loss~\eqref{eq:ce} is then extended as, 
\begin{equation}
\label{eq:vis-w-bipath}
\centering
    L_{\text{vis-PW-bi}} = \sum_{i'} \widehat{V}(i' \, ) \,\,  \mathcal{H} \left ( \widehat{P}_{I \leftarrow J \leftarrow I'}(\cdot | i') \, , P_W(\cdot | i') \right )
\end{equation}

Since we do not know the true $V$, we aim to find an estimate $\widehat{V}$, also visualized in Fig.~\ref{fig:valid-w-bipath}. 
We consider the predicted probability value $\widehat{P}_{I \leftarrow J \leftarrow I'}(M_W(i') | i') \in \left[0, 1\right]$ of a pixel $i'$ of $I'$ to be mapped to position $M_W(i')$ in $I$, according to the known mapping $M_W$. 
We assume that this value should be higher in matching regions, \ie the object, than in non-matching regions, \ie the background, where the constraint~\eqref{eq:compo-w-bipath-proba-matrix} doesn't hold.
We therefore compute our visibility mask by taking the highest $\gamma$ percent of $\widehat{P}_{I  \leftarrow J \leftarrow I'}(M_W(i') | i')$ over all $i'$ of $I'$. The scalar $\gamma$ is a hyperparameter controlling the sensitivity of the mask estimation. While we do not know the actual coverage of the object in the image, which might vary across training images, we found that taking a high estimate for $\gamma$ is sufficient in practise, as it simply removes the obvious non-matching  regions.
Moreover, while we could have instead computed $\widehat{V}$ by thresholding the probabilities as
$\widehat{V}(i' \, ) = \mathbbm{1}\left[ \widehat{P}_{I  \leftarrow J \leftarrow I'}(M_W(i') | i') > \beta \right]$, 
our approach avoids tedious continuous tuning of the $\beta$ parameter during training, necessary to follow the evolution of the probabilities. 
While valid as it is, the accuracy of the estimate $\widehat{V}$ can further be improved through explicit occlusion modelling. 

\parsection{Occlusion modelling} In order to explicitly model occlusion and non-matching regions into our probabilistic mapping $P_{I \leftarrow J}$, we \emph{predict} the probability of a pixel to be occluded or unmatched in one image, given that it is visible in the other. This can, for example, be achieved by augmenting the cost volume $C$ in~\eqref{eq:global-corr} with an unmatched bin~\cite{superpoint, SarlinDMR20} $\o$, such as $C(\o , j) = z \in \mathbb{R}$, where $z$ is a single learnable parameter. After converting the cost volume $C$ into a probabilistic mapping $P$ through~\eqref{eq:softmax}, $P_{I \leftarrow J}(\o | j)$ encodes the probability of pixel $j$ of image $J$ to map to the unmatched or occluded state $\o$, \ie to have no match in image $I$. 
We further specify the matching distribution given an unmatched state, to always be mapped to the unmatched state.
Specifically, we augment $\widehat{P}$ with a fixed column, forcing the distribution given an unmatched state to be as $\widehat{P}(\o | \o) = 1$.

\parsection{Occlusion aware PW-bipath} Our modelling of the unmatched state given the unmatched state, as $\widehat{P}_{I \leftarrow J}(\o | \o) = 1$ naturally ensures that the following scheme is respected. If a pixel $i'$ in image $I'$ is predicted as unmatched in image $J$, such as $\widehat{P}_{J \leftarrow I'}(\o|i' \,) = 1$, it will also be predicted unmatched in image $I$, \ie $\widehat{P}_{ I \leftarrow J  \leftarrow I'}(\o|i' \,) = 1$. This prevents enforcing~\eqref{eq:vis-w-bipath} on $\widehat{P}_{ I \leftarrow J  \leftarrow I'}$ for pixels of image $I'$ which are visible in $I$, but occluded in image $J$ (blue area in Fig. \ref{fig:valid-w-bipath}). Moreover, predicting a high probability for the occluded state   $\widehat{P}_{ I \leftarrow J \leftarrow I'}( \o |i')$ allows to identify occluded and non-matching areas $i'$ in $I'$. 
It further ensures that these regions are not selected in $\widehat{V}$, and therefore not supervised with~\eqref{eq:vis-w-bipath}.

\begin{figure}[t]
\centering%
\includegraphics[width=0.40\textwidth, trim=0 0 0 20, clip]{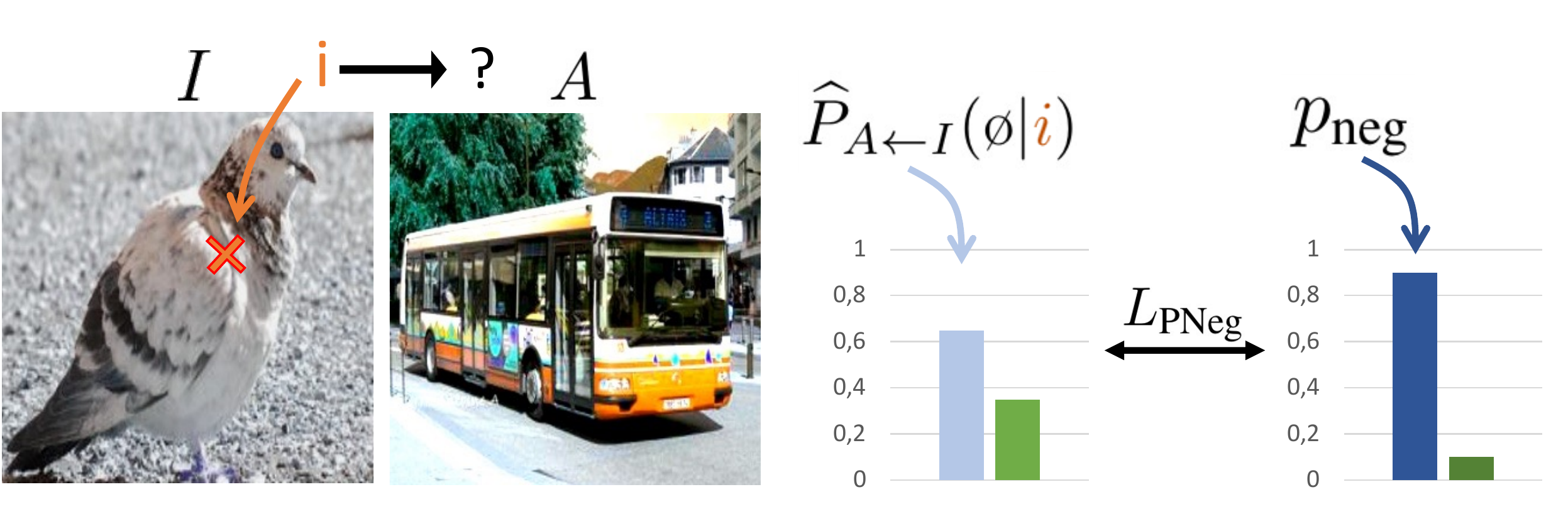}
\vspace{-4mm}
\caption{Learning objective on non-matching images $(I, A)$.}\vspace{-4mm}
\label{fig:neg-loss}
\end{figure}

\parsection{Supervision of the unmatched state} Our introduced objectives~\eqref{eq:ce-warpsup}-\eqref{eq:vis-w-bipath} do not impact the unmatched state $\o$. We thus propose an additional loss to supervise it. 
Particularly, we aim at encouraging background and occluded object regions in images $(I, I', J)$ depicting the same object class, to be predicted as unmatchable. 
Nevertheless, since the locations of the object in $(I, J)$ are unknown during training, we cannot get direct supervision. To overcome this, we introduce an image $A$, depicting a different semantic content than the triplet. 
We then supervise the unmatched state by guiding the mode of the distribution between $A$ and $I$ to be in the unmatched state for all pixels of the images. 
The corresponding learning objective on the non-matching image pair $(I, A)$ is defined as follows, and illustrated in Fig.~\ref{fig:neg-loss},
\begin{equation}
\label{eq:neg-loss}
    L_{\text{PNeg}} = \sum_i \mathcal{B}(\widehat{P}_{A \leftarrow I}( \o |  i) \,  , \,   p_{\text{neg}}) 
\end{equation}
$\mathcal{B}$ denotes the binary cross-entropy and we set $p_{\text{neg}}=0.9$. 

\subsection{Final Training Objectives}

Finally, we introduce our final weakly-supervised objective, the Probabilistic Warp Consistency, as a combination of our previously introduced PW-bipath~\eqref{eq:vis-w-bipath}, PWarp-supervision~\eqref{eq:ce-warpsup} and PNeg~\eqref{eq:neg-loss} objectives. We additionally propose a strongly-supervised approach, benefiting from our losses while also leveraging keypoint annotations.

\parsection{Weak supervision} In this setup, we assume that only image-level class labels are given, such that each image pair is either positive, \ie depicting the same object class, or negative, \ie representing different classes, following~\cite{Rocco2018b, DCCNet, MinLPC20}. We obtain our final weakly-supervised objective by combining the PW-bipath~\eqref{eq:vis-w-bipath} and PWarp-supervision~\eqref{eq:ce-warpsup} losses applied to positive image pairs, with our negative probabilistic objective~\eqref{eq:neg-loss} on negative image pairs. 
\begin{equation}
\label{eq:weak-loss}
    L_{\text{weak}} = L_{\text{vis-PW-bi}} + \lambda_{\text{P-warp-sup}} L_{\text{P-warp-sup}} + \lambda_{\text{PNeg}} L_{\text{PNeg}}
\end{equation}
Here, $\lambda_{\text{P-warp-sup}}$ and $\lambda_{\text{PNeg}}$ are weighting factors.

\parsection{Strong supervision} We extend our approach to the strongly-supervised regime, where keypoint match annotations are given for each training image pair. Previous approaches~\cite{CATs, CHM, ANCNet} leverage these annotations by training semantic networks with a keypoint objective $L_{\text{kp}}$.  
Our final strongly-supervised objective is defined as the combination of the keypoint loss with our PW-bipath~\eqref{eq:vis-w-bipath} and PWarp-supervision~\eqref{eq:ce-warpsup} objectives. Note that we do not include our explicit occlusion modelling, \ie the unmatched state and its corresponding loss~\eqref{eq:neg-loss} on negative image pairs. This is to ensure fair comparison to previous strongly-supervised approaches, which solely rely on keypoint annotations, and not on image-level labels, required for our loss~\eqref{eq:neg-loss}. 
\begin{equation}
\label{eq:strong-loss}
        L_{\text{strong}} =  L_{\text{vis-PW-bi}} +  \lambda_{\text{P-warp-sup}} L_{\text{P-warp-sup}} + \lambda_{\text{kp}} L_{\text{kp}}
\end{equation}
Here, $\lambda_{\text{vis-PW-bi}}$ and $\lambda_{\text{kp}}$ also are weighting factors.

%% file: 04_experimental_work.tex
\section{Experimental Results}
\label{sec:exp}

We evaluate our weakly-supervised learning approach for two semantic networks. The benefits brought by the combination of our probabilistic losses with keypoint annotations are also demonstrated for four recent networks.
We extensively analyze our method and compare it to previous approaches, setting a new state-of-the-art on multiple challenging datasets.

\subsection{Networks and Implementation Details}

For weak supervision, we integrate our approach~\eqref{eq:weak-loss} into baselines SF-Net~\cite{SFNet} and NC-Net~\cite{Rocco2018b}. It leads to our weakly-supervised \textbf{PWarpC-SF-Net} and \textbf{PWarpC-NC-Net} respectively.  
We also apply our strongly-supervised loss~\eqref{eq:strong-loss} to baselines SF-Net, NC-Net, DHPF~\cite{MinLPC20} and CATs~\cite{CATs}, resulting in respectively \textbf{PWarpC-SF-Net*}, \textbf{PWarpC-NC-Net*}, \textbf{PWarpC-DHPF} and \textbf{PWarpC-CATs}. 
For fair comparison, we additionally train a strongly-supervised baseline for both SF-Net and NC-Net, referred to as SF-Net* and NC-Net*. Note that for all methods, the strongly-supervised baseline is trained with only $L_{\text{kp}}$, which is defined as the cross-entropy loss for SF-Net*, NC-Net* and DHPF, and the End-Point-Error objective after applying soft-argmax~\cite{SFNet} for CATs. 
To convert the predicted probabilistic mapping to point-to-point matches for evaluation, all networks trained with our PWarpC objectives employ the argmax operation, except for PWarpC-CATs where we adopt the same soft-argmax as in the baseline CATs~\cite{CATs}.  
Additional details on the integration of our objectives for each architecture are provided in the appendix, Sec.~\ref{sec-sup:impl}-\ref{sec-sup:dhpf}. 
We train all networks on PF-Pascal~\cite{PFPascal}, using the splits of~\cite{SCNet}.
The results when trained on SPair-71K are further presented in the appendix, Sec.~\ref{sup:spair}. 

\newcommand{\best}[1]{\textcolor{BrickRed}{\textbf{#1}}}
\newcommand{\second}[1]{\textcolor{NavyBlue}{\textit{#1}}}
\newcommand{\lowreso}[1]{\textcolor{dark-gray}{#1}}

\begin{table*}[t]
\centering
\vspace{-5mm}
\resizebox{0.93\textwidth}{!}{%
\begin{tabular}{l@{~~}ll|ccc|ccc|cc|c@{~~}c@{~~}c@{~~}c}
\toprule
 & &  & \multicolumn{3}{c}{\textbf{PF-Pascal}} & \multicolumn{3}{c}{\textbf{PF-Willow}} & \multicolumn{2}{c}{\textbf{Spair-71K}} & \multicolumn{4}{c}{\textbf{TSS}}  \\
 & &   & \multicolumn{3}{c}{PCK @ $\alpha_{img}$} & \multicolumn{3}{c}{PCK @ $\alpha_{bbox-kp}$}  & \multicolumn{2}{c}{PCK @ $\alpha_{bbox}$} & \multicolumn{4}{c}{PCK @ $\alpha_{img} \,\, , \alpha=0.05$}\\
 
 & Methods  & Reso & $0.05$ &  $0.10$  &  $0.15$ & $0.05$ &  $0.10$  &  $0.15$  & $0.05$ &  $0.10$ & FG3DCar & JODS & Pascal & Avg.  \\ \midrule
S & UCN$_{\text{res101}}$~\cite{ucn}  & - & - & 75.1 & - & - & - & - & - & 17.7 & - & - & - & - \\
& SCNet$_{\text{VGG16}}$~\cite{SCNet}   & - & 36.2 & 72.2 & 82.0 & - & - & -  & - & -  & - & - & - & - \\
& HPF$_{\text{res101}}$~\cite{HPF} & max $300$ & 60.1 & 84.8 & 92.7 & 45.9 & 74.4 & 85.6 & - & - & 93.6 & 79.7 & 57.3 & 76.9  \\ 

& SCOT$_{\text{res101}}$~\cite{SCOT} & max 300 & 63.1 & 85.4 & 92.7 & 47.8 & 76.0 & 87.1 &  - & - & 95.3 & 81.3 & 57.7 & 78.1 \\ 
& ANC-Net$_{\text{res101}}$~\cite{ANCNet} & - & - & 86.1 &  - & - & - & - & - & 28.7 & - & - & - & - \\
& CHM$_{\text{res101}}$~\cite{CHM} & $240$ & \lowreso{80.1} & \lowreso{91.6} & - & - & - & -  &  - & - & - & - & - & - \\
& PMD$_{\text{res101}}$~\cite{PMD}  & - & - & 90.7 & - & - & 75.6 & - &  - & -  & - & - & - & - \\
& PMNC$_{\text{res101}}$~\cite{PMNC} & - & \best{82.4} & 90.6 & - & - & - & - &  - & 28.8 & - & - & - & -\\
& MMNet$_{\text{res101}}$~\cite{MMNet} & $224 \times 320$ & \lowreso{77.7} & \lowreso{89.1} & \lowreso{94.3} & \lowreso{-} & \lowreso{-}  & \lowreso{-} &  \lowreso{-}& \lowreso{-} & \lowreso{-} & \lowreso{-} & \lowreso{-} & \lowreso{-} \\
& DHPF$_{\text{res101}}$~\cite{MinLPC20} & $240$ & \lowreso{75.7} &  \lowreso{90.7} & \lowreso{95.0} & \lowreso{41.4} $^{\dagger}$ & \lowreso{67.4} $^{\dagger}$ & \lowreso{81.8} $^{\dagger}$  &  \lowreso{15.4} $^{\dagger}$ & \lowreso{27.4}  & \lowreso{-} & \lowreso{-} & \lowreso{-} & \lowreso{-}\\
& CATs$_{\text{res101}}$~\cite{CATs} & $256$ & \lowreso{67.5} & \lowreso{89.1} & \lowreso{94.9} & \lowreso{37.4} $^\dagger$  & \lowreso{65.8} $^\dagger$  & \lowreso{79.7} $^\dagger$  &  \lowreso{10.9} $^\dagger$  & \lowreso{22.4} $^\dagger$ & \lowreso{-} & \lowreso{-} & \lowreso{-} & \lowreso{-} \\
& CATs-ft-features$_{\text{res101}}$~\cite{CATs} & $256$ & \lowreso{75.4} & \lowreso{92.6} & \lowreso{96.4} & \lowreso{40.9} $^\dagger$  &  \lowreso{69.5} $^\dagger$ & \lowreso{83.2} $^\dagger$  &  \lowreso{13.6} $^\dagger$ & \lowreso{27.0} $^\dagger$ & \lowreso{-} & \lowreso{-} & \lowreso{-} & \lowreso{-} \ \\

\cdashlinelr{2-15}
& CATs$_{\text{res101}}$~\cite{CATs} & ori $^\dagger$ & 67.3 & 88.6 & 94.6 &  41.6 & 68.9 & 81.9 &  10.8 & 22.1 & 89.5 & 76.0 & 58.8 & 74.8 \\
& \textbf{PWarpC-CATs}$_{\text{res101}}$ & ori  & 67.1 & 88.5 & 93.8 & 44.2 & 71.2 & 83.5 & 12.2 & 23.3 & 93.2 & 83.4 & 70.7 & 82.4 \\ 

\cdashlinelr{2-15}
& CATs-ft-features$_{\text{res101}}$~\cite{CATs} & ori $^\dagger$ & 76.8 & \second{92.7} & \best{96.5} & 45.2 & 73.2 & 85.2  &  13.7 & 26.8  & 92.1 & 78.9 & 64.2 & 78.4 \\

& \textbf{PWarpC-CATs-ft-features}$_{\text{res101}}$ & ori  & \second{79.8} & 92.6 & \second{96.4} & \second{48.1} & 75.1 & 86.6 & 15.4 & 27.9 & \second{95.5} & \second{85.0} & \second{85.5} & \second{88.7} \\ 

\cdashlinelr{2-15}

& DHPF$_{\text{res101}}$~\cite{MinLPC20} & ori &  77.3 & 91.7 & 95.5 & 44.8 & 70.6 & 83.2 &  15.3 & 27.5 & 88.2 & 71.9 & 56.6 &  72.2 \\
& \textbf{PWarpC-DHPF}$_{\text{res101}}$ &  ori & 79.1 & 91.3 & 96.1 & \best{48.5} & 74.4 & 85.4 & 16.4 & 28.6 & 89.1 & 74.1 & 59.7  & 74.3 \\
\cdashlinelr{2-15}

& NC-Net*$_{\text{res101}}$  & ori  & 78.6 & 91.7 & 95.3 & 43.0 & 70.9 & 83.9 & \second{17.3} & 32.4 & 92.3 & 76.9 & 57.1 & 75.3 \\  
& \textbf{PWarpC-NC-Net*}$_{\text{res101}}$ & ori  & 79.2 & 92.1 & 95.6 & 48.0 & \second{76.2} & \second{86.8} & \best{21.5} & \best{37.1} & \best{97.5} & \best{87.8} & \best{88.4} & \best{91.2} \\ 

\cdashlinelr{2-15}

&  SF-Net*$_{\text{res101}}$ & ori &  78.7 & \best{92.9} & 96.0 & 43.2 & 72.5 & 85.9  & 13.3 &  27.9 & 88.0 & 75.1 & 58.4 & 73.8 \\  
& \textbf{PWarpC-SF-Net*}$_{\text{res101}}$ & ori & 78.3 & 92.2 & 96.2 & 47.5 & \best{77.7} & \best{88.8} & \second{17.3} & \second{32.5} & 94.9 & 83.4 & 74.3 & 84.2 \\ 

\midrule  \midrule
U & CNNGeo$_{\text{res101}}$~\cite{Rocco2017a} & ori  & 41.0 & 69.5 & 80.4 & 36.9 & 69.2 & 77.8  &  - & 18.1 & 90.1 & 76.4 & 56.3 & 74.4 \\
 & PARN$_{\text{res101}}$~\cite{Jeon} & ori & - & - & - & - & - & - & - & - & 89.5  & 75.9 & 71.2  & 78.8 \\
 & GLU-Net$_{\text{vgg16}}$~\cite{GLUNet} & ori    & 42.2 & 69.1 & 83.1 &  30.4 & 57.7 & 72.9 & - & - &  93.2    &   73.3  &   71.1    &    79.2  \\ 
 & Semantic-GLU-Net$_{\text{vgg16}}$~\cite{GLUNet, warpc} & ori &  48.3 & 72.5 & 85.1 & 39.7 & 67.5 & 82.1 &  7.6 & 16.5 & 95.3 & 82.2 & 78.2 & 85.2  \\  
& A2Net$_{\text{res101}}$~\cite{SeoLJHC18} & - & 42.8 &  70.8 & 83.3 & 36.3 & 68.8 & 84.4 &  - & 20.1 & - & - & - & - \\ 
& PMD$_{\text{res101}}$~\cite{PMD} & - &  - & 80.5 & - & - & 73.4 & - & - & - & - & - & - & - \\ \midrule

M & SF-Net$_{\text{res101}}$~\cite{SFNet} & $288$ / ori & \lowreso{53.6} & \lowreso{81.9} & \lowreso{90.6} & 46.3 & 74.0 & 84.2 &  - & - & - & - & - & -\\
 & SF-Net$_{\text{res101}}$~\cite{SFNet} & ori $^\dagger$ & 59.0 & 84.0 & 92.0 & 46.3 & 74.0 & 84.2  & 11.2 & 24.0 & 90.8 & 78.6 & 58.0 & 75.8  \\\midrule
W & \textbf{PWarpC-SF-Net}$_{\text{res101}}$ & ori & \best{65.7} & \best{87.6} & \second{93.1} & \second{47.5} & \best{78.3} & \best{89.0} & 17.6 & 33.5 &  95.1 & 84.7 & 76.8 & 85.5 \\

\cdashlinelr{2-15}

 & WarpC-SF-Net$_{\text{res101}}$ $\diamond$ ~\cite{SFNet, warpc}  & ori&  \second{64.9} & \second{86.1} & 92.2 & 46.9 & \second{76.6} & 87.9 & 13.1 & 26.6 & 95.7 & 82.3 & 68.8 & 82.2 \\

& WeakAlign$_{\text{res101}}$~\cite{Rocco2018a}&  ori/ ori / - &  49.0 & 75.8 & 84.0 & 38.2 & 71.2 & 85.8 &  - & 21.1 & 90.3  & 76.4 & 56.5  & 74.4\\ 
& RTNs$_{\text{res101}}$~\cite{Kim2018} & - &  55.2 & 75.9 & 85.2 & 41.3 & 71.9 & 86.2  & - & - & 90.1 & 78.2 & 63.3 & 77.2  \\

& DCCNet$_{\text{res101}}$~\cite{DCCNet}  & $240$ / ori / - &  \lowreso{55.6} & \lowreso{82.3} & \lowreso{90.5} &43.6 & 73.8 & 86.5  & - & 26.7 & 93.5  & 82.6 & 57.6  & 77.9 \\
& SAM-Net$_{\text{vgg19}}$~\cite{Kim2019} & - & 60.1 & 80.2 & 86.9 & - &  - & - & - & - & \second{96.1} & 82.2 & 67.2 & 81.8\\ 
& DHPF$_{\text{res101}}$~\cite{MinLPC20} & $240$ &  \lowreso{56.1} & \lowreso{82.1} & \lowreso{91.1} & \lowreso{40.5} $^{\dagger}$ & \lowreso{70.6} $^{\dagger}$ & \lowreso{83.8} $^{\dagger}$  &  \lowreso{14.7} $^{\dagger}$ & \lowreso{28.5} & \lowreso{-} & \lowreso{-} & \lowreso{-} & \lowreso{-} \\ 
& DHPF$_{\text{res101}}$~\cite{MinLPC20} & ori $^{\dagger}$ & 61.2 & 84.1 & 92.4 & 45.1 & 73.6 & 85.0  &  14.7 & 27.8 & - & - & - & - \\\
& GSF$_{\text{res101}}$~\cite{GSF} & - & 62.8 & 84.5 & \best{93.7} & 47.0 & 75.8 & \second{88.9} & - & 33.5 & - & - & - & - \\
& PMD$_{\text{res101}}$~\cite{PMD} & - & - & 81.2 & - & - & 74.7 & - & - & 26.5 & - & - & - & -  \\

 & WarpC-SemGLU-Net$_{\text{vgg16}}$~\cite{warpc} & ori & 62.1 & 81.7 & 89.7 & \best{49.0} & 75.1 & 86.9  &  13.4 $^{\dagger}$ & 23.8 $^{\dagger}$ & \best{97.1} & 84.7 & \second{79.7} & \second{87.2} \\ 
 
 & NC-Net$_{\text{res101}}$~\cite{Rocco2018b} &  $240$ / ori / - & \lowreso{54.3} & \lowreso{78.9} & \lowreso{86.0} & 44.0 & 72.7 & 85.4  &  - & \lowreso{26.4} & - & - & - & - \\ 

 & WarpC-NC-Net$_{\text{res101}}$ $\diamond$ ~\cite{Rocco2018b, warpc}  &  ori & 59.1 & 75.0 & 81.2 & 44.6 & 70.1 & 81.3 & \second{18.0} & \second{35.0} & 95.8 & \second{87.5} & 79.3 & 87.0 \\
\cdashlinelr{2-15}
 & NC-Net$_{\text{res101}}$~\cite{Rocco2018b} &  ori $^\dagger$  &  60.5 & 82.3 & 87.9 & 44.0 & 72.7 & 85.4 & 13.9 &  28.8 & 94.5 & 81.4 & 57.1 & 77.7 \\
& \textbf{PWarpC-NC-Net}$_{\text{res101}}$  &  ori & 64.2 & 84.4 & 90.5 & 45.0 & 75.9 & 87.9 & \best{18.2} & \best{35.3} & 95.9 & \best{88.8} & \best{82.9} & \best{89.2} \\  
\bottomrule
\end{tabular}%
}\vspace{-2mm}
\caption{PCK [\%] obtained by different state-of-the-art methods on the PF-Pascal~\cite{PFPascal}, PF-Willow~\cite{PFWillow}, SPair-71K~\cite{spair} and TSS~\cite{Taniai2016} datasets. All approaches are trained on the training set of PF-Pascal, except for~\cite{GLUNet}. The results when trained on SPair-71K are further presented in the appendix, Sec.~\ref{sup:spair}. \textbf{S} denotes strong supervision using keypoint match annotations, \textbf{M} refers to using ground-truth object segmentation mask, \textbf{U} is fully unsupervised requiring only single images, and \textbf{W} refers to weakly-supervised with image-level class labels. Each method evaluates with ground-truth annotations resized to a specific resolution. However, using different ground-truth resolutions leads to slightly different results. We therefore use the standard setting of evaluating on the original resolution (\textbf{ori}) and gray the results computed with the ground-truth annotations at a different size. When needed, we re-compute metrics of baselines using the provided pre-trained weights, indicated by $^\dagger$. For each of our PWarpC networks, we compare to its corresponding baseline within the dashed-lines. For completeness, we also train the baseline networks using the weakly-supervised mapping-based Warp Consistency objective~\cite{warpc}, indicated with $\diamond$. Best and second best results are in red and blue respectively. } 
\label{tab:sem}
\vspace{-4mm}
\end{table*}

\subsection{Experimental Settings}

We evaluate our networks on four standard benchmark datasets for semantic matching, namely PF-Pascal~\cite{PFPascal}, PF-Willow~\cite{PFWillow}, SPair-71K~\cite{spair} and TSS~\cite{Taniai2016}. Results on Caltech-101~\cite{caltech} are further shown in appendix~\ref{sup:caltech}. 

\parsection{Datasets} The \textbf{PF-Pascal}, \textbf{PF-Willow} and \textbf{SPair-71K} are keypoint datasets, which respectively contain 1341, 900 and 70958 image pairs from 20, 4 and 18 categories. Images have dimensions ranging from $102 \times 300$ to $500 \times 500$.  
\textbf{TSS} is the only dataset providing dense flow field annotations for the foreground object in each pair. It contains 400 image pairs, divided into three groups: FG3DCAR, JODS, and PASCAL, according to the origins of the images. 

\parsection{Metrics} We adopt the standard metric, Percentage of Correct Keypoints (PCK), with a pixel threshold of $\alpha_{\tau} \cdot \max(h_s^{\tau}, w_s^{\tau})$. Here, $h_{s}$ and $w_{s}$ are either the dimensions of the source image or the dimensions of the object bounding box in the source image, such as $\tau \in \left \{  \text{img} , \text{bbox} \right \}$. Because PF-WILLOW does not contain bounding-box annotations, approximate bounding box dimensions are computed based on keypoint annotations. However, previous methods use two different schemes to calculate the bounding box: the first (bbox-kp) uses two keypoint positions to approximate the bounding box that tightly wraps the object whereas in the second scheme (bbox-loose), the bounding box loosely covers the object as it uses only a single keypoint position. While the second scheme (bbox-loose) typically yields better PCK results, we adopt the first one  (bbox-kp) as it provides a better estimate of object bounding boxes. 

\begin{figure}[t]
\centering%
\subfloat[NC-Net \cite{Rocco2018b}]{\includegraphics[width=0.23\textwidth, trim=0 0 0 10, clip]{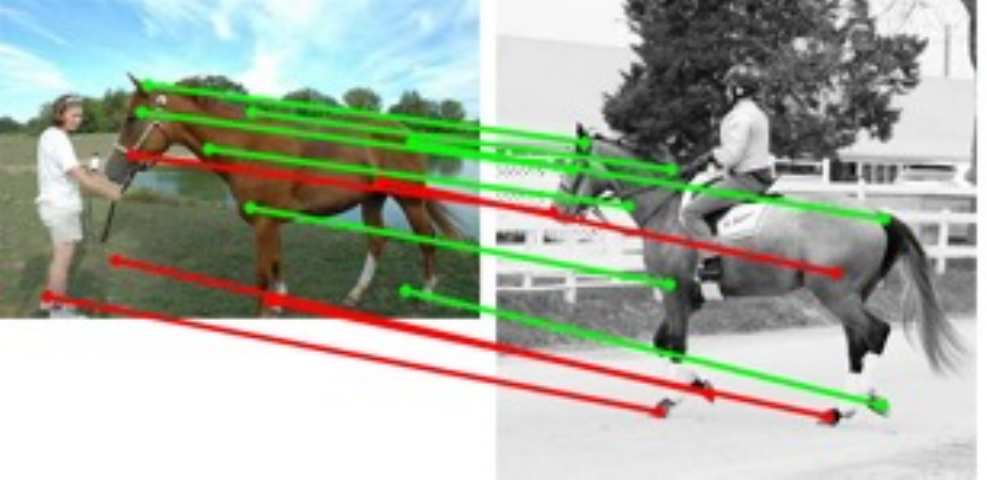}}~
\subfloat[\textbf{PWarpC-NC-Net} (Ours) ]{\includegraphics[width=0.23\textwidth, trim=0 0 0 10, clip]{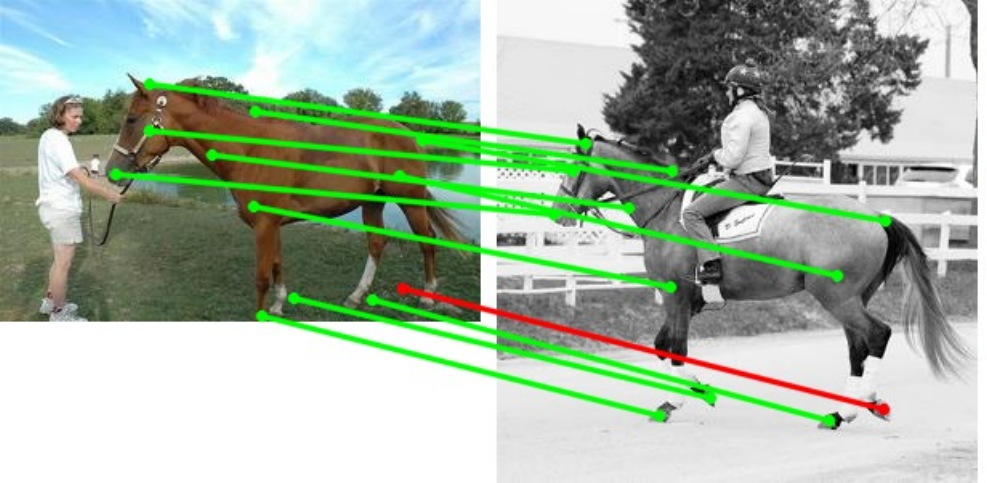}} \\

\subfloat[SF-Net \cite{SFNet}]{\includegraphics[width=0.23\textwidth]{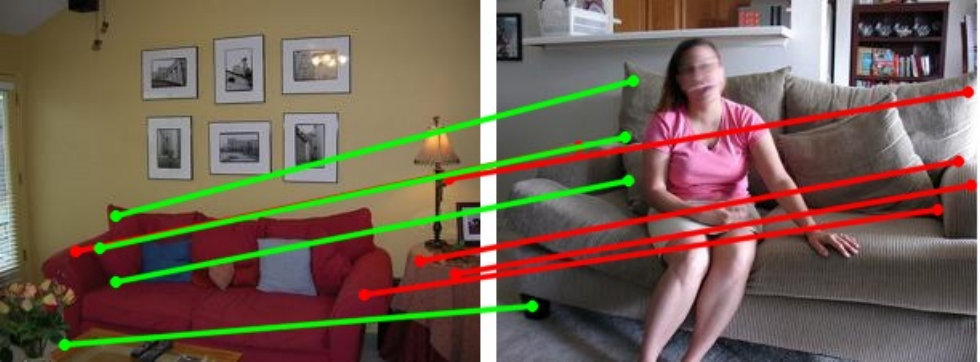}}~
\subfloat[\textbf{PWarpC-SF-Net} (Ours) ]{\includegraphics[width=0.23\textwidth]{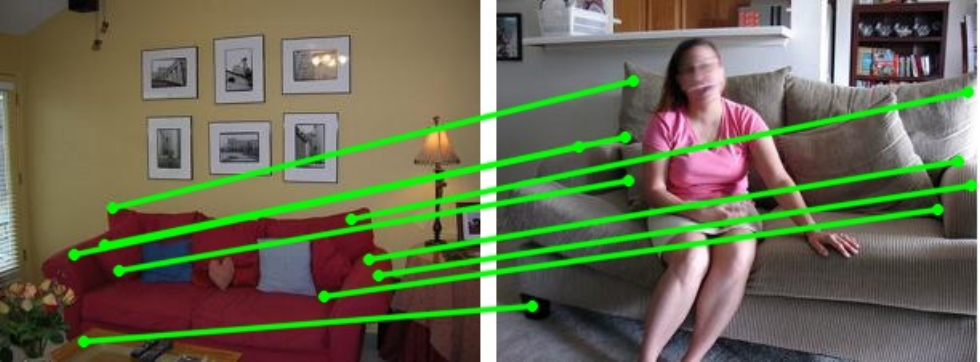}}
\vspace{-3mm}
\caption{Example predictions of baselines NC-Net~\cite{Rocco2018b} and SF-Net~\cite{SFNet}, compared to our weakly-supervised PWarpC-NC-Net and PWarpC-SF-Net. Green and red line denotes correct and wrong predictions, respectively, with respect to the ground-truth. 
}\vspace{-4mm}
\label{fig:examples}
\end{figure}

\subsection{Results} 

We present results on PF-Pascal, PF-Willow, SPair-71K and TSS in Tab.~\ref{tab:sem}. All approaches are trained on the training set of PF-Pascal, except for~\cite{GLUNet}. The results when trained on SPair-71K are further presented in the appendix, Sec.~\ref{sup:spair}. A few previous approaches compute the PCK metrics after resizing the annotations to a different resolution than the original. Nevertheless, we found that in practise, the annotation resolution can lead to notable variations in results, as evidenced for DHPF or CATs in Tab.~\ref{tab:sem}.
For fair comparison, we thus compute the metrics on the standard setting, \ie the original image size, and re-compute the PCK in this setting for baseline works if necessary. We also indicate the annotation size used, whenever reported by the authors or provided in their public implementation. 

\parsection{Weak supervision (W)} In Tab.~\ref{tab:sem}, bottom part, we compare approaches trained with weak-supervision in the form of image labels. In this setting, our PWarpC networks are trained with $L_{\text{weak}}$ in~\eqref{eq:weak-loss}. While bringing improvements on the PF-Pascal dataset itself, our approach PWarpC-NC-Net most notably achieves widely better generalization properties, with impressive $4.4 \%$ (+ $3.2$),  $22.6 \%$ (+ $6.5$) and $14.8 \%$ (+ $11.5$) relative (and absolute) gains compared to the baseline NC-Net on PF-Willow ($\alpha=0.1$), SPair-71K ($\alpha=0.1$) and TSS ($\alpha=0.05$) respectively. Our PWarpC-NC-Net thus sets a new state-of-the-art on SPair-71K and TSS among weakly-supervised methods trained on PF-Pascal.  

Even though it utilizes a lower degree of supervision, our approach PWarpC-SF-Net also significantly outperforms the baseline SF-Net, which is trained with mask supervision (M), on all datasets. In particular, it shows a relative (and absolute) gain of $4.3 \%$ (+ $3.6$), $5.8 \%$ (+ $4.3$) and $39.6 \%$ (+ $9.5$) on respectively PF-Pascal, PF-Willow and SPair-71K for $\alpha=0.1$, and of $10.9\%$ (+ $8.3$) on TSS for $\alpha=0.05$. This makes our PWarpC-SF-Net the new state-of-the-art across all unsupervised (U), weakly-supervised (W) and mask-supervised (M) approaches on PF-Pascal and PF-Willow. Example predictions are shown in Fig.~\ref{fig:examples}

\parsection{Strong supervision (S)} In the top part of Tab.~\ref{tab:sem}, we evaluate networks trained with strong supervision, in the form of key-point annotations. Our strongly-supervised PWarpC approaches are trained with our $L_{\text{strong}}$~\eqref{eq:strong-loss}. 
For all networks, while the results are on par with the baselines on PF-Pascal, the PWarpC networks show drastically better performance on PF-Willow, SPair-71K and TSS compared to their respective baselines. 
PWarpC-SF-Net* and PWarpC-NC-Net* thus set a new state-of-the-art on respectively PF-Willow, and the SPair-71K and TSS datasets, across all strongly-supervised approaches trained on PF-Pascal. 
Finally, while most works focus on designing novel semantic architectures, we here show that the right training strategy bridges the gap between architectures.

\begin{table}[b]
\centering
\vspace{-2mm}
\resizebox{0.47\textwidth}{!}{%
\begin{tabular}{l@{~~}l|cc|cc|c|c}
\toprule
& & \multicolumn{2}{c}{\textbf{PF-Pascal}} & \multicolumn{2}{c}{\textbf{PF-Willow}} & \multicolumn{1}{c}{\textbf{Spair-71K}}   & \multicolumn{1}{c}{\textbf{TSS}}  \\
& &  \multicolumn{2}{c}{$\alpha_{img}$} & \multicolumn{2}{c}{$\alpha_{bbox-kp}$} & \multicolumn{1}{c}{$\alpha_{bbox}$} & \multicolumn{1}{c}{$\alpha_{img}$} \\
 
& Methods  & $0.05$ &  $0.10$  &  $0.05$ &  $0.10$  & $0.10$ & $0.05$\\ \midrule
I & SF-Net baseline & 59.0 & 84.0 & 46.3 & 74.0 & 24.0 & 75.8 \\

II & PW-bipath~\eqref{eq:ce}  &  59.1 & 82.3 & 44.9 & 74.3 & 28.0 & 83.4 \\

III & + visibility mask~\eqref{eq:vis-w-bipath}& 61.2 & 83.7 & 46.1 & 75.8 & 28.5 & 78.4 \\ 
IV & + PWarp-supervision~\eqref{eq:ce-warpsup} & 63.0 & 84.9 & 47.0 & 76.9 & 30.7 & 83.5 \\  
V & + PNeg~\eqref{eq:neg-loss} (\textbf{PWarpC-SF-Net}) & \textbf{65.7} & \textbf{87.6} & \textbf{47.5} & \textbf{78.3} & \textbf{33.5} & \textbf{85.5} \\

\midrule 
V & \textbf{PWarpC-SF-Net} (Ours) &  \textbf{65.7} & \textbf{87.6} & \textbf{47.5} & \textbf{78.3} & \textbf{33.5} & \textbf{85.5}  \\
VI & Mapping Warp Consistency~\cite{warpc} &  64.9 & 86.1 & 46.9 & 76.6 & 26.6 & 82.2 \\

VII & PWarp-supervision only \eqref{eq:ce-warpsup} & 52.9 & 74.3 & 38.0 & 66.6 & 27.9 & 79.4  \\
VIII & Max-score~\cite{Rocco2018b} & 52.4 & 76.7 & 31.2 & 59.5 & 24.6 &  74.8 \\ 
IX & Min-entropy~\cite{MinLPC20} & 44.7 & 74.4 & 25.4 & 57.8 & 20.6 & 69.6 \\ 

\bottomrule
\end{tabular}%
}\vspace{-3mm}
\caption{Ablation study (top part) and comparison to alternative objectives (bottom part) for PWarpC-SF-Net. 
}

\label{tab:abl-sfnet}
\end{table}

\begin{figure}[t]
\centering%
(a) Training with mapping-based Warp Consistency~\cite{warpc} \\
\includegraphics[width=0.33\textwidth]{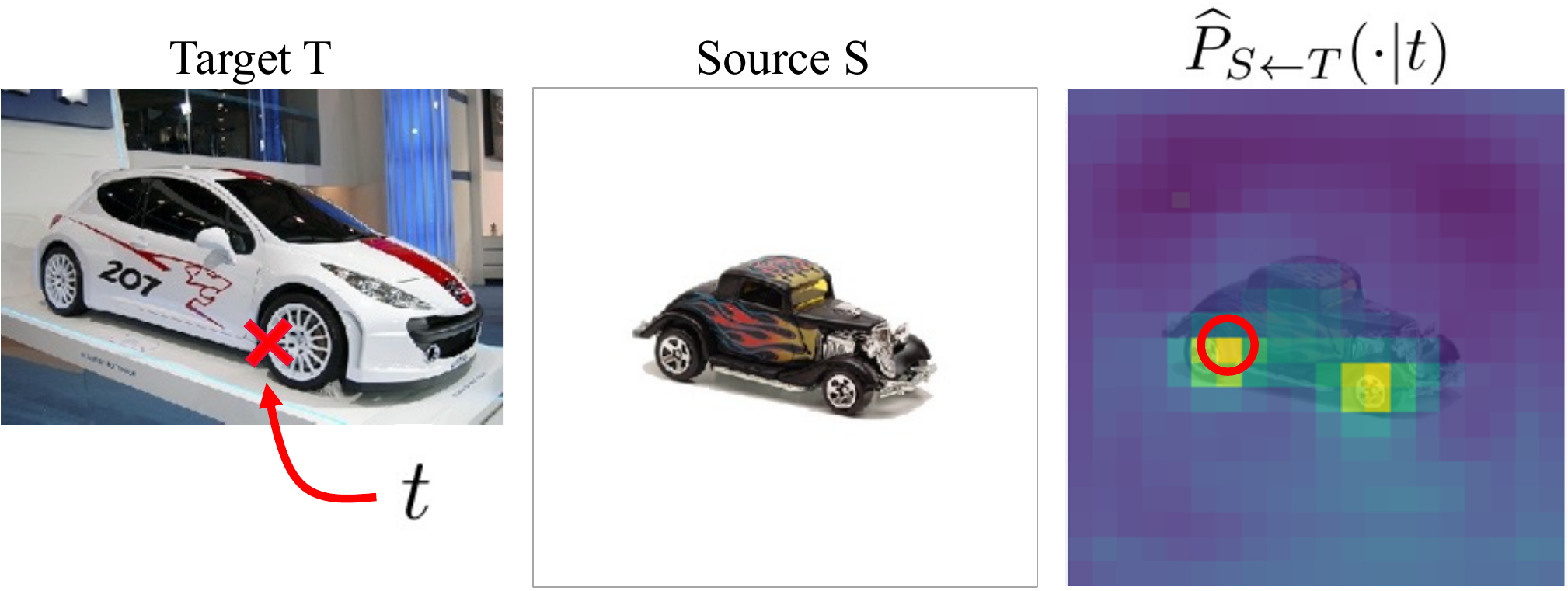} \\
(b) Training with Probabilistic Warp Consistency (\textbf{Ours}) \\
\includegraphics[width=0.45\textwidth]{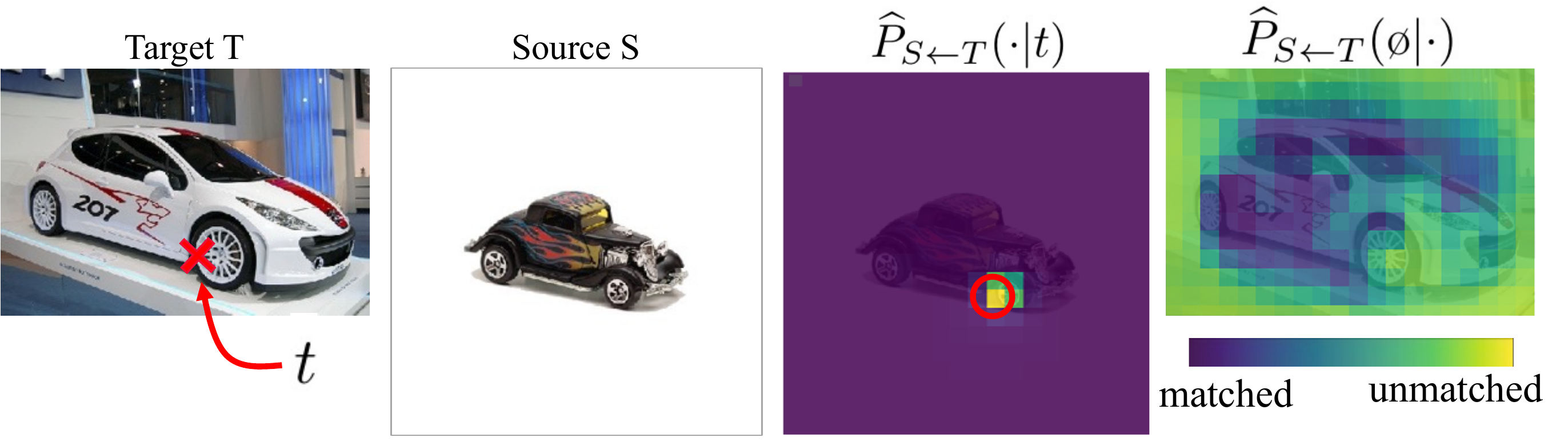}
\vspace{-2mm}
\caption{ In (a), SF-Net is trained using the mapping-based Warp Consistency approach~\cite{warpc}, after converting the cost volume to a mapping through soft-argmax~\cite{SFNet}. It predicts ambiguous matching scores, struggling to differentiate between the car wheels.
Our probabilistic approach (b) instead directly predicts a Dirac-like distribution, whose mode is correct. Here, we also show that our approach identifies most of the background areas as unmatched. 
}
\vspace{-5mm}
\label{fig:method-analysis}
\end{figure}

\subsection{Method Analysis}
\label{sec:method-analysis}

We here perform a comprehensive analysis of our approach in Tab.~\ref{tab:abl-sfnet}. We adopt SF-Net as the base architecture.

\parsection{Ablation study} In the top part of Tab.~\ref{tab:abl-sfnet}, we analyze key components of our approach. The version denoted as (II) is trained using our PW-bipath objective~\eqref{eq:ce}, without the visibility mask. Further introducing our visibility mask~\eqref{eq:vis-w-bipath} in (III) significantly boosts the results since it enables to supervise only in the common visible regions. Note that this version (III) already outperforms the baseline SF-Net (I), while using less annotation (class instead of mask). In (IV), we add our probabilistic warp-supervision~\eqref{eq:ce-warpsup}, leading to a small improvement for all thresholds and on all datasets. From (IV) to (V), we further introduce our explicit occlusion modelling associated with our negative loss~\eqref{eq:neg-loss}, which results in drastically better performance. This version corresponds to our final weakly-supervised PWarC-SF-Net, trained with~\eqref{eq:weak-loss}. 
An example of the regions identified as unmatched by PWarpC-SF-Net is shown in Fig.~\ref{fig:method-analysis}b.

\parsection{Comparison to other losses} In Tab.~\ref{tab:abl-sfnet}, bottom part, we first compare our probabilistic approach~\eqref{eq:weak-loss} corresponding to (V) with the mapping-based warp consistency objective~\cite{warpc}, denoted as (VI). Our approach (V) leads to better performance than warp consistency (VI), with a particularly impressive $6.9 \%$ absolute gain on the challenging SPair-71K dataset. We further illustrate the benefit of our approach on an example in Fig.~\ref{fig:method-analysis}.  Moreover, using \emph{only} the PWarp-supervision loss~\eqref{eq:ce-warpsup} in (VII) results in much worse performance than our Probabilistic Warp Consistency (V). 
Finally, we compare our approach (V) to previous losses applied on cost volumes. 
The versions (VIII) and (IX) are trained with respectively maximizing the max scores~\cite{Rocco2018b}, and minimizing the cost volume entropy~\cite{MinLPC20}. Both approaches lead to poor results, likely caused by the very indirect supervision signal that these objectives provide.

%% file: 05_conclusion.tex
\section{Conclusion}

We propose Probabilistic Warp Consistency, a weakly-supervised learning objective for semantic matching. We introduce multiple probabilistic losses derived both from a triplet of images generated based on a real image pair, and from a pair of non-matching images. When integrated into four recent semantic networks, our approach sets a new state-of-the-art on four challenging benchmarks. 

\vspace{-1mm}
\parsection{Limitations} Since our approach acts on cost volumes, which are memory expensive, it is limited to relatively coarse resolution. This might in turn impact its accuracy. 

%% file: supplementary.tex
\appendix

In this supplementary material, we provide additional details about our approach, experiment settings and results.
In Sec.~\ref{sec-sup:impl}, we first give general implementation details, that apply to all networks we considered. 
We follow by explaining the triplet image creation and the sampling process of our synthetic warps $M_W$ in Sec.~\ref{sec-sup:transfo}.

In Sec.~\ref{sec-sup:sfnet}, we then focus on the training procedure to obtain PWarpC-SF-Net and PWarpC-SF-Net* in more depth.  We continue by explaining the training details of PWarpC-NC-Net and PWarpC-NC-Net* in Sec.~\ref{sec-sup:ncnet}. We subsequently provide training details for PWarpC-CATs and PWarpC-DHPF in respectively Sec.~\ref{sec-sup:cats} and ~\ref{sec-sup:dhpf}. In all aforementioned sections, we provide information about the architecture, its original training strategy, our proposed training approach comprising the sampled transformations $M_W$ and weighting of our losses, as well as implementation details. We also provide additional ablative experiments. 

We then follow by analysing the effect of the strength of the sampled warps $M_W$ in Sec.~\ref{sec-sup:transfo-analysis}. In Sec. \ref{sup:spair}, we also provide results on all four benchmarks PF-Pascal~\cite{PFPascal}, PF-Willow~\cite{PFWillow}, SPair-71K~\cite{spair} and TSS~\cite{Taniai2016}, when the networks are trained or finetuned on SPair-71K instead of PF-Pascal. 
Finally we present more detailed quantitative and qualitative results in Sec.~\ref{sec-sup:results}. In particular, we extensively explain the evaluation datasets and set-up. We also analyze our approach in terms of robustness to view-point, scale, truncation and occlusion. Finally, we further evaluate our approach on the Caltech-101 dataset~\cite{caltech}.  

\section{General implementation details}
\label{sec-sup:impl}

In this section, we provide implementation details, which apply to all our PWarpC networks. 

\parsection{Creating of ground-truth probabilistic mapping $P_W$} Here, we describe how we obtain the ground-truth probabilisitc mapping $P_W$ from the known mapping $M_W$. We first rescale the mapping $M_W$ to the same resolution as the predicted probabilistic mapping $\widehat{P}$. 
We then convert the mapping into a ground-truth probabilistic mapping $P_W$, following this scheme. 
For each pixel position $i'$ in $I'$, we construct the ground-truth 2D conditional probability distribution $P_{W}( \cdot | i') \in \mathbb{R}^{h_{I'} \times w_{I'}}, \in \left[0, 1 \right]$ by assigning a one-hot or a smooth representation of $M(\mathbf{i'})$. In the latter case, following~\cite{ANCNet}, we pick the four nearest neighbours of $M(\mathbf{i'})$ and set their probability according to distance. Then we apply 2-D Gaussian smoothing of size 3 on that probability map. We then vectorize the two dimensions of $P_W( \cdot | i')$, leading to our final known warp probabilistic mapping $P_W \in \mathbb{R}^{h_{I}w_I \times h_{I'}w_{I'}}$. We will specify which representation we used, as either one-hot or smooth, for each loss and each network.

\parsection{Conversion of $P$ to correspondence set} The output of the model is a probabilistic mapping, encoding the matching probabilities for all pairwise match relating an image pair.  
However, for various applications, such as image alignment or geometric transformation estimation, it is desirable to obtain a set of point-to-point image correspondences $M_{I \leftarrow J}$ between the two images. This can be achieved by either performing a hard or soft assignment. 
In the former case, the hard assignment in one direction is done by just taking the most likely match, the mode of the distribution as, 
\begin{equation}
\label{eq:argmax}
    M_{I \leftarrow J}(\mathbf{j}) = \text{arg max}_i \,\, \,\,  P_{I \leftarrow J}(i|j) \\
\end{equation}
In the latter case,  the soft assignment corresponds to soft-argmax. It computes correspondences $M_{I \leftarrow J}(\mathbf{j})$ for individual locations $\mathbf{j}$ of image $J$, as the expected position in $I$ according to the conditional distribution $P_{I \leftarrow J} (.|j)$, 
\begin{equation}
\label{eq:soft-argmax}
    M_{I \leftarrow J}(\mathbf{j}) = \sum_i  \,\,\,\, \mathbf{i} \cdot P_{I \leftarrow J}(i|j) \\
\end{equation}

\parsection{Training details} All networks are trained with PyTorch, on a single NVIDIA TITAN RTX GPU with 24 GiB of memory, within 48 hours, depending on the architecture. 

\section{Triplet creation and sampling of warps $M_W$}
\label{sec-sup:transfo}

\subsection{Triplet creation}

Our introduced learning approach requires to construct an image triplet $(I, I', J)$ from an original image pair $(I, J)$, where all three images must have training dimensions $s \times s$. We follow a similar procedure than in~\cite{warpc}, further described here. 
The original training image pairs $(I, J)$ are first resized to a fixed size $s_r \times s_r$, larger than the desired training image size $s \times s$. 
We then sample a dense mapping $M_W$ of the same dimension $s_r \times s_r$, and create $I'$ by warping image $I$ with $M_W$, as $I' = I \circ M_W$. 
Each of the images of the resulting image triplet $(I, I', J)$ are then centrally cropped to the fixed training image size $s \times s$. The central cropping is necessary to remove most of the black areas in $I'$ introduced from the warping operation with large sampled mappings $M_W$ as well as possible warping artifacts arising at the image borders. 
We then additionally apply appearance transformations to all images of the triplet, such as brightness and contrast changes. 

\subsection{Sampling of warps $M_W$} 

A question raised by our proposed loss formulations~\eqref{eq:ce-warpsup}-\eqref{eq:vis-w-bipath} is how to sample the synthetic warps $M_W$. During training, we randomly sample it from a distribution $M_W \sim p_W$, which we need to design. Here, we also follow a similar procedure than in~\cite{warpc}. 

In particular, we construct $M_W$ by sampling homography, Thin-plate Spline (TPS), or affine-TPS transformations with equal probability. The transformations parameters are then converted to dense mappings of dimension $s_r \times s_r$. Then, we optionally apply horizontal flipping to the each dense mapping with a probability $p_{flip}$. 

Specifically, for homographies and TPS, the four image corners and a $3 \times 3$ grid of control points respectively, are randomly translated in both horizontal and vertical directions, according to a desired sampling scheme. The translated and original points are then used to compute the corresponding homography and TPS parameters. Finally, the transformations parameters are converted to dense mappings. 
For both transformation types, the magnitudes of the translations are sampled according to a uniform distribution with a range $\sigma_H$.
Note that for the uniform distribution, the sampling range is actually $\left [- \sigma_H, \sigma_H \right]$, when it is centered at zero, or similarly $\left [1 - \sigma_H, 1 + \sigma_H \right]$ if centered at 1 for example. 
Importantly, the image points coordinates are previously normalized to be in the interval $\left [ -1 , 1\right]$. Therefore $\sigma_H$ should be within $\left [ 0, 1\right]$. 

For the affine transformations, all parameters, \ie scale, translations, shearing and rotation angles, are sampled from a uniform distribution with range equal to $\tau$, $t$, $\alpha$ and $\alpha$ respectively. The affine scale parameters are sampled within $\left [1-\tau, 1 + \tau \right]$ with center at 1, while for all other parameters, the sampling interval is centered at zero. 

\subsection{List of Hyper-parameters} 

In summary, to construct our image triplet $(I, I', J)$, the hyper-parameters are the following: 

\bp{(i)} $s_r$, the resizing image size, on which is applied $M_W$ to obtain $I'$ before cropping.

\bp{(ii)} $s$, the training image size, which corresponds to the size of the training images after cropping. 

\bp{(iii)} $\sigma_H$, the range used for sampling the homography and TPS transformations.  

\bp{(iv)} $\tau$, the range used for sampling the scaling parameter of the affine transformations. 

\bp{(v)} $t$, the range used for sampling the translation parameter of the affine transformations. 

\bp{(vi)} $\alpha$, the range used for sampling the rotation angle of the affine transformations. It is also used as shearing angle.  

\bp{(vii)} $\sigma_{tps}$, the range used for sampling the TPS transformations, used for the Affine-TPS compositions. 

\bp{(viii)} The probability of horizontal flipping $p_{flip}$. 

\subsection{Hyper-parameters settings}

\parsection{Geometric transformations } For all our PWarpC networks, the mappings $M_W$ are created by sampling homographies, TPS and Affine-TPS transformations with equal probability.
For simplicity, we also use the same range for all three types of transformations. In particular,  we use a uniform sampling scheme with a range equal to $\left [-\sigma_H, \sigma_H \right ]$, where $\sigma_H = \sigma_{tps} = 0.4$.
For the affine transformations, we also sample all parameters, \ie scale, translation, shear and rotation angles, from uniform distributions with ranges respectively equal to $\tau = 0.45$, $t = 0.25$, and $\alpha = \pi/12$ for both angles.  We use these parameters when training on either PF-Pascal~\cite{PFPascal} or SPair-71K~\cite{spair}. 

\parsection{Probability of horizontal flipping} When training on PF-Pascal, we set the probability of horizontal flipping to $p_{flip} = 5 \%$ for all our PWarpC networks, except for PWarpC-NC-Net and PWarpC-NC-Net*, for which we use $p_{flip} = 30 \%$. For training on SPair-71K, we increase this value to $p_{flip} = 15 \%$ for all our PWarpC networks, except for PWarpC-NC-Net and PWarpC-NC-Net*, for which we keep $p_{flip} = 30 \%$. 

\parsection{Appearance transformations} For all experiments and networks, we apply the same appearance transformations to the image triplet $(I, I', J)$.
Specifically,  we convert each image to gray-scale with a probability of 0.2. We then apply color transformations, by adjusting contrast, saturation, brightness, and hue. The color transformations are larger for the synthetic image $I'$ then for the real images $(I, J)$. For the synthetic image $I'$, we additionally randomly invert the RGB channels. 
Finally, on all images of the triplet, we further use a Gaussian blur with a kernel between 3 and 7, and a standard deviation sampled within $\left[ 0.2, 2.0 \right]$, applied with probability of 0.2.

\section{PWarpC-SF-Net and PWarpC-SF-Net*}
\label{sec-sup:sfnet}

We first provide details about the SF-Net~\cite{SFNet} architecture. We also briefly review the training strategy of the original work. We then extensively explain our training approach and the corresponding implementation details, for both our weakly and strongly-supervised approaches, PWarpC-SF-Net and PWarpC-SF-Net* respectively. 
Finally, we provide additional method analysis for this architecture.

\subsection{Details about SF-Net}

\parsection{Architecture} SF-Net is based on a pre-trained ResNet-101 feature backbone, on which are added convolutional adaptation layers at two levels. The predicted feature maps are then used to construct two cost volumes, at two resolutions. After upsampling the coarsest one to the same resolution, the two cost volumes are combined with a point-wise multiplication. While the resulting cost volume is the actual output of the network, it is converted to a flow field through a kernel sotf-argmax operation. Specifically, a fixed Gaussian kernel is applied on the raw cost volume scores to post-process them, before applying SoftMax to convert the cost volume to a probabilistic mapping. From there, the soft-argmax operation transposes it to a mapping. 

For our PWarpC approaches, we do not  use the Gaussian kernel to post-process the predicted matching scores. We simply convert the predicted cost volume into a probabilistic mapping through a SoftMax operation, following eq.~\eqref{eq:softmax} of the main paper. 
Also note that only the adaptation layers are trained. 

\parsection{Training strategy in original work} The original work employs ground-truth foreground object masks as supervision. From single images associated with their segmentation masks, they create image pairs by applying random transformations to both the original images and segmentation masks. Subsequently, they train the network with a combination of multiple losses. In particular, they enforce the forward-backward consistency of the predicted flow, associated with a smoothness objective acting directly on the predicted flow. These losses are further combined with an objective enforcing the consistency of the warped foreground mask of one image with the ground-truth segmentation mask of the other image. 

\subsection{PWarpC-SF-Net and PWarpC-SF-Net*: our training strategy}

\parsection{Warps $W$ sampling} For the weakly-supervised version, we resize the image pairs $(I, J)$ to $s_r \times s_r = 340 \times 340$, sample a dense mapping $M_W$ of the same dimension and create $I'$. Each of the images of the resulting image triplet $(I, I', J)$ is then centrally cropped to $s \times s = 320 \times 320$.  

For the strongly-supervised version, we apply the transformations on images of the same size than the crop, \ie $s_r \times s_r  = s \times s = 320 \times 320$. This is to avoid cropping keypoint annotations.  

When training on PF-Pascal, we apply $5\%$ of horizontal flipping to sample the random mappings $M_W$, while it is increased to $15 \%$ when training on SPair-71K. 

\parsection{Weighting and details on the losses } We found it beneficial to define the known probabilistic mapping $P_W$ with a one-hot representation for our PW-bipath loss~\eqref{eq:vis-w-bipath}, while using a smooth representation instead for the PWarp-supervision~\eqref{eq:ce-warpsup} loss and the keypoint $L_{\text{kp}}$ objective in~\eqref{eq:strong-loss}. Each representation is described in Sec.~\ref{sec-sup:impl}. 

For the weakly-supervision version PWarpC-SF-Net, the weights in~\eqref{eq:weak-loss} are set to $\lambda_{\text{P-warp-sup}} = L_{\text{vis-PW-bi}} / L_{\text{P-warp-sup}}$ and $\lambda_{\text{PNeg}}=1$. 

For the strongly-supervised version, PWarpC-SF-Net*, we use the same weight $\lambda_{\text{P-warp-sup}} = L_{\text{vis-PW-bi}} / L_{\text{P-warp-sup}}$. We additionally set $\lambda_{\text{kp}} = ( L_{\text{P-warp-sup}} + L_{\text{vis-PW-bi}}  )/ L_{\text{kp}}$, which ensure that our probabilistic losses amount for the same than the keypoint loss $L_{\text{kp}}$. 
Moreover, the keypoint loss $L_{\text{kp}}$ is set as the cross-entropy loss, for both PWarpC-SF-Net* and its baseline SF-Net*.

\parsection{Implementation details} 
For our weakly-supervised PWarpC-SF-Net, we set the initial learnable parameter $z$, corresponding to the unmatched state $\o$ for our occlusion modeling, at $z = 0$. 

For both the weakly and strongly-supervised approaches, the SoftMax temperature, corresponding to equation~\eqref{eq:softmax} of the main paper, is set to $\tau= 1.0 / 50.0$, the same than originally used in the baseline for soft-argmax. 
The hyper-parameter used in the estimation of our visibility mask $\widehat{V}$ (eq.~\ref{eq:vis-w-bipath} of the main paper) is set to $\gamma = 0.7$ and to $\gamma = 0.2$ when trained on PF-Pascal or SPair-71K respectively. This is because in SPair-71K, the objects are generally much smaller than in PF-Pascal. 

For training, we use similar training parameters as in baseline SF-Net~\cite{SFNet}. We train with a batch size of 16 for maximum 100 epochs. The learning rate is set to $3.10^{-5}$ and halved after 50. We optionally finetune the networks on SPair-71K for an additional 20 epochs, with an initial learning rate of $1.10^{-5}$, halved after 10 epochs. 
The networks are trained using Adam optimizer~\cite{adam} with weight decay set to zero.

\subsection{Additional analysis}

Here, we first analyse the effect of the kernel applied in the original SF-Net baseline~\cite{SFNet} before converting the predicted cost volume to a probabilistic mapping representation. We also provide the ablation study of our strongly-supervised PWarpC-SF-Net*. Note that the ablation study of the weakly-supervised SF-Net is provided in Tab.~\ref{tab:abl-sfnet} of the main paper. Finally, we show the impact of different losses on negative image pairs, \ie depicting different object classes.

\begin{figure}[t]
\centering%
(a) Training with mapping-based Warp Consistency~\cite{warpc} \\
\includegraphics[width=0.48\textwidth]{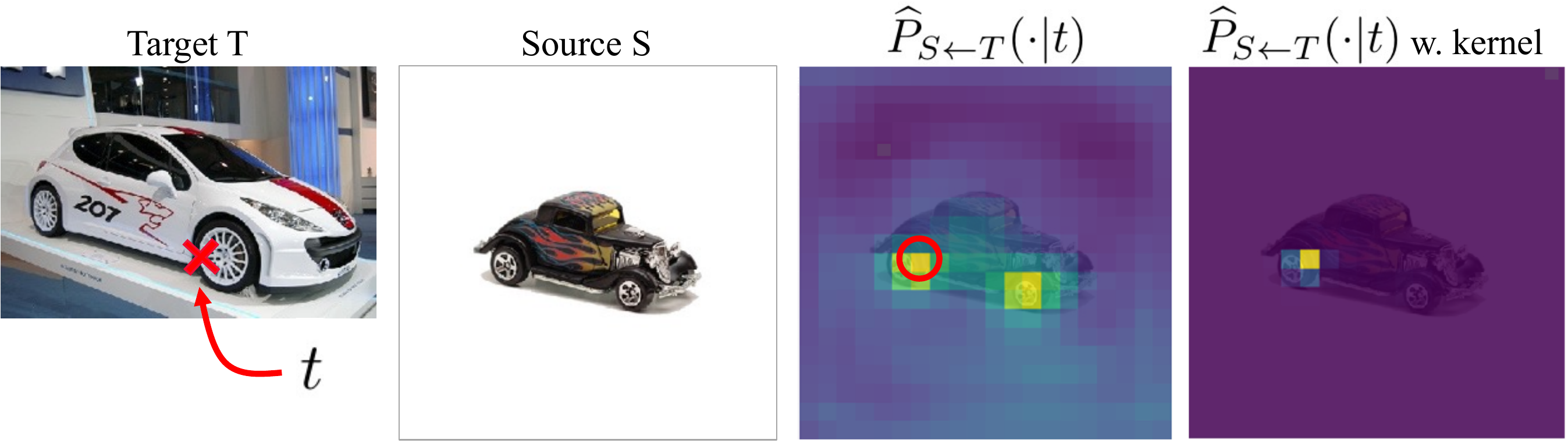}\\
(b) Training with Probabilistic Warp Consistency (\textbf{Ours}) \\
\includegraphics[width=0.33\textwidth]{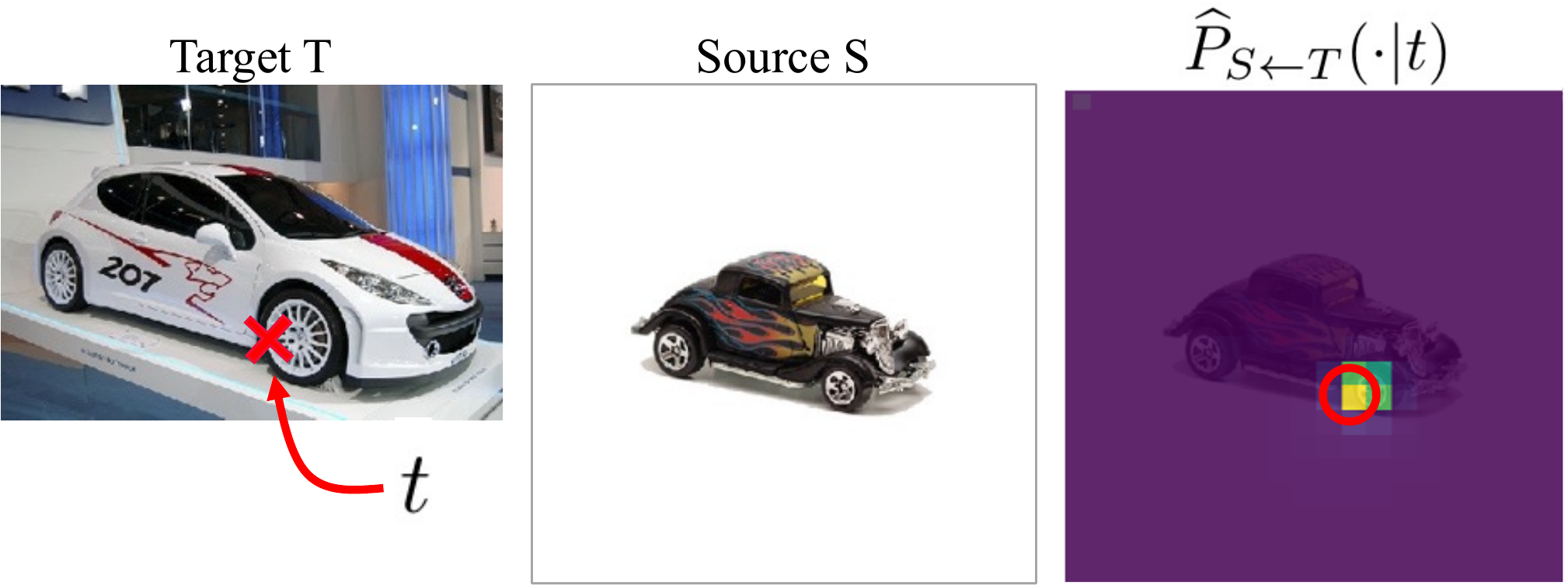}
\vspace{-2mm}
\caption{ In (a), SF-Net is trained using the mapping-based Warp Consistency approach~\cite{warpc}, after converting the cost volume to a mapping through soft-argmax~\cite{SFNet}. It predicts ambiguous matching scores, struggling to differentiate between the car wheels. After applying the kernel, the mode of the distribution corresponds to the wrong wheel. Also note that the kernel is extremely important in that case to post-process the multi-hypothesis distribution. 
Our probabilistic approach (b) instead directly predicts a Dirac-like distribution, whose mode is correct. 
}
\vspace{-5mm}
\label{fig:sfnet-kernel}
\end{figure}

\begin{table}[b]
\centering
\resizebox{0.47\textwidth}{!}{%
\begin{tabular}{l|cc|cc|c|c}
\toprule
& \multicolumn{2}{c}{\textbf{PF-Pascal}} & \multicolumn{2}{c}{\textbf{PF-Willow}} & \multicolumn{1}{c}{\textbf{Spair-71K}}   & \multicolumn{1}{c}{\textbf{TSS}}  \\
&  \multicolumn{2}{c}{$\alpha_{img}$} & \multicolumn{2}{c}{$\alpha_{bbox-kp}$} & \multicolumn{1}{c}{$\alpha_{bbox}$} & \multicolumn{1}{c}{$\alpha_{img}$} \\
 
Methods  & $0.05$ &  $0.10$  &  $0.05$ &  $0.10$  & $0.10$ & $0.05$\\ \midrule
 SF-Net* & \textbf{78.7} & \textbf{92.9} & 43.2 & 72.5 & 27.9 & 73.8  \\
+ Vis-aware PW-bipath~\eqref{eq:vis-w-bipath} & 77.1 & 91.6 & \textbf{47.8} & \textbf{77.9} & 31.1 &80.3  \\
+ PWarp-supervision~\eqref{eq:ce-warpsup} & 78.3 & 92.2 & 47.5 & 77.7 & \textbf{32.5} & \textbf{84.2} \\

\bottomrule
\end{tabular}%
}\vspace{-3mm}
\caption{Ablation study for strongly-supervised PWarpC-SF-Net*.  We incrementally add each component. 
}

\label{tab:abl-sfnet*}
\end{table}

\parsection{Effect of kernel} Baseline SF-Net relies on a kernel soft-argmax strategy to convert the predicted cost volume to a mapping. In particular, the kernel is applied on the cost volume before applying SoftMax (eq.~\eqref{eq:softmax}), which transposes it to a probabilistic mapping. From there, soft-argmax is used to obtain a mapping. Nevertheless, we observe that this kernel is extremely important in order to post-process the matching scores. This is shown in Fig.~\ref{fig:sfnet-kernel}. 
In contrast, our approach Probabilistic Warp Consistency, which directly acts on the predicted dense matching scores, produces clean, Dirac-like conditional distributions, without relying on any post-processing operations. 

\parsection{Ablation study for strongly-supervised PWarpC-SF-Net*} In Tab.~\ref{tab:abl-sfnet*}, we analyse key components of our approach PWarpC-SF-Net*. From the strongly-supervised baseline SF-Net*, adding our probabilistic PW-bipath objective leads to a significant improvement on the PF-Willow, SPair-71K and TSS datasets. Further including our PWarp-supervision objective results in additional gains on SPair-71K and TSS.

\parsection{Comparisons to alternative negative losses}
In Tab.~\ref{tab:neg-loss-analysis}, we compare combining our PW-bipath and PWarp-supervision objectives on image pairs of the same label, with  different losses on images pairs showing different object classes, \ie on negative image pairs. In the version denoted as (IV), we introduce our explicit occlusion modeling (Sec. 4.3 of the main paper), trained with our probabilistic negative loss $L_{\text{PNeg}}$. 
In (V) and (VI), we instead combine our probabilistic objectives on the positive image pairs~\eqref{eq:ce} (III), with an additional objective, minimizing the max scores or the negative entropy of the cost volume respectively. While it brings a small improvement with respect to version (III), the resulting network performances in (V) and (VI) are far lower than when trained with our final combination~\eqref{eq:weak-loss}, which corresponds to version (IV). 

\begin{table}[t]
\centering
\vspace{-3mm}
\resizebox{0.48\textwidth}{!}{%
\begin{tabular}{l@{~~}l|cc|cc|c|c}
\toprule
& & \multicolumn{2}{c}{\textbf{PF-Pascal}} & \multicolumn{2}{c}{\textbf{PF-Willow}} & \multicolumn{1}{c}{\textbf{Spair-71k}}   & \multicolumn{1}{c}{\textbf{TSS}}  \\
& &  \multicolumn{2}{c}{$\alpha_{img}$} & \multicolumn{2}{c}{$\alpha_{bbox-kp}$} & \multicolumn{1}{c}{$\alpha_{bbox}$} & \multicolumn{1}{c}{$\alpha_{img}$} \\
 
& Methods  & $0.05$ &  $0.10$  &  $0.05$ &  $0.10$  & $0.10$ & $0.05$\\ \midrule
I & SF-Net baseline (soft-argmax)& 59.0 & 84.0 & 46.3 & 74.0 & 24.0 & 75.8 \\
II & SF-Net baseline (argmax) & 60.3 & 81.3 & 43.7 & 71.0 & 26.9 & 74.1 \\

III & Vis-PW-bipath + PWarp-sup & 63.0 & 84.9 & 47.0 & 76.9 & 30.7 & 83.5 \\ 

IV & (III) + PNeg~\eqref{eq:neg-loss} (\textbf{PWarpC-SF-Net}) & \textbf{65.6} & \textbf{87.9} & \textbf{47.3} & \textbf{78.2} & \textbf{33.8} & \textbf{84.1} \\
V & (III) + Max-score~\cite{Rocco2018b} &  63.7 & 81.2 & 44.6 & 71.6 & 31.8 & 77.3 \\ 
VI & (III) + Min-entropy~\cite{MinLPC20} & 59.4 & 76.7 & 41.8 & 67.9 & 28.8 & 73.2  \\ %

\bottomrule
\end{tabular}%
}\vspace{-2mm}
\caption{Comparison of different losses applied on negative image pairs, \ie depicting different object classes, when associated with our introduced PW-bipath and PWarp-supervised losses on positive image pairs. We use SF-Net as baseline network. 
The evaluation results are computed using the annotations at original resolution. }

\label{tab:neg-loss-analysis}
\end{table}

\section{PWarpC-NC-Net and PWarpc-NC-Net*}
\label{sec-sup:ncnet}

In this section, we first provide details about the NC-Net architecture. We also briefly review the training strategy of the original work. We then extensively explain our training approach and the corresponding implementation details, for both our weakly and strongly-supervised approaches, PWarpC-NC-Net and PWarpC-NC-Net* respectively. 
Finally, we extensively ablate  our approach for this architecture.

\subsection{Details about NC-Net}

\parsection{Architecture} In~\cite{Rocco2018b}, Rocco \etal introduce a learnable consensus network, applied on the 4D cost volume constructed between a pair of feature maps. Specifically, they process the cost volume with multiple 4D convolutional layers, to establish a strong locality prior on the relationships between the matches. The cost volume before and after applying the 4D convolutions is also processed with a soft mutual nearest neighbor filtering.

\parsection{Training strategy in original work} The baseline NC-Net is trained with a weakly-supervised strategy, using image-level class labels as only supervision.  Their proposed objective maximizes the mean matching scores over all hard assigned matches from the predicted cost volume constructed between images pairs of the same class, while minimizing the same quantity for image pairs of different classes. By retraining the  NC-Net architecture with this strategy, we nevertheless found the training process to be quite unstable, multiple training runs leading to substantially different performance. 

\subsection{PWarpC-NC-Net and PWarpC-NC-Net*: our training strategy}

\parsection{Warps $W$ sampling} For the weakly-supervised version, we resize the image pairs $(I, J)$ to $s_r \times s_r = 430 \times 430$, sample a dense mapping $M_W$ of the same dimension and create $I'$. Each of the images of the resulting image triplet $(I, I', J)$ is then centrally cropped to $s \times s = 400 \times 400$.  

For the strongly-supervised version, we apply the transformations on images of the same size than the crop, \ie $s_r \times s_r = s \times = 400 \times 400$. This is to avoid cropping keypoint annotations.  

As for the random mapping $M_W$, we apply 30\% of horizontal flipping. We found increasing the percentage of horizontal flipping for our PWarpC-NC-Net and PWarpC-NC-Net* networks to be beneficial compared to the other networks, in order to help stabilize the learning.

\parsection{Weighting and details on the losses } For all losses, we use a smooth representation for the known probabilistic mapping $P_W$ (see Sec.~\ref{sec-sup:impl}). 

In general, we found the PWarp-supervision objective~\eqref{eq:ce-warpsup} to be slightly harmful for the PWarpC-NC-Net networks, and therefore did not include it in our final weakly and strongly-supervised formulations. This is particularly the case when finetuning the features, which is the setting we used for our final PWarpC-NC-Net and PWarpC-NC-Net*. This is likely due to the network 'overfitting' to the synthetic image pairs and transformations involved in the PWarp-supervision loss, at the expense of the real images considered in the PW-bipath~\eqref{eq:vis-w-bipath} and PNeg~\eqref{eq:neg-loss} objectives. 

As a result, for the weakly-supervision version PWarpC-NC-Net, the weights in~\eqref{eq:weak-loss} are set to $\lambda_{\text{P-warp-sup}} = 0$ and $\lambda_{\text{PNeg}}=1$. 
For the strongly-supervised version, PWarpC-NC-Net*, we use the same weight $\lambda_{\text{P-warp-sup}} = 0$. We additionally set $\lambda_{\text{kp}} = ( L_{\text{P-warp-sup}} + L_{\text{vis-PW-bi}}  )/ L_{\text{kp}}$, which ensure that our probabilistic losses amount for the same than the keypoint loss $L_{\text{kp}}$. 
Moreover, the keypoint loss $L_{\text{kp}}$ is set as the cross-entropy loss, for both PWarpC-NC-Net* and its baseline NC-Net*.

\begin{table}[b]
\centering
\resizebox{0.48\textwidth}{!}{%
\begin{tabular}{ll|cc|cc|c|c}
\toprule
& & \multicolumn{2}{c}{\textbf{PF-Pascal}} & \multicolumn{2}{c}{\textbf{PF-Willow}} & \multicolumn{1}{c}{\textbf{Spair-71k}}   & \multicolumn{1}{c}{\textbf{TSS}}  \\
& &  \multicolumn{2}{c}{$\alpha_{img}$} & \multicolumn{2}{c}{$\alpha_{bbox-kp}$} & \multicolumn{1}{c}{$\alpha_{bbox}$} & \multicolumn{1}{c}{$\alpha_{img}$} \\
 
& Methods  & $0.05$ &  $0.10$  &  $0.05$ &  $0.10$  & $0.10$ & $0.05$\\ 
\cdashlinelr{1-8}
I & NCNet baseline (Max-score)~\cite{Rocco2018b} & 60.5 & 82.3 & 44.0 & 72.7 & 28.8 &  77.7\\
II & PW-bipath &  diverged \\
III & + Visibility mask &  \textbf{64.7} & 83.8 & \textbf{45.4} & 75.9 &  32.8 & 82.7 \\
IV & + PWarp-supervision &  61.7 & 79.2 & 45.1 & 73.8 & 35.6 & 85.4 \\
\cdashlinelr{1-8}
III & PW-bipath Visibility mask &  \textbf{64.7} & 83.8 & \textbf{45.4} & 75.9 &  32.8 & 82.7 \\
V & + PNeg & 62.0& 82.2 & \textbf{45.4} & \textbf{76.2} & 33.2 & 87.9 \\ 
VI & + ft features (\textbf{PWarpC-NC-Net}) & 64.2 & \textbf{84.4} & 45.0 & 75.9 & 35.3 & \textbf{89.2} \\
\cdashlinelr{1-8}
VII & ft features from scratch  & 63.7 & 82.9 & 44.9 & 76.1 & \textbf{35.7} & 87.4 \\ 

\midrule
VI & \textbf{PWarpC-NC-Net} & \textbf{64.2} & \textbf{84.4} & \textbf{45.0} & \textbf{75.9} & \textbf{35.3} & \textbf{89.2} \\
I & Max-score (NC-Net baseline) & 60.5 & 82.3 & 44.0 & 72.7 & 28.8 & 77.7 \\
VIII & Min-entropy~\cite{MinLPC20} & 55.6 & 79.2 & 42.0 & 72.3 & 25.4 &  78.4 \\
IX & Warp Consistency~\cite{warpc} & 59.1 & 75.0 & 44.6 & 70.1 & 35.0 & 87.0  \\
\midrule

III & PW-bipath Visibility mask &  64.7 & \textbf{83.8} & \textbf{45.4} & 75.9 &  32.8 & 82.7 \\
V & (III) + PNeg (Ours) & 62.0& 82.2 & \textbf{45.4} & \textbf{76.2} & \textbf{33.2} & \textbf{87.9} \\ 
X & (III) + Max-score & \textbf{62.9} & 82.1 & \textbf{45.4} & 74.2 & 31.3 & 79.0 \\
XI & (III) + Min-entropy & 60.8 & 78.5 & 44.8 & 71.4 & 31.5 & 78.6 \\
\bottomrule
\end{tabular}%
}\vspace{-2mm}
\caption{In the top part, we conduct an ablation study for PWarpC-NC-Net. There, we incrementally add each component. In the middle part, we then compare our Probabilistic Warp Consistency objective to alternative weakly-supervised losses. In the bottom part, we compare the impact of combining different losses on non-matching pairs with our PW-bipath objective, applied on image pairs of the same class. We measure the PCK on the PF-Pascal~\cite{PFPascal}, PF-Willow~\cite{PFWillow}, SPair-71K~\cite{spair} and TSS~\cite{Taniai2016} datasets. The evaluation results are computed using ground-truth annotations at original resolution. }
\label{tab:abla-ncnet}
\end{table}

\parsection{Implementation details} 
For PWarpC-NC-Net, we set the initial learnable parameter $z$, corresponding to the unmatched state $\o$ for our occlusion modeling at $z = 10$. This is to ensure that it is in the same range than the cost volume, at initialization. 

The SoftMax temperature, corresponding to equation~\eqref{eq:softmax} of the main paper, is set to $\tau= 1.0$, the same than originally used in the baseline loss. 
The hyper-parameter used in the estimation of our visibility mask $\widehat{V}$ (eq.~\ref{eq:vis-w-bipath} of the main paper) is set to $\gamma = 0.2$. Indeed, for NC-Net, we found that using a more restrictive threshold, as compared to the other networks which use $\gamma=0.7$ (when training on PF-Pascal), is beneficial to stabilize the training. It offers a better guarantee that the PW-bipath loss~\eqref{eq:vis-w-bipath} is \emph{only} applied in common visible object regions between the triplet. 

Similarly to baseline NC-Net~\cite{Rocco2018b}, we train in two stages. In the first stage, we only train the consensus neighborhood network while keeping the  ResNet-101 feature backbone extractor fixed. We further finetune the last layer of the feature backbone as well as the consensus neighborhood network in a second stage. 
These two stages are used to train on PF-Pascal~\cite{PFPascal}, our final PWarpC-NC-Net and PWarpC-NC-Net* approaches, as well as strongly-supervised baseline NC-Net*. 

For training, we use similar training parameters as in baseline NC-Net. We train with a batch size of 16, which is reduced to 8 when the last layer of the backbone feature is finetuned. During the first training stage on PF-Pascal, we train for a maximum of 30 epochs with a learning rate set to a constant of $5 \cdot 10^{-4}$. During the second training stage on PF-Pascal, the learning rate is reduced to $1 \cdot 10^{-4}$ and the network trained for an additional 30 epochs. 

We optionally further finetune the networks on SPair-71K~\cite{spair} for 10 epochs, with the same learning rate equal to $1 \cdot 10^{-4}$. Note that in this setting, the last layer of the feature backbone is also finetuned. 
The networks are trained using Adam optimizer~\cite{adam} with weight decay set to zero.

\subsection{PWarpC-NC-Net: ablation study and comparison to previous works}

Similarly to Sec.~\ref{sec:method-analysis} of the main paper for PWarpC-SF-Net, we here provide a detailed analysis of our weakly-supervised approach PWarpC-NC-Net. 

\parsection{Ablation study} In the top part of Tab.~\ref{tab:abla-ncnet}, we analyze key components of our weakly-supervised approach. The version denoted as (II) is trained using our PW-bipath objective~\eqref{eq:ce}, without the visibility mask. NC-Net trained with this loss diverged. With the NC-Net architecture, we found it crucial to extend our loss with our visibility mask~\eqref{eq:vis-w-bipath}, resulting in version (III). We believe applying our PW-bipath loss on all pixels (II) confuses the NC-Net network, by enforcing matching even in \eg non-matching background regions.  Note that version (III) trained with our visibility aware PW-bipath objective~\eqref{eq:vis-w-bipath} already outperforms the baseline (I) on all datasets and for all thresholds. 
Further adding the PWarp-supervision loss~\eqref{eq:ce-warpsup} in (IV) leads to worse results than (III) on the PF-Pascal and PF-Willow datasets, despite bringing an improvement on SPair-71K and TSS. To obtain a final network achieving competitive results on all four datasets, we therefore do not include the PWarp-supervision objective~\eqref{eq:ce-warpsup} in our final formulation. 

From (III), including our occlusion modeling, \ie the unmatched state and its corresponding probabilistic negative loss~\eqref{eq:neg-loss} in (V) leads to notable gains on the PF-Willow, SPair-71K and TSS datasets. In (VI), we further finetune the last layer of the feature backbone with the neighborhood consensus network in a second training stage. It leads to substantial improvements on all datasets, except for PF-Willow, where results remain almost unchanged. 

From (VI) to (VII), we compare finetuning the feature backbone in a second training stage (VI), or directly in a single training stage (VII). The former leads to better performance on the PF-Pascal dataset. As a result, version (VI) corresponds to our final weakly-supervised PWarpC-NC-Net, trained with two stages on PF-Pascal.

\parsection{Comparison to other losses} In the middle part of Tab.~\ref{tab:abla-ncnet}, we compare our Probabilistic Warp Consistency approach to previous weakly-supervised alternatives. The baseline NC-Net, corresponding to version (I), is trained with maximizing the max scores of the predicted cost volumes for matching images. It leads to significantly worse results than our approach (VI) on all datasets and threshold. The same conclusions apply to version (VIII), trained with minimizing the cost volume entropy for matching images. Finally, we compare our probabilistic approach (VI) to the mapping-based Warp Consistency method, corresponding to (IX). While Warp Consistency (IX) achieves good performance on the SPair-71K and TSS datasets, it leads to poor results on the PF-Pascal and PF-Willow datasets.

\parsection{Comparison of objectives on negative image pairs} Finally, in the bottom part of Tab.~\ref{tab:abla-ncnet}, we compare multiple alternative losses applied on image pairs depicting different object classes. In particular, we combine our visibility-aware PW-bipath loss (III) with either our introduced probabilistic negative loss~\eqref{eq:neg-loss}, minimizing the maximum scores~\cite{Rocco2018b} or maximizing the cost volume entropy~\cite{MinLPC20} in respectively (V), (X) and (XI). Our probabilistic negative loss~\eqref{eq:neg-loss} leads to significantly better results on the PF-Willow, SPair-71K and TSS datasets. We believe it is because it enables to explicitly model occlusions and unmatched regions through our extended probabilistic formulation, including the unmatched state.

\section{PWarpC-CATs}
\label{sec-sup:cats}

In this section, we first briefly review the CATs architecture and the original training strategy. We then provide details about the integration of our probabilistic approach into this architecture. Finally, we analyse the key components of our resulting strongly-supervised networks PWarpC-CATs and PWarpC-CATs-ft-features.

\subsection{Details about CATs}

\parsection{Architecture} CATs~\cite{CATs} finds matches which are globally consistent by leveraging a Transformer architecture applied to slices of correlation maps constructed from multi-level features. The Transformer module alternates self-attention layers across points of the same correlation map, with inter-correlation self-attention across multi-level dimensions.

\parsection{Training strategy in original work} While the final output of the CATs architecture is a cost volume, the latter is converted to a dense mapping by transposing into a probabilistic mapping with SoftMax, and then applying soft-argmax. The network is then trained with the End-Point Error objective, by leveraging the keypoint match annotations. 

\subsection{PWarpC-CATs: our training strategy}

\parsection{Warps $W$ sampling} We apply the transformations on images with dimensions $s_r \times s_r = s \times s = 256 \times 256$. We do not further crop central images to avoid cropping keypoint annotations.  

When training on PF-Pascal, we apply $5\%$ of horizontal flipping to sample the random mappings $M_W$, while it is increased to $15 \%$ when training on SPair-71K. 

\parsection{Weighting and details on the losses } We define the known probabilistic mapping $P_W$ with a one-hot representation for our PW-bipath and PWarp-supervision losses~\eqref{eq:vis-w-bipath}-\eqref{eq:ce-warpsup} (see Sec.~\ref{sec-sup:impl}). 

To obtain PWarpC-CATs, we set the weights in~\eqref{eq:strong-loss} as $\lambda_{\text{P-warp-sup}} = L_{\text{vis-PW-bi}} / L_{\text{P-warp-sup}}$ and $\lambda_{\text{kp}} = ( L_{\text{P-warp-sup}} + L_{\text{vis-PW-bi}}  )/ L_{\text{kp}}$, which ensure that our probabilistic losses amount for the same than the keypoint loss $L_{\text{kp}}$. 

To obtain PWarpC-CATs-ft-features, where the ResNet-101 backbone feature is additionally finetuned, we found the PWarp-supervision objective~\eqref{eq:ce-warpsup} to be slightly harmful, and therefore did not include it in this case. This is consistent with the findings of PWarpC-NC-Net and PWarpC-NC-Net*, for which the PWarp-supervised objective was also found harmful when the feature backbone is finetuned. This is likely due to the network 'overfitting' to the synthetic image pairs and transformations involved in the PWarp-supervision loss, at the expense of the real images considered in the PW-bipath~\eqref{eq:vis-w-bipath} objectives. 
As a result, for the PWarpC-CATs-ft-features version, we set the weights in~\eqref{eq:strong-loss} as $\lambda_{\text{P-warp-sup}} = 0$ and $\lambda_{\text{kp}} = ( L_{\text{P-warp-sup}} + L_{\text{vis-PW-bi}}  )/ L_{\text{kp}}$. 

Moreover, to be consistent with the baseline CATs, the keypoint loss $L_{\text{kp}}$ is set as End-Point-Error loss, after converting the probabilistic mapping to a mapping through soft-argmax.

\parsection{Implementation details} 
The softmax temperature, corresponding to equation~\eqref{eq:softmax} of the main paper, is set to $\tau= 0.02$, the same than originally used in the baseline. 
The hyper-parameter used in the estimation of our visibility mask $\widehat{V}$ (eq.~\ref{eq:vis-w-bipath} of the main paper) is set to $\gamma = 0.7$ and to $\gamma = 0.2$ when trained on PF-Pascal or SPair-71K respectively. This is because in SPair-71K, the objects are generally much smaller than in PF-Pascal. 

For training, we use similar training parameters as in baseline CATs. We train with a batch size of 16 when the feature backbone is frozen, and  reduce it to 7 when finetuning the backbone. The initial learning rate is set to $3 \cdot 10^{-6}$ for the feature backbone, and $3 \cdot 10^{-5}$ for the rest of the architecture. It is halved after 80, 100 and 120 epochs and we train for a maximum of 150 epochs. We use the same training parameters when training on either PF-Pascal or SPair-71K. 
The networks are trained using AdamW optimizer~\cite{AdamW} with weight decay set to $0.05$.

\begin{table}[t]
\centering
\resizebox{0.48\textwidth}{!}{%
\begin{tabular}{l|cc|cc|c|c}
\toprule
& \multicolumn{2}{c}{\textbf{PF-Pascal}} & \multicolumn{2}{c}{\textbf{PF-Willow}} & \multicolumn{1}{c}{\textbf{Spair-71k}}   & \multicolumn{1}{c}{\textbf{TSS}}  \\
&  \multicolumn{2}{c}{$\alpha_{img}$} & \multicolumn{2}{c}{$\alpha_{bbox-kp}$} & \multicolumn{1}{c}{$\alpha_{bbox}$} & \multicolumn{1}{c}{$\alpha_{img}$} \\
 
Methods  & $0.05$ &  $0.10$  &  $0.05$ &  $0.10$  & $0.10$ & $0.05$\\ 
\toprule
CATs baseline (EPE) & 67.3 & \textbf{88.6} & 41.6 &  68.9 & 22.1 & 74.8\\
+ Vis-aware-PW-bipath & \textbf{68.1} & 88.5 & 44.0 & 70.6 & 21.4 & 76.3  \\  
+ PWarp-supervision (\textbf{PWarpC-CATs}) & 67.1 & 88.5 & \textbf{44.2} & \textbf{71.2} & \textbf{23.3} & \textbf{82.4} \\

\cdashlinelr{1-7}
CATs-ft-features (EPE) & \textbf{79.8} & 92.7 & 45.2 & 73.2 & 26.8 &  78.4 \\

\begin{tabular}{c}  \hspace{-15mm}+ Vis-aware-PW-bipath  \\ (\textbf{PWarpC-CATs-ft-features})\end{tabular}
  & \textbf{79.8} & 92.6 & \textbf{48.1} & \textbf{75.1} & \textbf{27.9} & \textbf{88.7} \\
+ PWarp-supervision &  79.6 & 92.4 & 46.7 & 74.4 & 26.0  & \textbf{88.7} \\
\bottomrule
\end{tabular}%
}\vspace{-3mm}
\caption{Ablation study for PWarpC-CATs and PWarpC-CATs-ft-features. We incrementally add each component. We measure the PCK on the PF-Pascal~\cite{PFPascal}, PF-Willow~\cite{PFWillow}, SPair-71K~\cite{spair} and TSS~\cite{Taniai2016} datasets. The evaluation results are computed using ground-truth annotations at original resolution. }
\label{tab:abla-cats}
\end{table}

\subsection{Ablation study}

In Tab.~\ref{tab:abla-cats}, we analyse the key components of our strongly-supervised approaches PWarpC-CATs (top part) and PWarpC-CATs-ft-features (bottom part). 
From the CATs baseline, which is trained with the End-Point Error (EPE) objective while keeping the backbone feature frozen, adding our visibility-aware PW-bipath loss~\eqref{eq:vis-w-bipath} leads to a subtantial gain on the PF-Willow and TSS dataset. Further including our PWarp-supervision objective results in improved performance on PF-Willow, SPair-71K and TSS. 
For the versions with finetuning the feature backbone (bottom part of  Tab.~\ref{tab:abla-cats}), our visibility-aware PW-bipath objective brings major gains on PF-Willow, SPair-71K and TSS. However, further adding the PWarp-supervision leads to a small drop in performance on all datasets. For this reason, we use the combination of the EPE loss with our visibility-aware PW-bipath objective to train our final PWarpC-CATs-ft-features.

\section{PWarpC-DHPF}
\label{sec-sup:dhpf}

As in previous sections, we first review the DHPF~\cite{MinLPC20} architecture and its original training strategy. We then provide training details for our strongly-supervised PWarpC-DHPF. Finally, we provide an ablation study for our approach applied to this architecture.

\subsection{Details about DHPF} 

\parsection{Architecture} DHPF learns to compose hypercolumn features, \ie aggregation of different layers, on the fly by selecting a small number of relevant layers from a deep convolutional neural network. In particular, it proposes a gating mechanism to choose which layers to include in the hypercolumn. 
The hypercolumns features are then correlated, leading to the final output cost volume.

\parsection{Training strategy in original work} The original work proposes both a weakly and strongly-supervised approach. The weakly-supervised approach is trained with minimizing the cost volume entropy computed between image pairs depicting the same class, while maximizing it for pairs depicting a different semantic content.

The strongly-supervised approach is instead trained with the cross-entropy loss, after converting the keypoint match annotations to probability distributions. In both cases, the authors also include a layer selection loss. It is a soft constraint to encourage the network to select each layer of the feature backbone at a certain rate. 

\subsection{PWarpC-DHPF: our training strategy}

\parsection{Warps $W$ sampling} We apply the transformations on images with dimensions $s_r \times s_r = s \times s = 240 \times 240$. Similarly to PWarpC-CATs, we do not further crop central images to avoid cropping keypoint annotations.  

When training on PF-Pascal, we apply $5\%$ of horizontal flipping to sample the random mappings $M_W$, while it is increased to $15 \%$ when training on SPair-71K. 

\parsection{Weighting and details on the losses } We define the known probabilistic mapping $P_W$ with a smooth representation for our PW-bipath and PWarp-supervision losses~\eqref{eq:vis-w-bipath}-\eqref{eq:ce-warpsup} (see Sec.~\ref{sec-sup:impl}). 

To obtain PWarpC-DHPF, we set the weights in~\eqref{eq:strong-loss} as $\lambda_{\text{P-warp-sup}} = L_{\text{vis-PW-bi}} / L_{\text{P-warp-sup}}$ and $\lambda_{\text{kp}} = ( L_{\text{P-warp-sup}} + L_{\text{vis-PW-bi}}  )/ L_{\text{kp}}$, which ensure that our probabilistic losses amount for the same than the keypoint loss $L_{\text{kp}}$. 

Moreover, in the strongly-supervised baseline DHPF, they train with a keypoint loss $L_{\text{kp}}$ corresponding to the cross-entropy with the ground-truth keypoint matches converted to one-hot probabilistic mapping representations. We nevertheless found that the baseline is slightly improved when the ground-truth keypoint matches are instead converted to smooth probability distributions. We denote this version as DHPF* and compare it to our final PWarpC-DHPF in Tab.~\ref{tab:abla-dhpf}. As a result, for our PWarpC-DHPF, we set the keypoint loss $L_{\text{kp}}$ in~\eqref{eq:strong-loss} to the cross-entropy with a smooth representation of the ground-truth keypoint match distributions. 
Finally, for fair comparison, we add the layer selection loss used in baseline DHPF to our strongly-supervised loss~\eqref{eq:strong-loss}.

\parsection{Implementation details} 
The softmax temperature, corresponding to equation~\eqref{eq:softmax} of the main paper, is set to $\tau= 1$, as in the baseline loss. Note that following the baseline DHPF, we apply gaussian normalization on the cost volume before applying the SoftMax operation~\eqref{eq:softmax} to convert it to a probabilistic mapping.
The hyper-parameter used in the estimation of our visibility mask $\widehat{V}$ (eq.~\ref{eq:vis-w-bipath} of the main paper) is set to $\gamma = 0.7$ and to $\gamma = 0.2$ when trained on PF-Pascal or SPair-71K respectively. This is because in SPair-71K, the objects are generally much smaller than in PF-Pascal. 

For training, we use similar training parameters as in baseline DHPF. We train on PF-Pascal with a batch size of 6 for a maximum of 100 epochs. The initial learning rate is set $3 \cdot 10^{-2}$ and halved after 50 epochs.  
We optionally further finetune the network on SPair-71K, with an additional 10 epochs and a constant learning rate of $1 \cdot 10^{-2}$. 
The networks are trained using SGD optimizer~\cite{ruder2016overview}.

\begin{table}[t]
\centering
\resizebox{0.47\textwidth}{!}{%
\begin{tabular}{l|cc|cc|c|c}
\toprule
& \multicolumn{2}{c}{\textbf{PF-Pascal}} & \multicolumn{2}{c}{\textbf{PF-Willow}} & \multicolumn{1}{c}{\textbf{Spair-71k}}   & \multicolumn{1}{c}{\textbf{TSS}}  \\
&  \multicolumn{2}{c}{$\alpha_{img}$} & \multicolumn{2}{c}{$\alpha_{bbox-kp}$} & \multicolumn{1}{c}{$\alpha_{bbox}$} & \multicolumn{1}{c}{$\alpha_{img}$} \\
 
Methods  & $0.05$ &  $0.10$  &  $0.05$ &  $0.10$  & $0.10$ & $0.05$\\ 
\cdashlinelr{1-7}
DHPF baseline (CE with one-hot) & 77.3 & \textbf{91.7} & 44.8 & 70.6 &  27.5 & 72.2   \\
DHPF* (CE with smooth) & \textbf{78.1} & 90.7 & 44.7 & 70.1 &27.9 & 74.02\\
+ Vis-aware-PW-bipath &  76.3 & 90.7 & 47.3 & 73.6 & 28.0 & 73.7 \\
+ PWarp-supervision (\textbf{PWarpC-DHPF}) &  77.7 & \textbf{91.7} & \textbf{47.7} & \textbf{74.3} &  \textbf{28.6} & \textbf{74.3} \\
\bottomrule
\end{tabular}%
}\vspace{-3mm}
\caption{Ablation study for PWarpC-DHPF. We incrementally add each component. We measure the PCK on the PF-Pascal~\cite{PFPascal}, PF-Willow~\cite{PFWillow}, SPair-71K~\cite{spair} and TSS~\cite{Taniai2016} datasets. The evaluation results are computed using ground-truth annotations at original resolution. }
\vspace{-3mm}
\label{tab:abla-dhpf}
\end{table}

\subsection{Ablation study} In Tab.~\ref{tab:abla-dhpf}, we conduct ablative experiments on PWarpC-DHPF. Training with the cross-entropy loss using a smooth representation of the ground-truth in DHPF* leads to slightly better results than DHPF on PF-Pascal and SPair-71K. For this reason, we use it as baseline. Further including our visibility-aware PW-bipath loss and PWarp-supervision leads to incremental gains on PF-Willow and SPair-71K.

\section{Analysis of transformations W}
\label{sec-sup:transfo-analysis}
 
In this section, we analyse the impact of the sampled transformations' strength on the performance of the corresponding trained PWarpC networks.
As explained in Sec.~\ref{sec-sup:transfo}, the strength of the warps $M_W$ is mostly controlled by the range $\sigma_H$, used to sample the base homography, TPS and Affine-TPS transformations. The probability of horizontal flipping $p_{flip}$ also has a large impact. 
We thus analyse the effect of the sampling range $\sigma_H$ and the probability of horizontal flipping $p_{flip}$ on the evaluation results of the corresponding PWarpC networks. In particular, we provide the analysis for our weakly-supervised PWarpC-SF-Net. The trend is the same for the other PWarpC networks.

\begin{figure*}[t]
\centering%
(A) Impact of sampling range \\
\subfloat[PF-Pascal]{\includegraphics[width=0.48\textwidth]{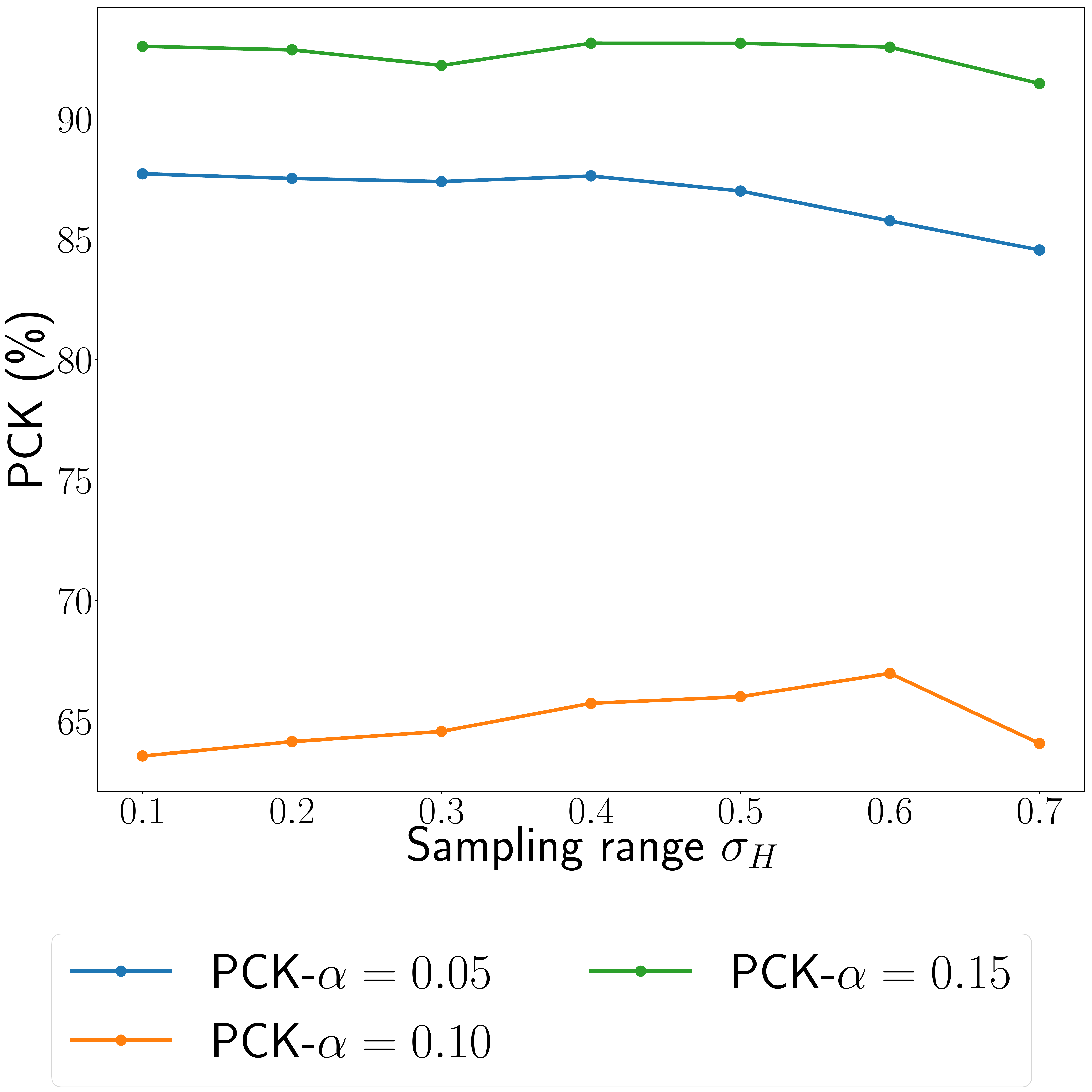}} \hspace{1mm}
\subfloat[SPair-71K]{\includegraphics[width=0.48\textwidth]{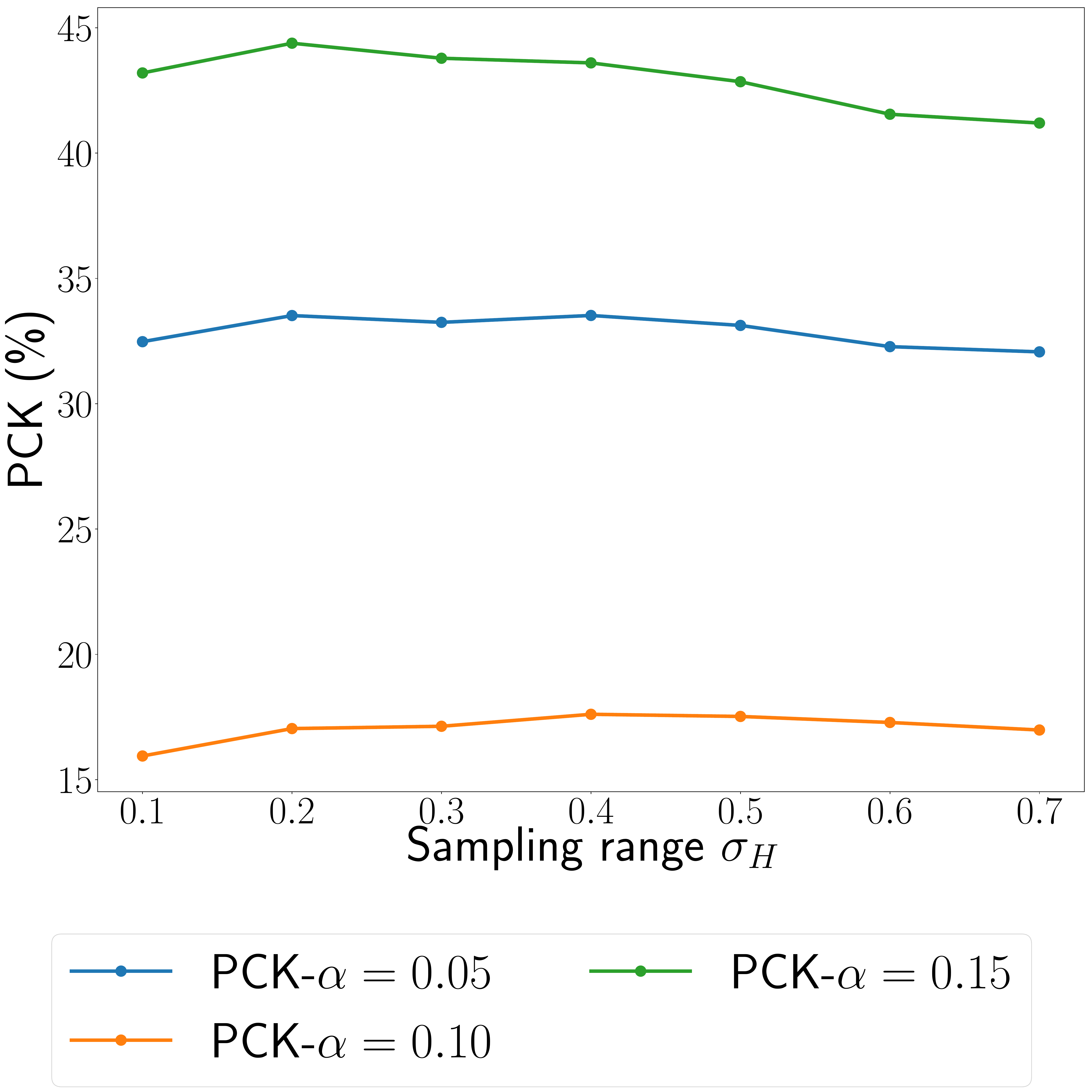}} \\
\vspace{7mm}
(B) Impact of probability of horizontal flipping \\
\subfloat[PF-Pascal]{\includegraphics[width=0.48\textwidth]{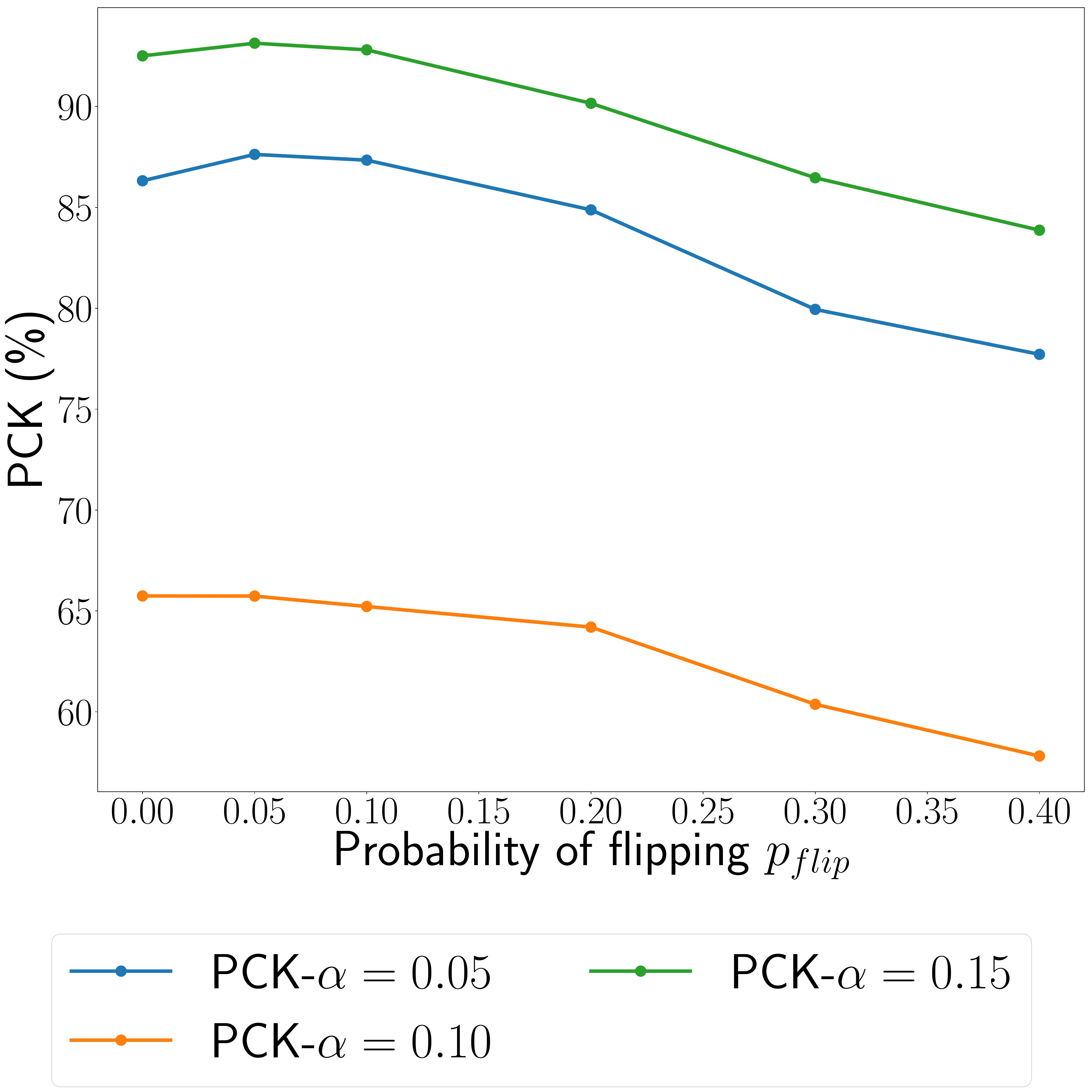}} \hspace{1mm}
\subfloat[SPair-71K]{\includegraphics[width=0.48\textwidth]{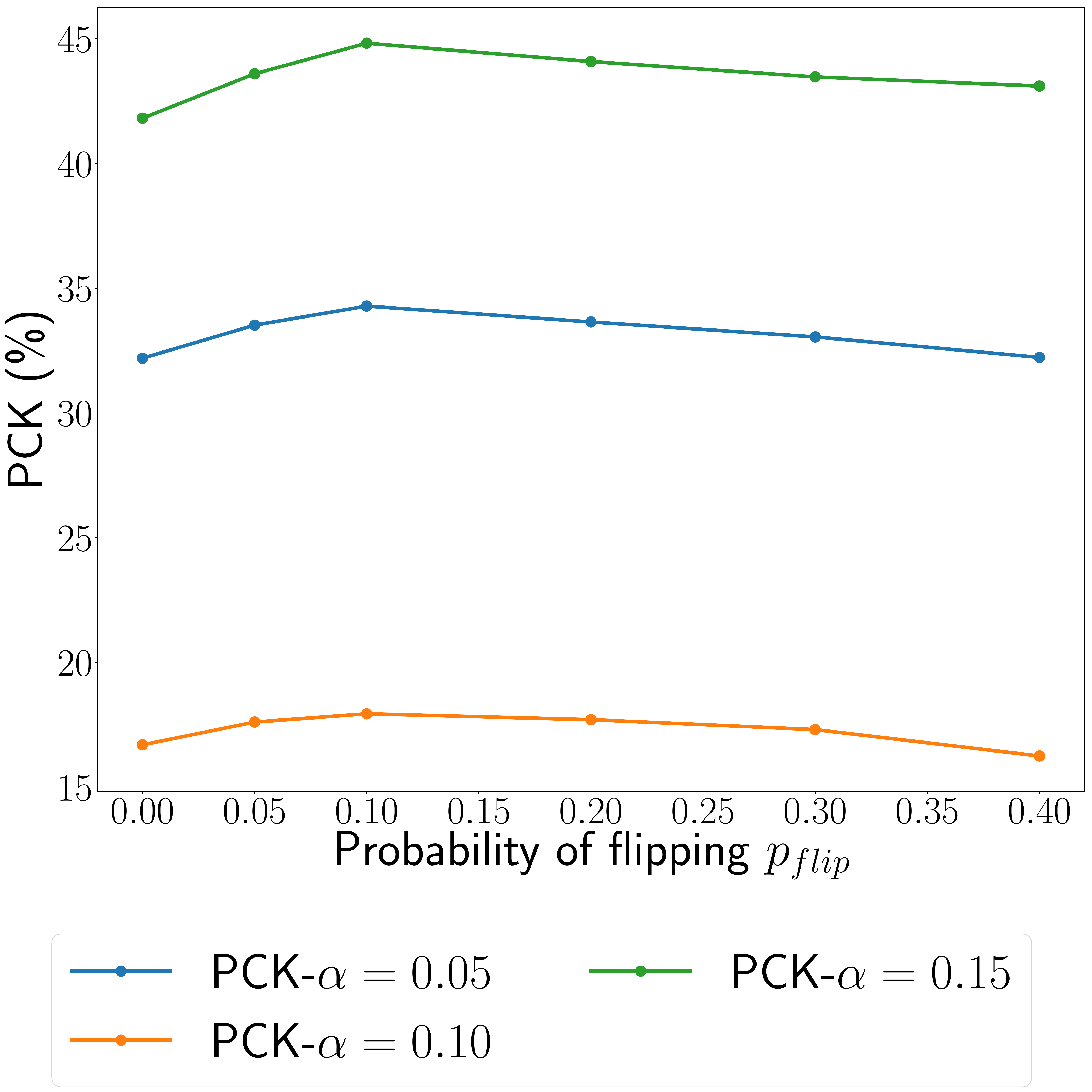}} \\

\caption{Impact of the strength of the transformations $M_W$, on the performance of the weakly-supervised PWarpC-SF-Net network. We look at the PCK for $\alpha$ thresholds in $\left \{0.05, 0.1, 0.15 \right\}$ obtained on the PF-Pascal~\cite{PFPascal} and SPair-71K~\cite{spair} datasets, for different sampling ranges $\sigma_H$ and probability of horizontal flipping $p_{flip}$, used to create the synthetic transformations $M_W$ during training.
}\vspace{-4mm}
\label{fig:transfo-analysis}
\end{figure*}

While we choose a specific distribution to sample the transformations parameters used to construct the mapping $M_W$, our experiments show that the performance of the trained networks according to our proposed Probabilistic Warp Consistency loss  is relatively insensitive to the strength of the transformations $M_W$, if they remain in a reasonable bound. We present these experiments in Fig.~\ref{fig:transfo-analysis}.  

In Fig.~\ref{fig:transfo-analysis} (A), we analyse the impact of the sampling range on the performance of PWarpC-SF-Net. Any range within $\left[ 0.1, 0.7 \right]$ leads to similar performance, for $\alpha=0.1$ and for $\alpha=0.15$. Only for $\alpha=0.05$ on PF-Pascal, increasing the range up to $0.6$ leads to better results, with a drop for $\sigma_H= 0.7$.  We select $\sigma_H = 0.4$ in our final setting.

We then look at the impact of the probability of horizontal flipping in Fig.~\ref{fig:transfo-analysis} (B). On PF-Pascal, increasing the probability of flipping up to $5 \%$ leads to an increase in performance. Increasing it further nevertheless results in a gradual drop in performance. The trend is the same on SPair-71K, except that the best results are achieved for $p_{flip}=10\%$. We therefore set $p_{flip}=5\%$ for our final PWarpC networks.

\begin{table*}[t]
\centering
\resizebox{0.98\textwidth}{!}{%
\begin{tabular}{l@{~~}ll|ccc|ccc|cc|c@{~~}c@{~~}c@{~~}c}
\toprule
 & &  & \multicolumn{3}{c}{\textbf{PF-Pascal}} & \multicolumn{3}{c}{\textbf{PF-Willow}} & \multicolumn{2}{c}{\textbf{Spair-71k}} & \multicolumn{4}{c}{\textbf{TSS}}  \\
 & &   & \multicolumn{3}{c}{PCK @ $\alpha_{img}$} & \multicolumn{3}{c}{PCK @ $\alpha_{bbox-kp}$}  & \multicolumn{2}{c}{PCK @ $\alpha_{bbox}$} & \multicolumn{4}{c}{PCK @ $\alpha_{img} \,\, , \alpha=0.05$}\\
 
 & Methods  & Reso & $0.05$ &  $0.10$  &  $0.15$ & $0.05$ &  $0.10$  &  $0.15$  & $0.05$ &  $0.10$ & FG3DCar & JODS & Pascal & Avg.  \\ \midrule
S & HPF$_{\text{res101}}$~\cite{HPF} & max $300$ &  - &- &- & - &- &- & - &  28.2 & - &- &- & - \\ 

& SCOT$_{\text{res101}}$~\cite{SCOT} & max 300 &   - &- &- & - &- &- & - & 35.6 & - &- &- & - \\ 
& CHM$_{\text{res101}}$~\cite{CHM} & $240$ & \lowreso{-} &\lowreso{-} &\lowreso{-} & \lowreso{-} & \lowreso{-} & \lowreso{-} & \lowreso{-}  &  \lowreso{46.3} & \lowreso{-} & \lowreso{-} & \lowreso{-} & \lowreso{-} \\
& PMD$_{\text{res101}}$~\cite{PMD}  & - &   - &- &- & - &- &- & - &   37.4  & - &- &- & -\\
& PMNC$_{\text{res101}}$~\cite{PMNC} & - & - &- &- & - &- &- & - & 50.4 & - &- &- & -  \\
& MMNet$_{\text{res101}}$~\cite{MMNet} & $224 \times 320$  &   \lowreso{-} &\lowreso{-} &\lowreso{-} & \lowreso{-} & \lowreso{-} & \lowreso{-} & \lowreso{-}  &  \lowreso{40.9}  &\lowreso{-} & \lowreso{-} & \lowreso{-} & \lowreso{-}  \\
& DHPF$_{\text{res101}}$~\cite{MinLPC20} & $240$ &   \lowreso{52.6} $^{\dagger}$ &\lowreso{75.4} $^{\dagger}$ &\lowreso{84.8} $^{\dagger}$ & \lowreso{37.4} $^{\dagger}$ & \lowreso{63.9} $^{\dagger}$ & \lowreso{77.0} $^{\dagger}$ & \lowreso{20.7} $^{\dagger}$ &  \lowreso{37.3} & \lowreso{-} & \lowreso{-} & \lowreso{-} & \lowreso{-}  \\
& CATs$_{\text{res101}}$~\cite{CATs} & $256$ &  \lowreso{45.3} $^{\dagger}$ & \lowreso{67.7} $^{\dagger}$ & \lowreso{77.0} $^{\dagger}$ & \lowreso{31.8} $^{\dagger}$ & \lowreso{56.8} $^{\dagger}$ & \lowreso{69.1} $^{\dagger}$ & \lowreso{21.9} $^{\dagger}$ &  \lowreso{42.4} & \lowreso{-} & \lowreso{-} & \lowreso{-} & \lowreso{-} \\
& CATs-ft-features$_{\text{res101}}$~\cite{CATs} & $256$ &  \lowreso{54.4} $^{\dagger}$ &\lowreso{74.1} $^{\dagger}$ &\lowreso{81.9} $^{\dagger}$ & \lowreso{39.7} $^{\dagger}$ & \lowreso{66.3} $^{\dagger}$ & \lowreso{78.3} $^{\dagger}$ & \lowreso{27.9} $^{\dagger}$  &   \lowreso{49.9} & \lowreso{-} & \lowreso{-} & \lowreso{-} & \lowreso{-}  \\

\cdashlinelr{2-15}
& CATs-ft-features$_{\text{res101}}$~\cite{CATs} & ori $^\dagger$ & 57.7 & 75.2 & 82.9 & 43.5 & 69.1 & 80.8 & 27.1 & 48.8 &  88.9 & 73.9 & 57.1 & 73.3 \\

& \textbf{PWarpC-CATs-ft-features}$_{\text{res101}}$ & ori  & 58.8 & 77.4 & 84.6 & \second{46.4} & \second{73.6} & \second{85.0} & 28.2 & 48.4 & 91.1 & \best{85.8} & \second{69.1} & \second{82.0} \\  

\cdashlinelr{2-15}

& DHPF$_{\text{res101}}$~\cite{MinLPC20} & ori $^\dagger$ &  56.9 & 77.2 & 86.3 & 40.9 & 66.8 & 79.9 & 20.6 & 36.3 & 83.8 & 69.7 & 57.3 & 70.3 \\
& \textbf{PWarpC-DHPF}$_{\text{res101}}$ &  ori &  65.8 & \second{85.5} & 92.3 & \best{47.6} & 72.9 & 84.5 & 23.3 & 38.7 &  87.5 & 73.7 & 60.3 & 73.8 \\ 
\cdashlinelr{2-15}

& NC-Net*$_{\text{res101}}$  & ori  & 59.8 & 75.6 & 82.1 & 38.9 & 62.6 &  74.7 & \second{29.1} & \second{50.7} & 81.1 & 66.7 & 45.4 &  64.4 \\  
& \textbf{PWarpC-NC-Net*}$_{\text{res101}}$ & ori  & \second{67.8} & 82.3 & 86.9 & 46.1 & 72.6 & 82.7 & \best{31.6} & \best{52.0} & \best{93.0} & \second{84.6} & \best{70.6} & \best{82.7} \\ 

\cdashlinelr{2-15}

&  SF-Net*$_{\text{res101}}$ & ori & 66.5 & 85.0 & 90.8 & 43.5 & 70.4 & 82.9 & 26.2 & 50.0 & 88.3 & 75.3 & 57.2  & 73.6 \\  

& \textbf{PWarpC-SF-Net*}$_{\text{res101}}$ & ori & \best{72.1} & \best{89.6} & \best{93.5} & 46.3 & \best{75.2} & \best{87.0} & 27.0 & 48.8 & \second{92.5} & 81.1 & 66.2 & 79.9 \\  

\midrule  \midrule
U & CNNGeo$_{\text{res101}}$~\cite{Rocco2017a}  (results from~\cite{spair})  & -  &   - &- &- & - &- &- & - &  20.6 &- & - &- &- \\
& A2Net$_{\text{res101}}$~\cite{SeoLJHC18} (results from~\cite{spair})  &  - &- &- & - &- &- & - & - & 22.3  &- & - &- &- \\ \midrule

M & SF-Net$_{\text{res101}}$~\cite{SFNet} (results from~\cite{PMNC}) & -  &    - &- &- & - &- &- & - &  26.3 &- & - &- &- \\ 
\midrule
W & \textbf{PWarpC-SF-Net}$_{\text{res101}}$ & ori & \best{64.5} & \best{86.9} & \best{92.6} & \best{47.1} & \best{78.1} & \best{89.9} & \best{18.6} & \second{37.1} & 91.0 & 81.6 & 67.4 & 80.0\\  

\cdashlinelr{2-15}

& WeakAlign$_{\text{res101}}$~\cite{Rocco2018a} (results from~\cite{spair}) &   - &  - &- &- & - &- &- & - &  20.9\\ 

& DHPF$_{\text{res101}}$~\cite{MinLPC20} & $240$ & \lowreso{46.1} $^{\dagger}$ &\lowreso{78.1} $^{\dagger}$ &\lowreso{88.4} $^{\dagger}$ & \lowreso{34.9 } $^{\dagger}$ & \lowreso{66.2} $^{\dagger}$ & \lowreso{82.5} $^{\dagger}$ & \lowreso{12.4} $^{\dagger}$ &  \lowreso{27.7} & \lowreso{-} &\lowreso{-} &\lowreso{-} & \lowreso{-} \\ 
& DHPF$_{\text{res101}}$~\cite{MinLPC20} & ori $^{\dagger}$ & 53.3 & 81.3 & 90.3 & 40.9 & 70.1 & 84.6 & 12.7 & 27.2 \\\
& PMD$_{\text{res101}}$~\cite{PMD} &  - &  - &- &- & - &- &- & - & 26.5 & - &  - &- &- \\

 & WarpC-SemGLU-Net$_{\text{vgg16}}$~\cite{warpc} & ori &  57.0 $^{\dagger}$ & 78.7 $^{\dagger}$ & 88.7 $^{\dagger}$ & \second{46.1} $^{\dagger}$ & 72.8 $^{\dagger}$ &  84.9 $^{\dagger}$ &  12.8 $^{\dagger}$ &  23.5 & \best{96.3} $^{\dagger}$ & \second{84.2} $^{\dagger}$ & \second{80.2}  $^{\dagger}$& \second{86.9}  \\ 
 
\cdashlinelr{2-15}
 & NC-Net$_{\text{res101}}$~\cite{Rocco2018b} (results from~\cite{spair}) &  - &  - &- &- & - &- &- & - &  20.1  &- & - &- &- \\ 
& \textbf{PWarpC-NC-Net}$_{\text{res101}}$  &  ori &  \second{61.7} & \second{82.6} & \second{88.5} & 43.6 & \second{74.6} & \second{86.9} & \second{18.5} & \best{38.0} & \second{95.4} & \best{88.9} & \best{85.6} & \best{90.0} \\ 
\bottomrule
\end{tabular}%
}\vspace{-2mm}
\caption{PCK [\%] obtained by different state-of-the-art methods on the PF-Pascal~\cite{PFPascal}, PF-Willow~\cite{PFWillow}, SPair-71K~\cite{spair} and TSS~\cite{Taniai2016} datasets. All approaches are trained or finetuned on the training set of Spair-71K. \textbf{S} denotes strong supervision using key-point annotation, \textbf{M} refers to using ground-truth object segmentation mask, \textbf{U} is fully unsupervised requiring only single images, and \textbf{W} refers to weakly supervised with image class labels. Each method evaluates with images and ground-truth annotations resized to a specific resolution. However, using different ground-truth resolution leads to slightly different results. We therefore use the standard setting of evaluating on the original resolution (\textbf{ori}) and gray the results computed at a different resolution. When needed, we re-compute metrics of baselines using the provided pre-trained weights, indicated by $^\dagger$. }
\label{tab:sem-spair}
\end{table*}

\begin{table*}[t]
\centering
\resizebox{0.98\textwidth}{!}{%
\begin{tabular}{ cllcccccccccccccccccc|c } 
\toprule 
 & Method & Reso & \multicolumn{19}{c}{\textbf{SPair-71K Category}} \\ \cmidrule(lr){4-22}
& & & Aero & Bike & Bird & Boat &Bottle& Bus  & Car  & Cat  & Chair& Cow  & Dog  & Horse& Motor&Person& Plant& Sheep & Train & TV & \textbf{All}\\  
\midrule 

S & CATs-ft-features ~\cite{CATs} & ori$^\dagger$ & 51.7 & 32.1 & 71.3 & 33.7 & \best{52.1} & \second{58.1} & \second{45.6} & 65.9 & 21.8 & 60.8 & \second{55.2} & 42.9 & 42.6 & 37.2 & 40.6 & 31.8 & 75.3 & 57.6 & 48.8 \\

& \textbf{PWarpC-CATs-ft-features} & ori & 49.6 & 33.4 & 72.8 & \best{37.2} & 49.0 & 53.4 & 40.6 & 66.4 & 15.7 & 56.5 & 51.0 & \second{41.3} & \second{48.0} & 40.2 & 41.4 & 36.5 & 72.7 & 61.1 & 48.8\\

\cdashlinelr{2-22}

& DHPF & ori$^\dagger$ & 39.1 & 20.4 & 65.4 & 20.3 & 44.0 & 28.2 & 21.5 & 60.5 & 18.3 & 45.6 & 45.0 & 29.9 & 26.3 & 34.4 & 25.8 & 26.3 & 54.8 & 44.5 & 36.3 \\ 

& \textbf{PWarpC-DHPF} & ori & 42.9 & 28.2 & 70.2 & 22.5 & 43.9 & 31.0 & 23.1 & 60.9 & 17.7 & 47.8 & 46.1 & 32.5 & 33.9 & 35.2 & 28.0 & 27.1 & 58.0 & 44.2 & 38.7 \\
\cdashlinelr{2-22}

& NC-Net* & ori & \second{55.2} & 33.7 & 72.7 & 35.6 & 48.5 & 54.3 & \best{47.5} & \best{71.6} & 20.8 & \best{64.8} & 54.7 & \second{46.0} & 43.5 & \best{47.5} & \second{43.0} & \second{37.3} & 74.4 & 61.3 & \second{50.7} \\

& \textbf{PWarpC-NC-Net*} & ori & 54.7 & \best{37.0} & \best{77.2} & \second{36.9} & 48.1 & 51.5 & 42.3 & \second{70.4} & \second{22.9} & \second{64.3} & \best{55.9} & \best{46.2} & \best{49.1} & \second{45.2} & \best{47.4} & \best{38.5} & 75.3 & \best{70.5} & \best{52.0}\\
\cdashlinelr{2-22}

& SF-Net* & ori & \best{57.4} & 34.2 & \second{73.1} & 34.9 & 48.4 & \best{58.4} & 45.4 & 66.6 & 20.7 & 61.3 & 53.4 & 40.4 & 44.6 & 20.1 & 40.4 & 34.2 & \best{78.7} & 66.3 & 50.0 \\
& \textbf{PWarpC-SF-Net*} & ori & 50.1 & \second{36.2} & 67.7 & 32.1 & \second{48.6} & 57.9 & 44.7 & 64.8 & \best{23.9} & 60.0 & 49.6 & 39.2 & 46.3 & 38.8 & 37.1 & 32.8 & \second{77.7} & \second{69.0} & 48.8\\

\midrule
\midrule
U & CNNGeo~\cite{Rocco2017a} & - & 23.4 & 16.7 & 40.2 & 14.3 & 36.4 & 27.7 & 26.0 & 32.7 & 12.7 & 27.4 & 22.8 & 13.7 & 20.9 & 21.0 & 17.5 & 10.2 & 30.8 & 34.1 & 20.6\\
& A2Net~\cite{SeoLJHC18}    & -   & 22.6 & 18.5 & 42.0 & 16.4 & 37.9 & \best{30.8} & \second{26.5} & 35.6 & 13.3 & 29.6 & 24.3 & 16.0 & 21.6 & 22.8 & \second{20.5} & 13.5 & 31.4 & 36.5 & 22.3\\
\midrule

M & SFNet~\cite{SFNet}     & ori $^{\dagger}$       & 26.9 & 17.2 & 45.5 & 14.7 &  38.0 & 22.2 & 16.4 & 55.3 & 13.5 & 33.4 & 27.5 & 17.7 & 20.8 & 21.1 & 16.6 & 15.6 & 32.2 & 35.9 & 26.3 \\ \midrule
W & \textbf{PWarpC-SF-Net} & ori & \second{41.4} & \second{28.9} & \second{59.0} & \second{21.1} & \best{44.9} & 28.0 & 20.0 & \best{61.8} & \best{23.0} & \best{46.0} & \second{39.0} & \second{25.6} & \second{27.9} & \second{27.8} & \best{30.2} & \second{24.1} & \best{46.4} & \best{67.5} & \second{37.1}\\ 

\cdashlinelr{2-22}
 & WeakAlign~\cite{Rocco2018a}   & -     & 22.2 & 17.6 & 41.9 & 15.1 & 38.1 & 27.4 & \best{27.2} & 31.8 & 12.8 & 26.8 & 22.6 & 14.2 & 20.0 & 22.2 & 17.9 & 10.4 & 32.2 & 35.1 & 20.9\\

& PMD    & -   & 26.2 & 18.5 & 48.6 & 15.3 & 38.0 & 21.7 & 17.3 & 51.6 & 13.7 & 34.3 & 25.4 & 18.0 & 20.0 & 24.9 & 15.7 & 16.3 & 31.4 & 38.1 & 26.5\\
\cdashlinelr{2-22}
& NCNet~\cite{Rocco2018b} & ori $^{\dagger}$   & 17.9 & 12.2 & 32.1 & 11.7 & 29.0 & 19.9 & 16.1 & 39.2 &  9.9 & 23.9 & 18.8 & 15.7 & 17.4 & 15.9 & 14.8 &  9.6 & 24.2 & 31.1 & 20.1\\

& \textbf{PWarpC-NC-Net} & ori  & \best{43.8} & \best{30.4} & \best{64.7} & \best{28.0} & \second{41.6} & \second{30.5} & 24.6 & \second{57.3} & \second{20.4} & \second{41.5} & \best{41.3} & \best{34.4} & \best{36.0} & \best{30.9} & \best{30.2} & \best{25.2} & \second{44.0} & \second{56.7} & \best{38.0} \\

\bottomrule
\end{tabular}
}
\vspace{-2mm}
\caption{Per-class PCK ($\alpha_{bbox} = 0.1$) results on SPair-71K. All models are trained on the training set of SPair-71K. They all use ResNet101 as the backbone. \textbf{S} denotes strong supervision using keypoint match annotations, \textbf{M} refers to using ground-truth object segmentation mask, \textbf{U} is fully unsupervised requiring only single images, and \textbf{W} refers to weakly-supervised with image-level class labels. Each method evaluates with ground-truth annotations resized to a specific resolution. However, using different ground-truth resolutions leads to slightly different results. We therefore use the standard setting of evaluating on the original resolution (\textbf{ori}). When needed, we re-compute metrics of baselines using the provided pre-trained weights, indicated by $^\dagger$. For each of our PWarpC networks, we compare to its corresponding baseline within the dashed-lines. Best and second-best results are in red and blue respectively. }
\label{tab:per-cat-spair-spair}
\end{table*}

\newpage
\subsection{Results when training on SPair}
\label{sup:spair}

In this section, all approaches are trained or fine-tuned on the training set of SPair-71K. In Tab.~\ref{tab:sem-spair}, we present results on PF-Pascal, PF-Willow, SPair-71K and TSS. 
In Tab.~\ref{tab:per-cat-spair-spair}, we present the detailed per-category results on SPair-71K.

\parsection{Weakly-supervised} In the bottom part of  Tab.~\ref{tab:sem-spair}, we compare approaches trained with a weakly-supervised approach. Our PWarpC-SF-Net and PWarpC-NC-Net trained on PF-Pascal were further finetuned on SPair-71K with our Probabilistic Warp Consistency objective~\eqref{eq:weak-loss}. 
Note that baselines SF-Net and NC-Net were obtained by finetuning on SPair-71K the original models trained on PF-Pascal, with their respective original training strategies. Our weakly-supervised approaches PWarpC-SF-Net and PWarpC-NC-Net lead to a particularly impressive improvement compared to their respective baselines, with $41 \%$ (+ 10.8) and $89.1 \%$ (+ 17.9) relative (and absolute) gains. As a result, PWarpC-SF-Net and PWarpC-NC-Net set a new state-of-the-art on respectively the PF-Willow and PF-Pascal datasets, and the SPair-71K and TSS datasets, across all unsupervised (U), weakly-supervised (W) and mask-supervised (M) approaches trained on SPair-71K.

\parsection{Strongly-supervised}  In the top part of~\ref{tab:sem-spair}, we report results of models trained with a strongly-supervised approach, leveraging keypoint match annotations. While training on SPair-71K with our approach leads to similar results than the baselines on SPair-71K, our PWarpC networks show drastically better generalization properties to PF-Pascal, PF-Willow and TSS. 
Our strongly-supervised PWarpC-NC-Net* sets a new state-of-the-at on SPair-71K and TSS, across all strongly-supervised approaches trained on SPair-71K. Our PWarpC-SF-Net* also obtains state-of-the-art results on the PF-Pascal and PF-Willow datasets.

\newpage

\begin{table*}[t]
\centering

\resizebox{0.95\textwidth}{!}{%
\begin{tabular}{ll@{~~}l|ccc|ccc|cccc|cccc|c}
\toprule
& Methods & Reso & \multicolumn{3}{c}{\textbf{View-point}} & \multicolumn{3}{c}{\textbf{Scale}} &   \multicolumn{4}{c}{\textbf{Truncation}} & \multicolumn{4}{c}{\textbf{Occlusion}}  \\
& &  & easy & medi & hard & easy & medi & hard & none & src & trg & both &  none & src & trg & both & All \\  \toprule
U & CNNGeo (from~\cite{spair}) & - & 25.2 & 10.7 & 5.9 & 22.3 & 16.1 & 8.5 & 21.1 & 12.7 & 15.6 & 13.9 & 20.0 & 14.9 & 14.3 & 12.4 & 18.1 \\
& A2Net (from~\cite{spair}) & - & 27.5 & 12.4 & 6.9 & 24.1 & 18.5 & 10.3 & 22.9 & 15.2 & 17.6 & 15.7 & 22.3 & 16.5 & 15.2 & 14.5 & 20.1 \\

\midrule
M & SF-Net & ori $^\dagger$ & 32.0 & 15.5 & 10.0 & 28.4 & 22.0 & 13.2 & 27.0 & 20.1 & 20.0 & 18.7 & 26.6 & 18.5 & 18.9 & 18.0 & 24.0 \\
\midrule

W & \textbf{PWarpC-SF-Net} & ori &  \second{41.9} & \second{24.2} & \second{20.7} & \second{39.1} & \second{31.8} & \second{18.8} & \second{36.3} & \second{29.7} & \second{30.4} & \best{28.4} & \second{36.5} & \second{27.7} & \second{27.9} & \second{24.7} & \second{33.5} \\
\cdashlinelr{2-18}

&  WeakAlign (from~\cite{spair}) & - & 29.4 & 12.2 & 6.9 & 25.4 & 19.4 & 10.3 & 24.1 & 16.0 & 18.5 & 15.7 & 23.4 & 16.7 & 16.7 & 14.8 & 21.1 \\
&  NC-Net (from~\cite{spair}) & - & 34.0 & 18.6 & 12.8 & 31.7 & 23.8 & 14.2 & 29.1 & 22.9 & 23.4 & 21.0 & 29.0 & 21.1 & 21.8 & 19.6 & 26.4 \\

\cdashlinelr{2-18}
&  NC-Net & ori $^\dagger$ &  37.6 & 19.4 & 13.8 & 34.7 & 26.0 & 14.9 & 31.7 & 25.2 & 25.1 & \second{23.5} & 31.5 & 23.4 & 24.3 & 20.9 & 28.8 \\

&  PWarpC-NC-Net & ori & \best{42.6} & \best{27.1} & \best{24.6} & \best{40.8} & \best{33.4} & \best{20.8} & \best{38.5} & \best{30.1} & \best{32.8} & \best{28.4} & \best{38.1} & \best{29.1} & \best{31.2} & \best{25.9} & \best{35.3} \\

\midrule

S & DHPF & ori $^\dagger$ & 34.5 & 20.0 & 15.4 & 32.4 & 25.7 & 14.7 & 31.1 & 22.5 & 22.7 & 22.1 & 30.2 & 21.8 & 22.9 & 18.7 & 27.5 \\
& \textbf{PWarpC-DHPF} & ori & 35.8 & 21.0 & 16.7 & 33.5 & 26.6 & 16.5 & 32.2 & 23.8 & 24.5 & 21.7 & 31.5 & 22.7 & 23.9 & 20.2 & 28.6 \\
\cdashlinelr{2-18}
& CATS &  ori $^\dagger$ & 29.7 & 13.8 & 9.56 & 26.4 & 20.2 & 11.6 & 25.2 & 17.8 & 18.1 & 17.3 & 24.3 & 17.2 & 18.4 & 16.3 & 22.1 \\
& \textbf{PWarpC-CATs} & ori & 30.7 & 15.1 & 11.2 & 28.2 & 21.0 & 11.6 & 26.7 & 19.1 & 18.4 & 18.0 & 25.8 & 17.8 & 19.0 & 16.6 & 23.3 \\
\cdashlinelr{2-18}
&  CATs-ft-features & ori $^\dagger$ & 35.6 & 17.0 & 12.8 &  31.5 & 24.9 & 14.7 &  29.9 & 22.6 & 22.3 & 22.4 &  29.7 & 20.5 & 21.6 & 18.5 & 26.8 \\
& \textbf{PWarpC-CATs-ft-features} &  ori & 35.6 & 19.6 & 15.7 & 33.7 &  25.3 & 14.2 &  32.0 & 22.5 & 22.9 & 21.1 &  31.2 & 21.0 & 22.1 & 19.2 & 27.9 \\
\cdashlinelr{2-18}
 & SF-Net* & ori & 36.8 & 18.6 & 12.2 & 32.8 & 25.8 & 16.0 & 30.1 & 25.4 & 25.0 & 23.7 & 30.6 & 22.7 & 23.2 & 18.4 & 27.9 \\
&  \textbf{PWarpC-SF-Net*} & ori & 41.5 & \second{22.8} & \second{18.1} & \second{38.1} & \second{30.6} & 18.2 & \second{35.6} & 28.6 & 28.9 & 26.2 & \second{35.4} & \second{26.4} & 27.4 & \second{25.3} & \second{32.5} \\

\cdashlinelr{2-18}

 & NC-Net* & ori &  \second{42.0} & 22.3 & 15.4 & 37.5 & 30.3 & \second{19.7} & 34.9 & \second{28.8} & \second{30.0} & \second{26.3} & 35.2 & 26.2 & \second{28.1} & 23.8 & 32.4  \\
 
 &  \textbf{PWarpC-NC-Net*} & ori & \best{45.4} & \best{27.7} & \best{24.7}   &  \best{42.6} & \best{35.2} & \best{22.5}  &  \best{40.3} & \best{32.5} & \best{33.7} & \best{29.7} &  \best{40.0} & \best{30.5} & \best{32.2} & \best{29.6} & \best{37.1}  \\

\bottomrule
\end{tabular}%
}\vspace{-2mm}
\caption{PCK analysis for state-of-the-art approaches, by variation factors on SPair-71K. The variation factors include view-point, scale, truncation, and occlusion with various difficulty levels. All models in this table use ResNet101 as the backbone, and are trained on the training set of PF-Pascal. \textbf{S} denotes strong supervision using keypoint match annotations, \textbf{M} refers to using ground-truth object segmentation mask, \textbf{U} is fully unsupervised requiring only single images, and \textbf{W} refers to weakly-supervised with image-level class labels. Each method evaluates with ground-truth annotations resized to a specific resolution. However, using different ground-truth resolutions leads to slightly different results. We therefore use the standard setting of evaluating on the original resolution (\textbf{ori}). When needed, we re-compute metrics of baselines using the provided pre-trained weights, indicated by $^\dagger$. For each of our PWarpC networks, we compare to its corresponding baseline within the dashed-lines. Best and second best results are in red and blue respectively. }
\label{tab:occ-state}
\end{table*}

\begin{table*}[t]
\centering
\resizebox{0.90\textwidth}{!}{%
\begin{tabular}{l@{~~}l|ccc|ccc|cccc|cccc|c}
\toprule
& Methods & \multicolumn{3}{c}{\textbf{View-point}} & \multicolumn{3}{c}{\textbf{Scale}} &   \multicolumn{4}{c}{\textbf{Truncation}} & \multicolumn{4}{c}{\textbf{Occlusion}}  \\
& & easy & medi & hard & easy & medi & hard & none & src & trg & both &  none & src & trg & both & All \\  \toprule
I & SF-Net & 32.0 & 15.5 & 10.0 & 28.4 & 22.0 & 13.2 & 27.0 & 20.1 & 20.0 & 18.7 & 26.6 & 18.5 & 18.9 & 18.0 & 24.0 \\
II & PW-bipath~\eqref{eq:ce}  & 35.0 & 20.3 & 17.4 & 33.6 & 25.9 & 14.2 & 31.7 & 23.1 &  23.4 & 21.8 &  30.7 & 21.9 & 23.8 & 21.0 & 28.0 \\
III & + visibility mask~\eqref{eq:vis-w-bipath}& 37.4 & 19.0 & 13.3 & 33.6 & 26.6 & 15.8 & 31.7 & 24.7 & 24.4 & 22.1 & 31.3 & 22.3 & 24.2 & 21.4 & 28.5 \\
IV & + PWarp-supervision~\eqref{eq:ce-warpsup} &  38.0 & 22.2 & 18.9 & 35.9 & 28.5 & 16.5 & 33.7 & 25.9 & 26.8 & 25.3 & 33.4 & 24.6 & 25.3 & 22.9 & 30.7\\
V & + PNeg~\eqref{eq:neg-loss} (\textbf{PWarpC-SF-Net}) &  \textbf{41.9} & \textbf{24.2} & \textbf{20.7} & \textbf{39.1} & \textbf{31.8} & \textbf{18.8} & \textbf{36.3} & \textbf{29.7} & \textbf{30.4} & \textbf{28.4} & \textbf{36.5} & \textbf{27.7} & \textbf{27.9} & \textbf{24.7} & \textbf{33.5} \\
\midrule

V & \textbf{PWarpC-SF-Net} (Ours) &   \textbf{41.9} & \textbf{24.2} & \textbf{20.7} & \textbf{39.1} & \textbf{31.8} & \textbf{18.8} & \textbf{36.3} & \textbf{29.7} & \textbf{30.4} & \textbf{28.4} & \textbf{36.5} & \textbf{27.7} & \textbf{27.9} & \textbf{24.7} & \textbf{33.5} \\
VI & Mapping Warp Consistency~\cite{warpc} & 34.4 & 18.2 & 14.0 & 31.7 & 24.7 & 14.1 & 30.5 & 22.3 & 21.2 & 19.3 & 29.4 & 20.9 & 21.8 & 18.3 & 26.6 \\
VII & PWarp-supervision only~\eqref{eq:ce-warpsup} &  35.3 & 21.6 & 16.6 & 33.4 &  26.9 & 16.2 & 31.1 & 24.4 & 26.5 & 23.2 & 31.4 & 23.0 & 23.3 & 21.2 & 27.9  \\

VIII & Max-score~\cite{Rocco2018b} & 34.0 & 20.8 & 15.9 & 33.0 & 25.3 & 14.2 & 30.0 & 24.7 & 24.2 & 22.4 & 30.3 & 21.6 & 22.5 & 20.4  & 24.6 \\
IX & Min-entropy~\cite{MinLPC20} &  28.3 & 17.5 & 12.3 & 27.7 & 20.4 & 11.9 & 25.3 & 19.3 & 19.9 & 19.8 & 25.4 & 18.2 & 18.1 & 15.5 & 20.6 \\

\bottomrule
\end{tabular}%
}\vspace{-2mm}
\caption{Ablation study (top part) and comparison to alternative weakly-supervised or unsupervised losses (bottom part). We compare the PCK by variation factors on SPair-71K. The variation factors include view-point, scale, truncation, and occlusion with various difficulty levels. }
\label{tab:spair-occ}
\end{table*}

\section{Detailed results when trained on PF-Pascal}
\label{sec-sup:results}

In this section, we first provide additional details on the validation datasets and experimental setting in Sec.~\ref{suc:data}. In Sec.~\ref{sup:occ}, we then analyze the robustness of our approach to different variation factors, \ie occlusion, truncation, scale and view-point, on the SPair-71K dataset. We then provide the per-category results on SPair-71K in Sec.~\ref{sup:per-cat-pfpascal}. Subsequently, we show additional prediction examples of the unmatched state for our weakly-supervised approaches in Sec.~\ref{sup:occ-pre}. We follow by analysing our approach in terms of robustness of the predicted confidence scores in Sec.~\ref{sup:conf}. 
In Sec.~\ref{sup:caltech}, we further compare state-of-the-art methods on the Caltech-101 dataset. Finally, we provide extensive qualitative comparisons in Sec~\ref{sup:qual}.

\subsection{More details on datasets and metrics}
\label{suc:data}

\parsection{Evaluation metrics} For evaluation, we adopt the standard evaluation metric, percentage of correct keypoints (PCK). Given a set of $M$ predicted and ground-truth keypoint $\left \{ \widehat{k} \right \}_{m=1}^M$ and $\left \{ k \right \}_{m=1}^M$, the PCK for the corresponding image pair is calculated as $\text{PCK} = \frac{1}{M} \sum _{m=1}^M \mathds{1} \left [   \left \| \widehat{k}_m - k_m \right \|  \leqslant \alpha_{\tau} \cdot \max(h_s^{\tau}, w_s^{\tau})  \right ]$. Here, $h_{s}$ and $w_{s}$ are either the dimensions of the source image or the dimensions of the object bounding box in the source image.

\noindent\textbf{PF-Pascal} contains 1341 image pairs from 20 categories. Images have dimensions randing from $102 \times 300$ to $300 \times 300$. We use the splits proposed in~\cite{SCNet} where training, validation and test sets respectively contain 700, 300 and 300 image pairs. In line with~\cite{SCNet}, we report the PCK with respect to the dimensions of the source image. 

\noindent\textbf{PF-Willow} comprises 900 images from 4 categories with small variations in view-point and scale, and 10 keypoint annotations per pair. Images have dimensions ranging from $153 \times 300$ to $300 \times 300$. 
Due to the absence of real bounding box annotations in PF-WILLOW, the evaluation threshold of a bounding box, $\max(w_s^{\text{bbox}} , h_s^{\text{bbox}})$, is computed using the two furthest key-point positions to approximate a bounding box that tightly wraps the object (which we refer to as bbox-kp). However, note that a few previous works sometimes use a different bounding box definition (referred to as bbox-loose), which loosely covers the object by using only a single keypoint position. Since this definition is not as accurate as the former, we do not report the results using this bounding box definition.

\noindent\textbf{SPair-71K} is a highly challenging dataset, comprising 70958 image pairs from 18 categories with extreme and diverse viewpoint and scale variations. Images have dimensions ranging from $188 \times 312$ to $500 \times 500$.  The dataset contains rich annotations for each image pair, \eg keypoints, scale difference, truncation
and occlusion difference, and a clear data split. In line with previous works, we report the PCK with respect to source bounding box dimensions. 

\noindent\textbf{TSS} is the only dataset proving dense flow field annotations for the foreground object in each pair. It contains 400 image pairs, divided into three groups: FG3DCAR, JODS, and PASCAL, according to the origins of the images. Images have dimensions ranging from $237 \times 250$ to $600 \times 800$. Evaluation is done on 800 pairs, by also exchanging source and target images. The PCK is computed with respect to source image size.

\subsection{Robustness to specific challenges}
\label{sup:occ}

To better understand the performance of our training approach under complex conditions, we report the results according to different variation factors with various difficulty levels. In particular, the SPair-71k dataset contains diverse variations in view-point, scale, truncation and occlusion. We are particularly interested in the occlusion setting.

\parsection{Comparison to state-of-the-art} In Tab.~\ref{tab:occ-state}, we compare state-of-the-art approaches by variation factor on the SPair-71K dataset. Both our weakly-supervised (W) approaches PWarpC-NC-Net and PWarpC-SF-Net bring a significant improvement compared to their respective baselines, for all variation factor and all difficulty levels. Specifically for occlusion, our PWarpC-SF-Net and PWarpC-NC-Net bring an absolute gain of $8.7\%$ and $5.9 \%$, from their respective baselines SF-Net and NC-Net. 

As for strongly-supervised approaches (S), each of our PWarpC network shows an improvement compared to its baseline, for all four variation factors. 

\parsection{Ablation study} In Tab.~\ref{tab:spair-occ} top part, we show the impact of the key components of our weakly-supervised approach. From version (II), which corresponds to training with our PW-bipath objective without the visibility mask, to our final PWarpC-SF-Net, denoted as (V), incrementally adding each of our losses brings a significant improvement for all variation factors and levels of difficulty. In particular, in (V), our explicit occlusion modeling, \ie through introducing an unmatched state and our probabilistic negative loss,  leads to a particularly impressive gain compared to (IV), of $3.3\%$ and $1.8\%$ for truncation and occlusion respectively. 

\parsection{Comparison to alternative weakly-supervised cost volume losses} In Tab.~\ref{tab:spair-occ} bottom part, we further compare our final weakly-supervised PWarpC-SF-Net to alternative weakly-supervised objectives. We first compare our probabilistic approach (V) to the mapping-based Warp Consistency~\cite{warpc}, denoted as (VI). Here, note that to train with the Warp Consistency objective~\cite{warpc}, the predicted cost volume is converted to a mapping, by applying kernel soft-argmax as in the original work~\cite{SFNet}. Our probabilistic approach (V) obtains significantly better results than the mapping-based warp consistency (VI), for all variations factors and level of difficulties. 
Version (VII) corresponds to training the SF-Net architecture with only the PWarp-supervision objective. The performance is much worse than when trained with our Probabilistic Warp Consistency (V). 
Finally, both training with maximizing the max scores in (VIII), or minimizing the cost volume entropy in (IX) also lead to poor results compared to our approach (V) for all variation factors.

\subsection{Per-category SPair-71K results when trained on PF-Pascal}
\label{sup:per-cat-pfpascal}

\begin{table*}[t!]
\centering
\vspace{-3mm}
\resizebox{0.98\textwidth}{!}{%
\begin{tabular}{l@{~~}ll|ccccccccccccccccccc}
\toprule
& Methods & Reso & aero & bike & bird & boat & bottle & bus & car & cat & chair & cow & dog & horse & moto & person & plant & sheep & train & tv & all \\ \toprule
U & CNNGeo (from~\cite{spair})  & - & 21.3 & 15.1 & 34.5 & 12.8 & 31.2 & 26.3 & 24.0 & 30.6 & 11.6 & 24.3 & 20.4 & 12.2 & 19.7 & 15.6 & 14.3 & 9.6 & 28.5 & 28.8 & 18.1 \\
& A2Net (from~\cite{spair}) & - & 20.8 & 17.1 & 37.4 & 13.9 & 33.6 & 29.4 & 26.5 & 34.9 & 12.0 & 26.5 &  22.5 & 13.3 & 21.3 & 20.0 & 16.9 & 11.5 & 28.9 & 31.6 & 20.1 \\  \midrule

M & SF-Net & ori $^\dagger$ & 24.1 & 15.7 & 43.3 & 14.6 & 35.6 & 20.8 & 13.8 & 47.7 & 14.9 & 26.7 & 24.2 & 13.4 & 19.4 & 20.7 & 15.2 &  14.2 & 28.8 & 34.9 & 24.0 \\ \midrule
W & \textbf{PWarpC-SF-Net} & ori & \best{38.8} & \second{27.6} & \second{58.3} & \second{18.9} & \best{41.3} & \best{30.3} & \second{21.7} & \second{56.2} & \best{20.1} & \best{38.3} & \second{33.8} & \second{20.0} & \second{28.6} & \second{24.2} & \second{21.7} & \second{18.2} & \second{42.2} & \best{60.0} & \second{33.5}\\
 
\cdashlinelr{2-22}

 & WeakAlign (from~\cite{spair}) & - &  23.4 & 17.0 & 41.6 & 14.6 & 37.6 & 28.1 & 26.6 & 32.6 & 12.6 & 27.9 & 23.0 & 13.6 & 21.3 & 22.2 & 17.9 & 10.9 & 31.5 & 34.8 & 21.2 \\
& NC-Net (from~\cite{spair})  & - & 24.0 & 16.0 & 45.0 & 13.7 & 35.7 & 25.9 & 19.0 & 50.4 & 14.3 & 32.6 & 27.4 & 19.2 & 21.7 & 20.3 & 20.4 & 13.6 & 33.6 & 40.4 & 26.4 \\
 
 \cdashlinelr{2-22}
 & NC-Net & ori $^\dagger$ & 25.9 & 18.1 & 45.6 & 16.7 & 39.7 & 26.6 & 20.0 & 52.7 & 15.5 & 33.4 & 30.6 & 18.4 & 24.4 & 23.5 & 24.2 & 16.0 & 36.1 & 47.3 & 28.8 \\

& \textbf{PWarpC-NC-Net} & ori & \second{37.4} & \best{28.8} & \best{60.8} & \best{22.9} & \second{40.5} & 29.4 & \best{22.8} & \best{60.1} & 19.5 & 37.8 & \best{38.4} & \best{27.9} & \best{32.1} & \best{29.7} & \best{29.2} & \best{20.2} & \best{44.5} & 50.0 & \best{35.3} \\

 \midrule   \midrule
S & DHPF & ori $^\dagger$& 23.1 & 21.6 & 56.0 & 16.6 & 36.6 & 21.7 & 15.8 & 49.6 & 16.5 & 31.3 & 34.8 & \second{19.2} & 25.0 & 25.8 & 20.3 & 14.8 & 31.7 & 31.5 & 27.5 \\
 & \textbf{PWarpC-DHPF} & ori & 25.2 & 23.8 & 62.0 & 17.1 & 35.1 & 23.5 & 16.8 & 51.2 & 16.9 & 34.7 & \second{34.6} & 19.1 & 25.6 & 25.3 & 18.1 & 16.0 & 31.6 & 37.0 & 28.6 \\
  \cdashlinelr{2-22}
& CATs-ft-features & ori $^\dagger$& 23.7 & 18.7 & 49.5 & 16.3 & 37.3 & 20.8 & 14.6 & 47.1 & 17.7 & 32.5 & 30.3 & 15.2 & 22.4 & 22.6 & 20.2 & 15.4 & 34.7 & 37.7 & 26.8 \\
& \textbf{PWarpC-CATs-ft-features} & ori & 24.5 & 21.3 & 56.3 & 16.6 & 35.0 & 23.7 & 16.0 & 54.0 & 15.3 & 34.5 & 36.2 & 14.6 & 21.1 & 19.8 & 17.3 & 15.4 & 39.4 & 37.6 & 27.9 \\
 \cdashlinelr{2-22}
& CATs & ori $^\dagger$& 22.5 & 15.0 & 41.9 & 14.0 & 34.2 & 19.5 & 14.1 & 40.7 & 13.8 & 24.9 & 24.2 & 13.6 & 17.2 & 16.8 & 13.6 & 13.1 & 27.9 & 27.1 & 22.1 \\
& \textbf{PWarpC-CATs} & ori & 24.2 & 14.9 & 44.5 & 14.4 & 34.4 & 21.0 & 15.2 & 44.4 & 13.6 & 27.6 & 26.1 & 14.0 & 17.4 & 16.2 & 15.5 & 12.9 & 31.1 & 28.2 & 23.3 \\
 \cdashlinelr{2-22}
 & SF-Net* & ori & 26.1 & 21.6 & 48.7 & 16.7 & \second{39.6} & 23.1 & 17.8 & 52.6 & 17.7 & 32.2 & 31.9 & 15.7 & 21.8 & \second{27.1} & 22.5 & 16.3 & 31.9 & 35.5 & 27.9 \\
 & \textbf{PWarpC-SF-Net*} &ori &  \second{33.8} & \second{28.3} & \second{56.1} & \second{18.6} & 38.9 & \best{30.4} & 20.5 & 56.3 & \second{19.3} & \second{36.8} & 32.4 & 18.4 & \second{28.9} & 26.1 & \second{23.4} & \second{18.6} & \second{42.2} & \second{53.1} & \second{32.5} \\
 \cdashlinelr{2-22}
 
& NC-Net* & ori & 28.8 & 24.0 & 53.6 & 19.2 & 41.1 & 27.8 & \second{21.4} & \second{61.1} & 18.8 & 38.5 & 35.4 & 22.9 & 25.2 & 25.9 & 28.1 & 20.7 & 41.3 & 45.3 & 32.4 \\ 
& \textbf{PWarpC-NC-Net*} & ori & \best{40.1} & \best{31.0} & \best{65.5} & \best{23.4} & \best{43.1} & \second{29.4} & \best{21.9} & \best{61.8} & \best{21.4} & \best{41.2} & \best{39.2} & \best{28.1} & \best{32.0} & \best{30.8} & \best{30.0} & \best{22.5} & \best{43.9} & \best{58.2} & \best{37.1} \\

\bottomrule
\end{tabular}%
}\vspace{-2mm}
\caption{Per-class PCK ($\alpha_{bbox} = 0.1$) results on SPair-71K. All models in this table use ResNet101 as the backbone, and are trained on the training set of PF-Pascal. \textbf{S} denotes strong supervision using keypoint match annotations, \textbf{M} refers to using ground-truth object segmentation mask, \textbf{U} is fully unsupervised requiring only single images, and \textbf{W} refers to weakly-supervised with image-level class labels. Each method evaluates with ground-truth annotations resized to a specific resolution. However, using different ground-truth resolutions leads to slightly different results. We therefore use the standard setting of evaluating on the original resolution (\textbf{ori}). When needed, we re-compute metrics of baselines using the provided pre-trained weights, indicated by $^\dagger$. For each of our PWarpC networks, we compare to its corresponding baseline within the dashed-lines. Best and second best results are in red and blue respectively. }
\label{tab:per-cat-spair-pfpascal}
\end{table*}

For completeness, we provide the results per category on the SPair-71K dataset in Tab.~\ref{tab:per-cat-spair-pfpascal}. All approaches are trained on the PF-Pascal dataset. It corresponds to results provided in Tab.~\ref{tab:sem} of the main paper.

\begin{figure}[t]
\centering%
\vspace{-4mm}
\begin{tabular}{c}
\small{
\hspace{0.3cm}Target \hspace{1.1cm} Source  \hspace{0.7cm}   $P_{S \leftarrow T}(\cdot | t)$  \hspace{0.5cm}   $P_{S \leftarrow T}(\o | \cdot)$ }
\end{tabular}
\vspace{-2mm}
\includegraphics[width=0.48\textwidth]{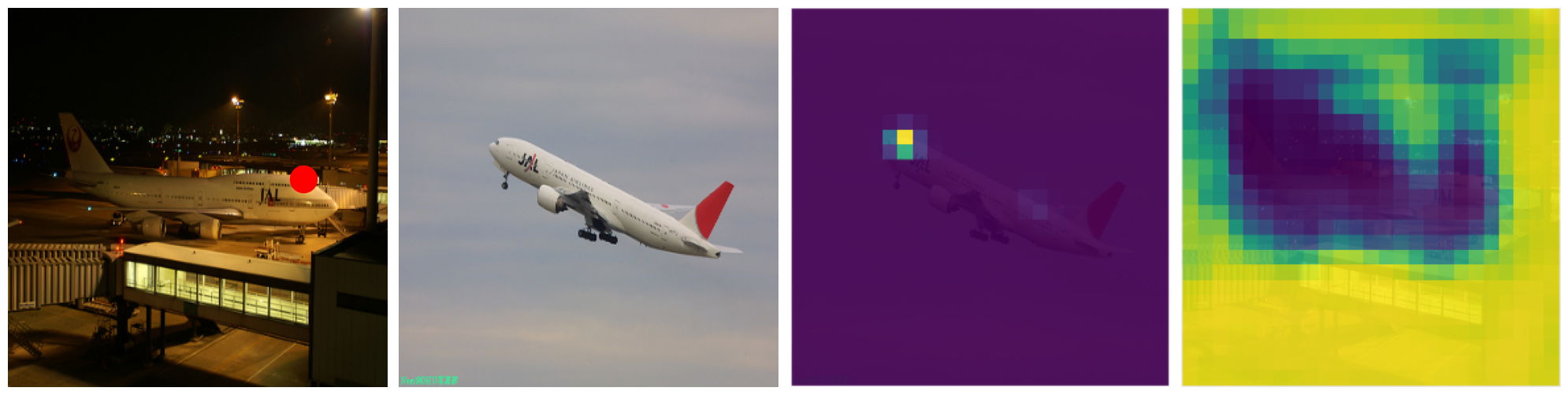} \\
\includegraphics[width=0.48\textwidth]{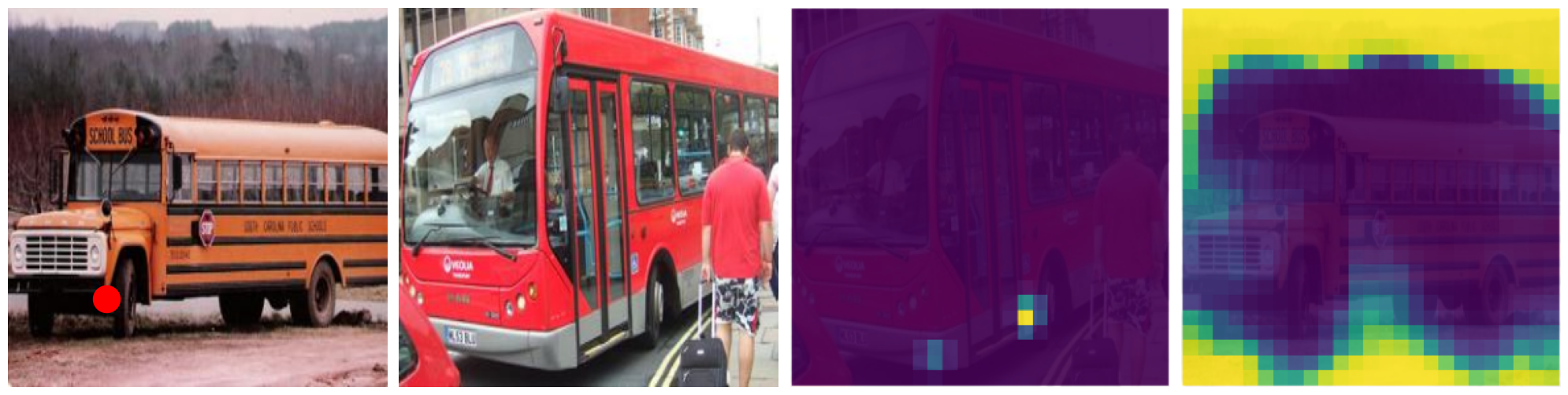} \\
\includegraphics[width=0.48\textwidth]{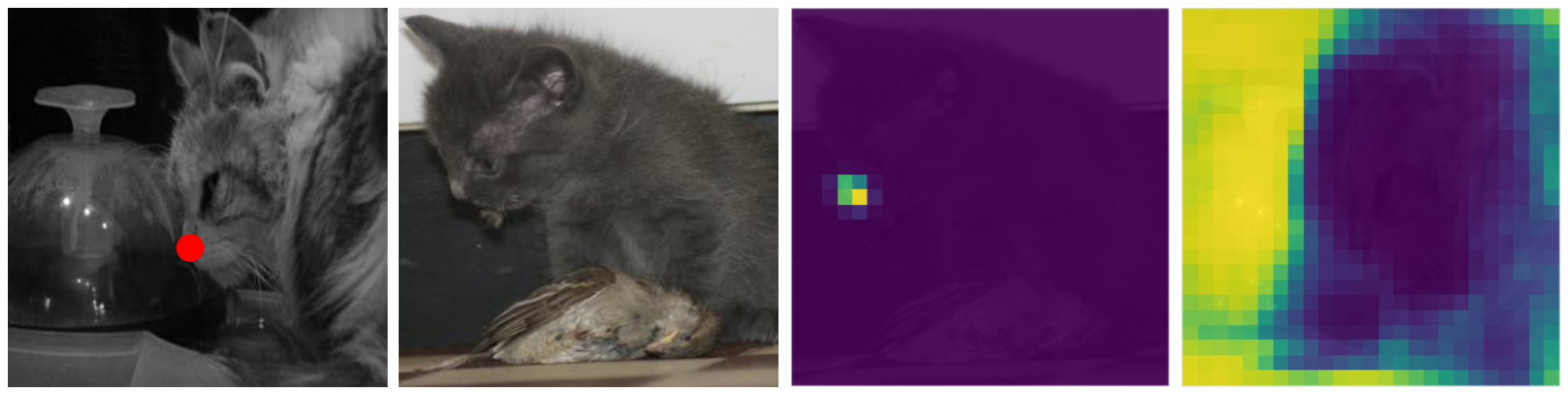} \\
\includegraphics[width=0.48\textwidth]{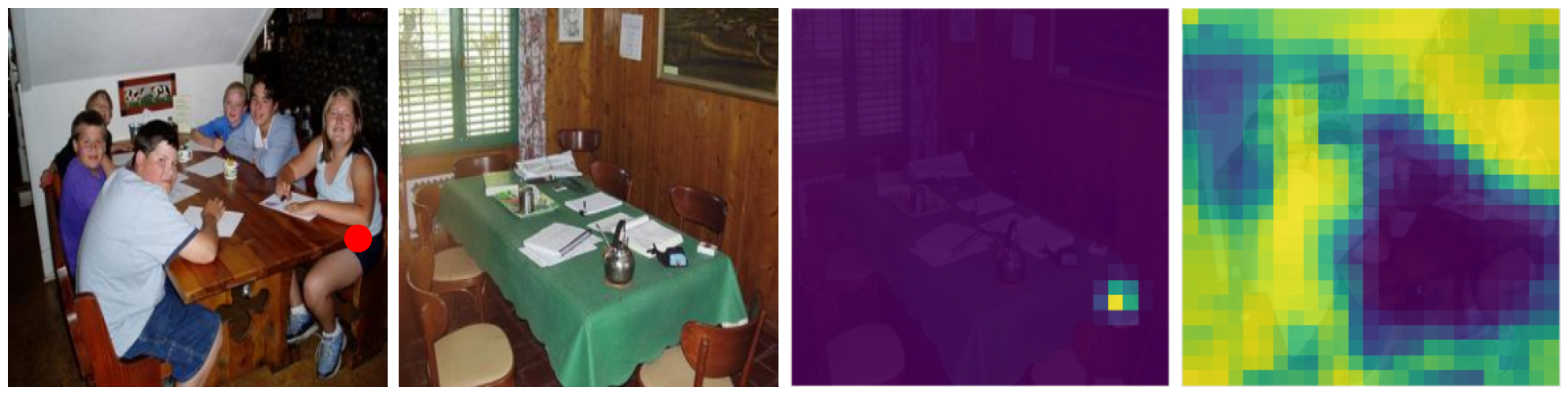} \\
\includegraphics[width=0.48\textwidth]{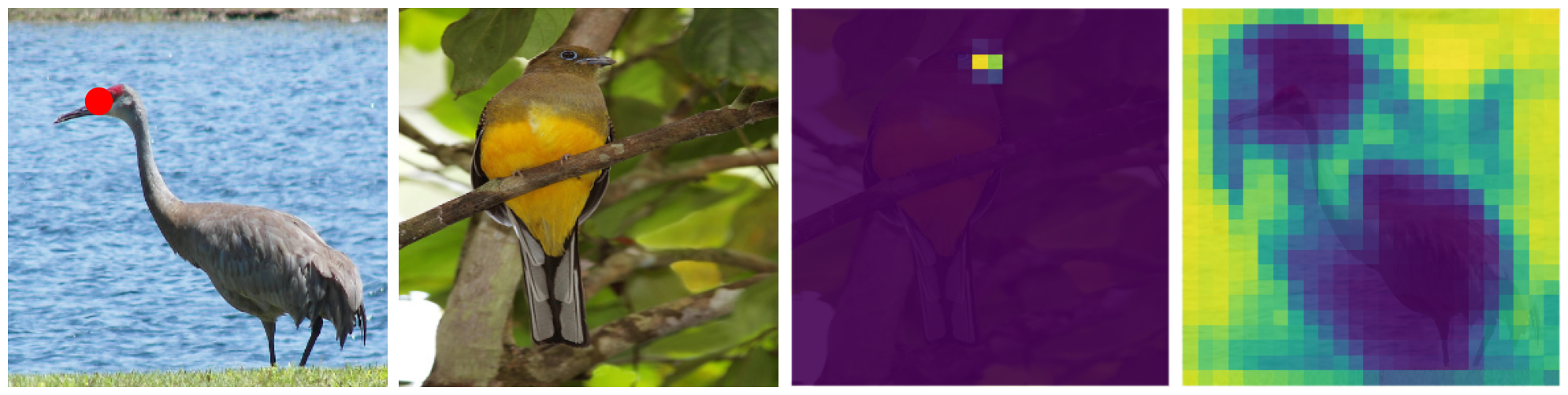} \\
\includegraphics[width=0.48\textwidth]{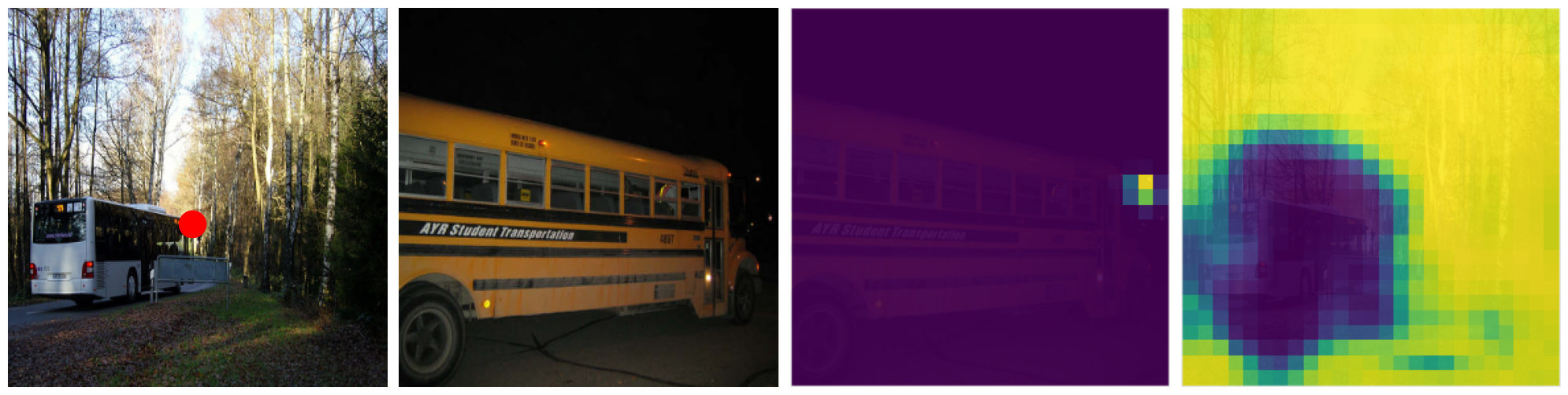} \\
\includegraphics[width=0.48\textwidth]{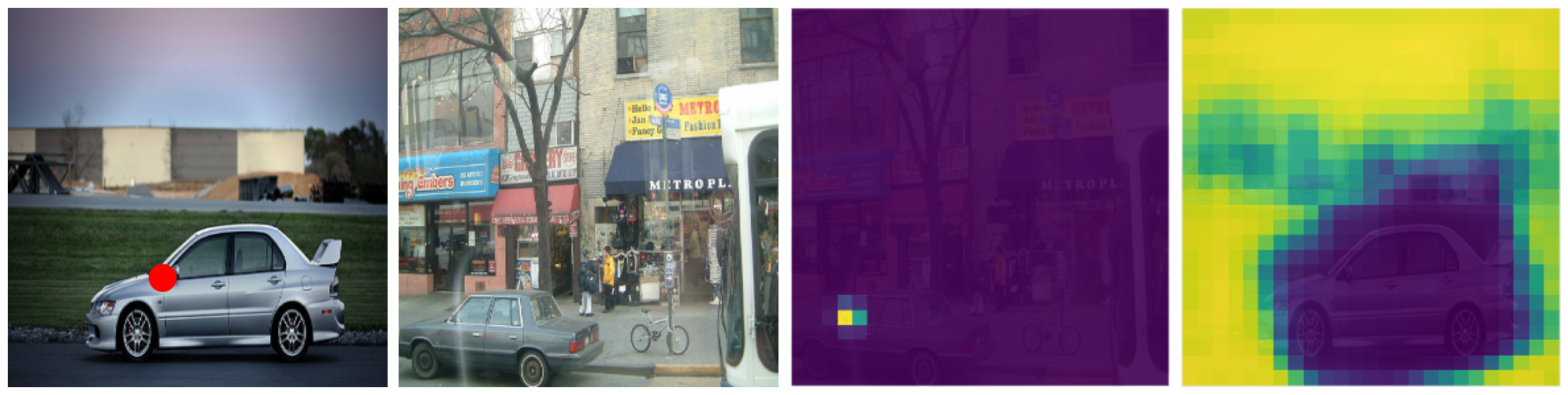} \\
\includegraphics[width=0.48\textwidth]{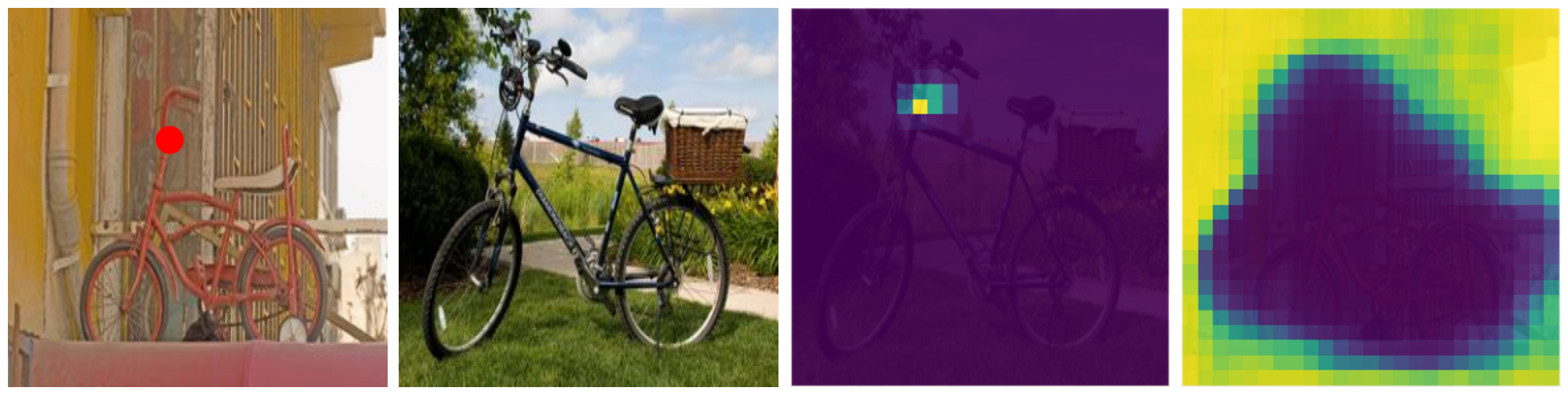} \\
\includegraphics[width=0.48\textwidth]{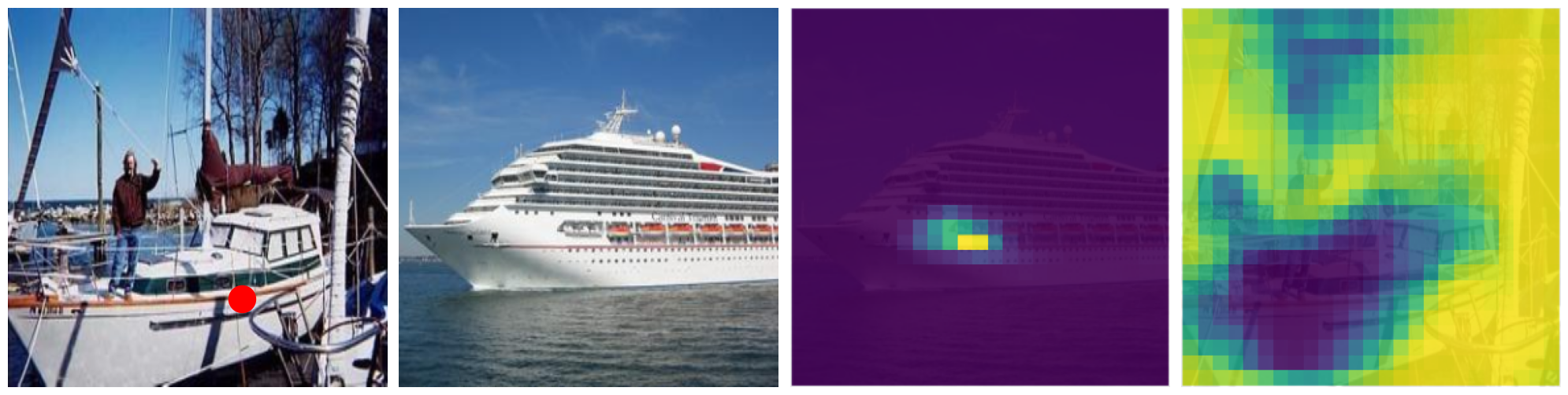} \\
\caption{Examples of the predicted probability distribution relating the target to the source image, given a pixel location $t$ (red dot) in the target image. We also show the predicted unmatched state for all pixels of the target image. Yellow and purple corresponds respectively to values of $1$ and $0$.  Here, we use our weakly-supervised PWarpC-NC-Net for the predictions. 
}\vspace{-4mm}
\label{fig:occ-pred}
\end{figure}

\subsection{Example predictions for the unmatched state}
\label{sup:occ-pre}

In Fig.~\ref{fig:occ-pred}, we provide examples of the unmatched state predictions of our weakly-supervised approach PWarpC-NC-Net. The results are similar for PWarpC-SF-Net. Our probabilistic approach predicts Dirac-like distributions, whose mode are correct.
Furthermore, through our explicit occlusion modeling approach (Sec.~\ref{sec:occ} of the paper), the network successfully identifies the object in most examples.

\begin{figure}[t]
\centering%
\includegraphics[width=0.47\textwidth, trim=0 0 0 0]{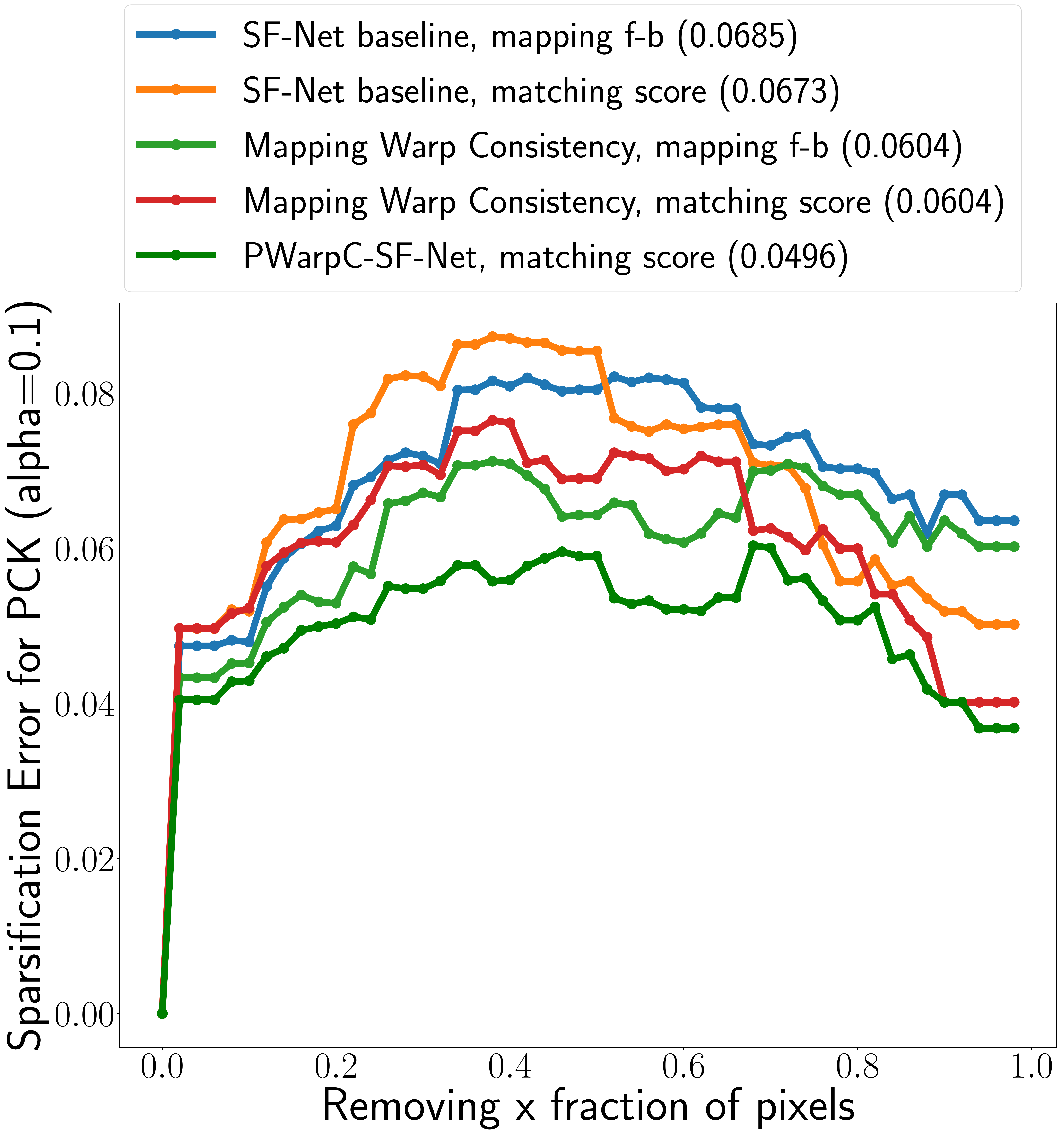}
\vspace{-3mm}
\caption{Sparsification error curves and AUSE on the PF-Pascal dataset, for our PWarpC-SF-Net, baseline SF-Net and SF-Net trained with the mapping-based Warp Consistency objective. For the two latter, the predicted cost volume is converted to a mapping before applying the loss. For this reason, we show two alternative confidence estimation schemes, based on the matching scores, or on the forward-backward consistency of the flow. For our approach PWarpC-SF-Net, our Probabilistic Warp Consistency objective is applied directly on the probabilistic mapping, therefore we directly use the probabilities of the hard assigned matches as confidence measures. Smaller AUSE is better. 
}
\vspace{-4mm}
\label{fig:sparsi-base}
\end{figure}

\begin{figure}[t]
\centering%
\includegraphics[width=0.47\textwidth, trim=0 0 0 0]{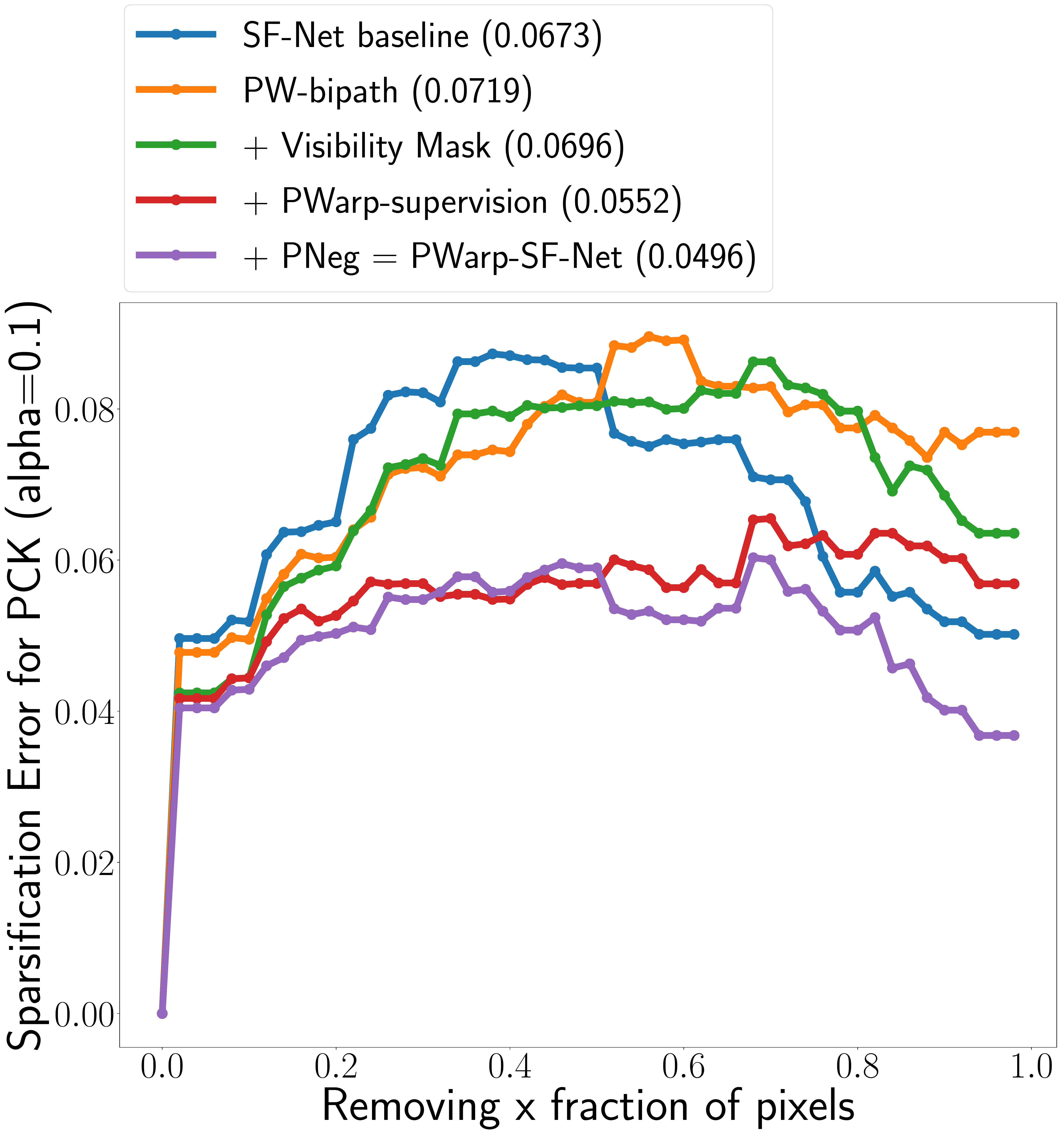}
\vspace{-3mm}
\caption{Ablation study for weakly-supervised PWarpC-SF-Net in terms of sparsification error curves and AUSE on the PF-Pascal dataset. As confidence measure, we use the probabilities of the hard assigned matches.  Smaller AUSE is better. 
}
\vspace{-4mm}
\label{fig:sparsi-ablation}
\end{figure}

\newpage
\subsection{Confidence analysis }
\label{sup:conf}

Most semantic matching architectures predict a cost volume as the final network output. The cost volume, after conversion to a probabilistic mapping through SoftMax, inherently encodes the confidence of each predicted tentative match. It is not the case when directly regression a mapping or flow output instead. Nevertheless, a confidence estimation for each of the predicted matches can in the case be obtained, by \eg forward-backward consistency of the flow field~\cite{Meister2017}. 
In this section, we analyse the quality of the confidence predictions,  when the networks are trained with our weakly-supervised Probabilistic Warp Consistency approach. 

A common technique to assess the quality of a confidence estimate is to rely on sparsification and error curves.

\parsection{Sparsification and error curves} 
To assess the quality of the uncertainty estimates, we rely on sparsification plots, in line with~\cite{Aodha2013LearningAC,Ilg2017a, ProbFlow, pdcnet, pdcnet+}. 
The pixels having the highest uncertainty are progressively removed and the PCK of the remaining pixels is plotted in the sparsification curve. These plots reveal how well the estimated uncertainty relates to the true errors. Ideally, larger uncertainty should correspond to larger errors. Gradually removing the predictions with the highest uncertainties should therefore monotonically improve the accuracy of the remaining correspondences. The sparsification plot is compared with the best possible ranking of the predictions, according to their actual errors computed with respect to the ground-truth flow. We refer to this curve as the oracle plot.

Note that, for each network the oracle is different. Hence, an evaluation using a single sparsification plot is not possible. To this end, we use the Sparsification Error, constructed by directly comparing each sparsification plot to its corresponding oracle plot by taking their difference. Since this measure is independent of the oracle, a fair comparison between different methods is possible. As evaluation metric, we use the Area Under the Sparsification Error curve (AUSE). We compute the sparsification error curve on each image pair. The final error curve is the average over all image pairs of the dataset. 

The sparsification and error plots provide an insightful and directly relevant assessment of the uncertainty. In particular, the AUSE directly evaluates the ability to filter out inaccurate and incorrect correspondences, which is the main purpose of the uncertainty estimate.

\begin{table}[t]
\centering
\vspace{-3mm}
\resizebox{0.47\textwidth}{!}{%
\begin{tabular}{l@{~~}l@{~~}l|cc}
\toprule
Sup. & Methods & Reso & LT-ACC $\uparrow$  & IoU $\uparrow$ \\

\toprule
S & SCNet$_{\text{VGG16}}$~\cite{SCNet}  (from~\cite{HPF}) & -  & 0.79 & 0.51 \\
& HPF$_{\text{res101}}$~\cite{HPF}  & ori & \second{0.87} & \second{0.63} \\
& DHPF$_{\text{res101}}$~\cite{MinLPC20} & 240 &  \lowreso{0.87} & \lowreso{0.62} \\
\cdashlinelr{2-5}
& CATs$_{\text{res101}}$~\cite{CATs} & ori $^\dagger$ & 0.84 & 0.58 \\
& \textbf{PWarpC-CATs}$_{\text{res101}}$ & ori  & 0.85 & 0.60 \\
\cdashlinelr{2-5}
& CATs-ft-features$_{\text{res101}}$~\cite{CATs} & ori $^\dagger$ & 0.84 & 0.59 \\
& \textbf{PWarpC-CATs-ft-features}$_{\text{res101}}$ & ori  & 0.86 & 0.60 \\
\cdashlinelr{2-5}
& DHPF$_{\text{res101}}$~\cite{MinLPC20} & ori $^\dagger$ &  \second{0.87} & 0.62  \\
& \textbf{PWarpC-DHPF}$_{\text{res101}}$ &  ori &  \best{0.88} & \best{0.64} \\
\cdashlinelr{2-5}
& NC-Net*$_{\text{res101}}$  & ori  &  0.81 & 0.54 \\
& \textbf{PWarpC-NC-Net*}$_{\text{res101}}$ & ori  & \second{0.87} & 0.62  \\
\cdashlinelr{2-5}
&  SF-Net*$_{\text{res101}}$ & ori &  0.85 & 0.59  \\
& \textbf{PWarpC-SF-Net*}$_{\text{res101}}$ & ori & \second{0.87} & 0.62  \\
\midrule  \midrule

U & CNNGeo$_{\text{res101}}$~\cite{Rocco2017a}  (from~\cite{Rocco2018a}) & -  &  0.83 & 0.61 \\
& A2Net$_{\text{res101}}$~\cite{SeoLJHC18} & - &  0.80 & 0.57 \\

\midrule
M & SF-Net$_{\text{res101}}$~\cite{SFNet} & ori $\dagger$ & \best{0.87} & \best{0.64} \\

\midrule
W & \textbf{PWarpC-SF-Net}$_{\text{res101}}$ & ori & \second{0.86} & 0.61  \\
\cdashlinelr{2-5}
 & WeakAlign$_{\text{res101}}$~\cite{Rocco2018a}  &   &  0.85 & \second{0.63} \\
& DHPF$_{\text{res101}}$~\cite{MinLPC20} &  240 &  0.86 & 0.61 \\
& DHPF$_{\text{res101}}$~\cite{MinLPC20} &  ori $^\dagger$ &  \best{0.87} & \second{0.63} \\
& NC-Net$_{\text{res101}}$~\cite{Rocco2018b} & -  &  0.85 & 0.60 \\
\cdashlinelr{2-5}
& NC-Net$_{\text{res101}}$~\cite{Rocco2018b} & ori $^\dagger$  &  0.85 & 0.58 \\
& \textbf{PWarpC-NC-Net}$_{\text{res101}}$  &  ori & \second{0.86} & 0.61 \\
\bottomrule
\end{tabular}%
}\vspace{-2mm}
\caption{State-of-the-art comparison on the Caltech-101 dataset. All approaches are trained on the PF-Pascal dataset.  \textbf{S} denotes strong supervision using keypoint annotation, \textbf{M} refers to using ground-truth object segmentation mask, \textbf{U} is fully unsupervised requiring only single images, and \textbf{W} refers to weakly-supervised with image-level class labels. Each method evaluates with ground-truth annotations resized to a specific resolution. However, using different ground-truth resolutions leads to slightly different results. We therefore use the standard setting of evaluating on the original resolution (\textbf{ori}) and gray the results computed at a different size. When needed, we re-compute metrics of baselines using the provided pre-trained weights, indicated by $^\dagger$. For each of our PWarpC network, we compare to its corresponding baseline within the dashed-lines. Best and second best results are in red and blue respectively. }
\label{tab:caltech}
\end{table}

\parsection{Results}  For this confidence analysis, we use SF-Net as baseline. In Fig.~\ref{fig:sparsi-base}, we report the sparsification error curves and AUSE obtained by our approach PWarpC-SF-Net, the baseline SF-Net, and SF-Net trained with the mapping-based Warp Consistency~\cite{warpc} objective. For the baseline and the Warp Consistency versions, using directly the predicted matching scores of the hard-assigned matches or the forward-backward mapping as confidence measure leads to similar results. Nevertheless, our probabilistic PWarpC-SF-Net outperforms the other approaches in AUSE. It demonstrates the benefit of our probabilistic approach, acting directly on dense matching scores.  In particular, it shows our approach can filter out inaccurate and incorrect correspondences better than previous methods, which is extremely important for usability in end tasks.

\parsection{Ablation study} In Fig.~\ref{fig:sparsi-ablation}, we show the impact of the key components of our weakly-supervised Probabilistic Warp Consistency approach, on the confidence estimation robustness. Adding the visibility mask, our PWarp-supervision loss and our occlusion modelling consistency improves the AUSE.

\subsection{Additional results on Caltech} 
\label{sup:caltech}

Here, we additionally evaluate our approach on the Caltech dataset. 

\noindent\textbf{Caltech-101} contains images depicting 101 diverse object classes. Each image comes with a ground-truth foreground object segmentation mask. Although originally introduced for the image classification task, this dataset was adopted for assessing semantic alignment. Particularly, the predicted dense correspondences relating the target  to the source image are used to warp the ground-truth segmentation mask of the source towards the target. The overlap between the warped source segmentation mask and the ground-truth target segmentation mask is then measured, and used as a proxy to assess the quality of the predicted dense correspondences.  
We follow the standard set-up, according to which the evaluation is performed on 1515 semantically related image pairs, \ie 15 pairs for each of the 101 object categories of the dataset. The semantic alignment is evaluated using two different metrics: the label transfer accuracy (LT-ACC) and the intersection-over-union (IoU). They both measure the overlap between the annotated foreground object segmentation masks, with former putting more emphasis on the background class and the latter on the foreground object. 

Note that compared to other benchmarks described above, the Caltech-101 dataset provides image pairs from more diverse classes, enabling us to evaluate our method under more general correspondence settings. 

\parsection{Results} In Tab.~\ref{tab:caltech}, we present results of semantic networks on the Caltech dataset. All approaches are trained on the PF-Pascal dataset.
For both weakly-supervised (W) and strongly-supervised (S) approaches, our PWarpC networks show significantly better performance than their respective baselines. The only exception is our weakly-supervised PWarpC-SF-Net, which obtains worse results than its baseline SF-Net. This is because on Caltech-101, the evaluation is conducted by warping the foreground mask of the source image, according to the predicted dense flow or mapping. In that case, a smooth flow is very beneficial. Baseline SF-Net explicitly enforces smoothness of the predicted flow fields as part of the training strategy~\cite{SFNet}, which explains its good performance. On the other hand, we do not specifically enforce any smoothness priors, which is why PWarpC-SF-Net lacks in performance compared to SF-Net for this particular dataset and evaluation. Nevertheless, on all other datasets (see Tab.~\ref{tab:sem}), where the metrics are based directly on the predicted matches, our approach PWarpC-SF-Net significantly outperforms baseline SF-Net. 
Moreover, our PWarpC-DHPF sets a new state-of-the-art on Caltech-101 across all approaches, independently of their level of supervision.

\subsection{Qualitative results}
\label{sup:qual}

\begin{figure*}[t]
\centering%
\vspace{-20mm}
\includegraphics[width=0.75\textwidth]{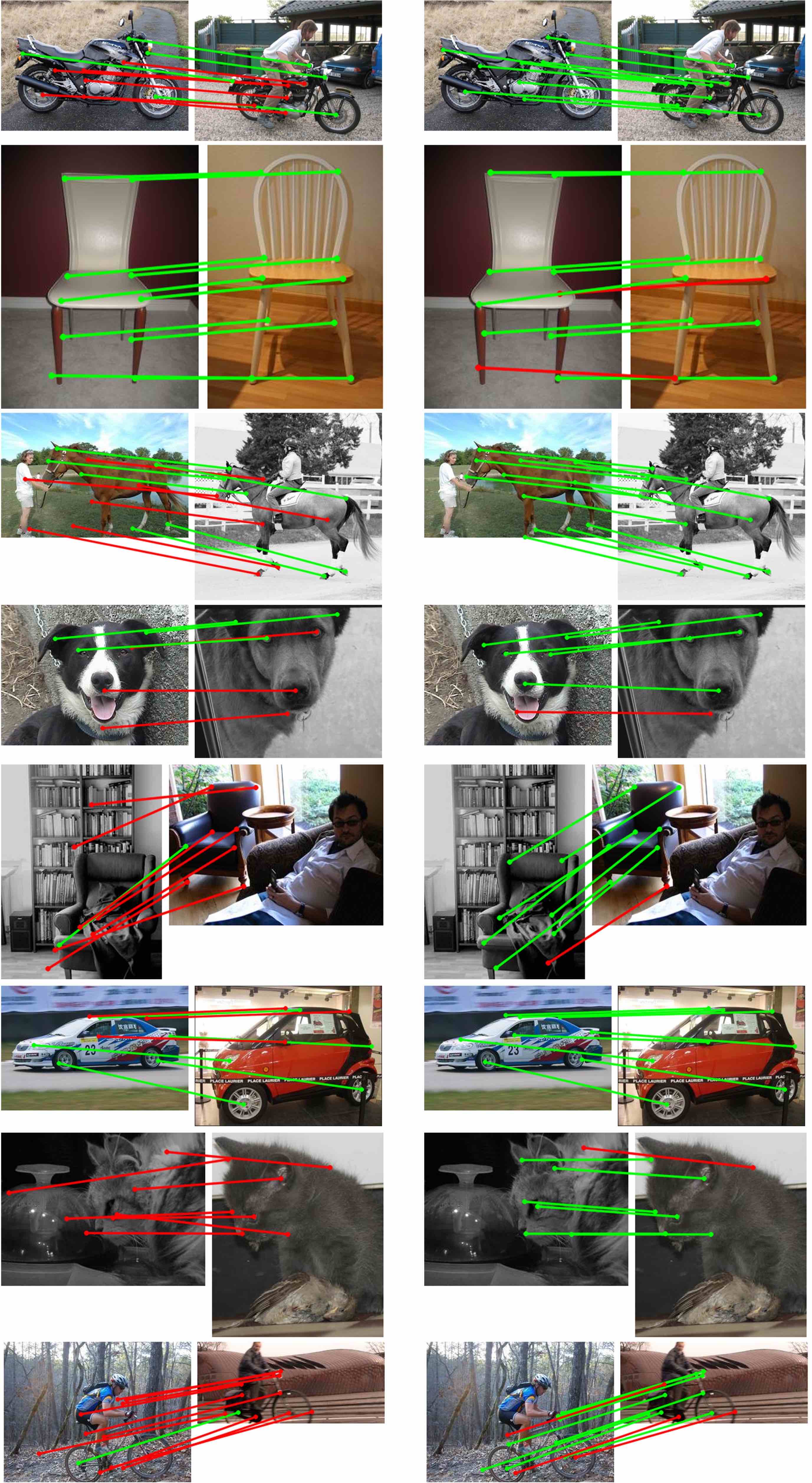}
\caption{Example predictions on PF-Pascal and PF-Willow, of baseline SF-Net~\cite{SFNet} (\textbf{left}) compared to our weakly-supervised PWarpC-SF-Net (\textbf{right}). Green and red line denotes correct and wrong predictions, respectively, with respect to the ground-truth. 
}\vspace{-4mm}
\label{fig:sfnet-pfpascal}
\end{figure*}

\begin{figure*}[t]
\centering%
\vspace{-20mm}
\includegraphics[width=0.60\textwidth]{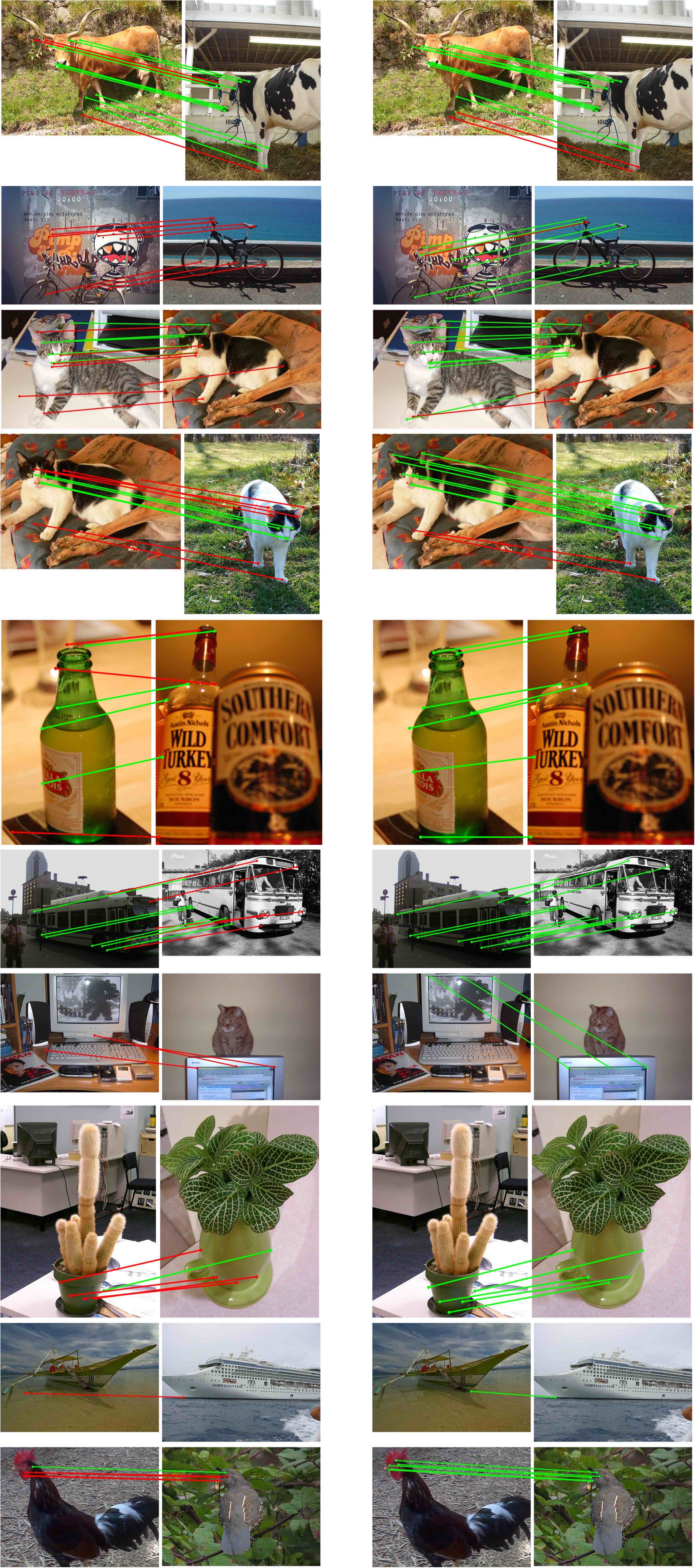}
\vspace{-3mm}
\caption{Example predictions on SPair-71K, of baseline SF-Net~\cite{SFNet} (\textbf{left}) compared to our weakly-supervised PWarpC-SF-Net (\textbf{right}). Green and red line denotes correct and wrong predictions, respectively, with respect to the ground-truth. 
}\vspace{-4mm}
\label{fig:sfnet-spair}
\end{figure*}

\begin{figure*}[t]
\centering%
\includegraphics[width=0.90\textwidth]{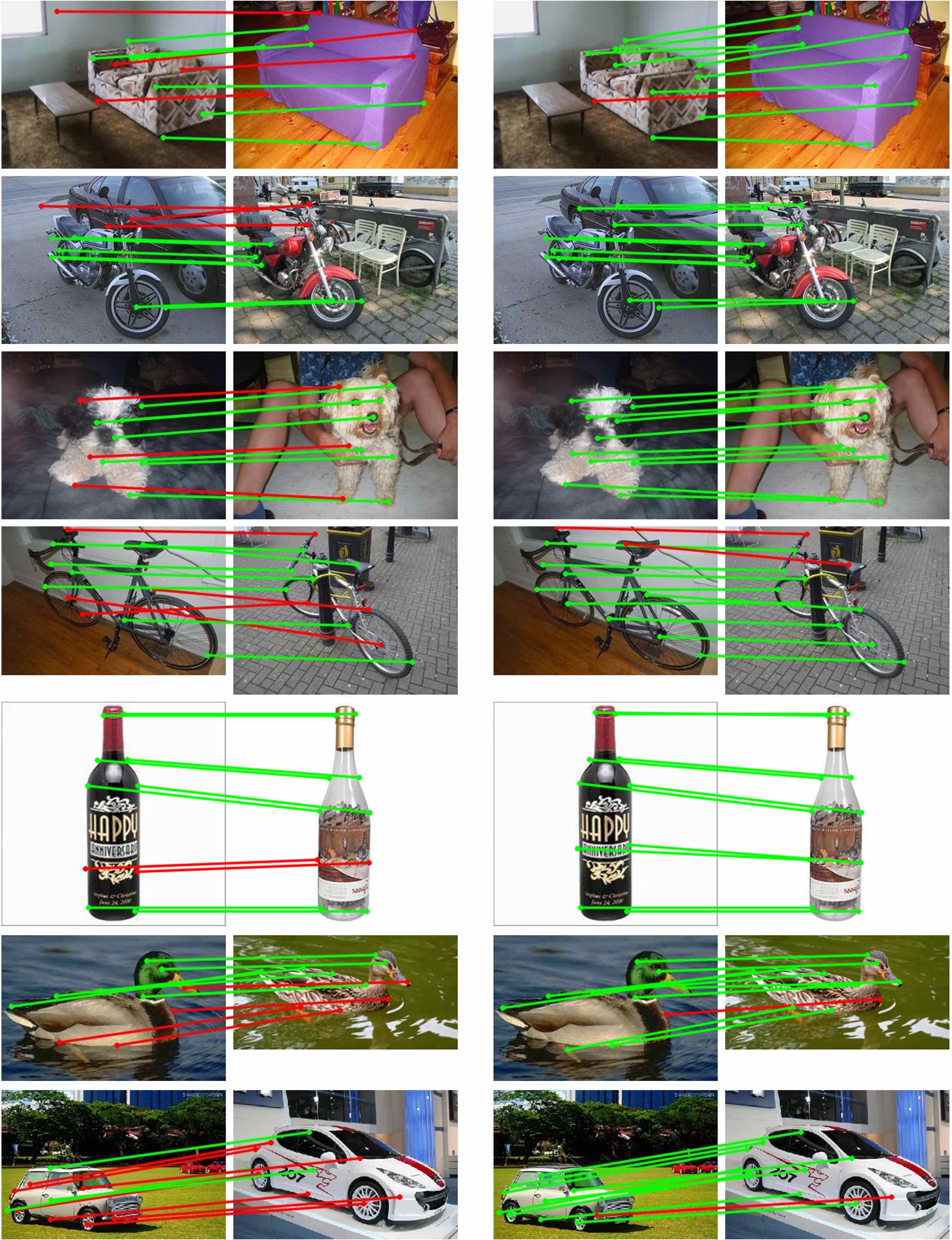}
\vspace{-3mm}
\caption{Example predictions on PF-Pascal and PF-Willow, of baseline NC-Net~\cite{Rocco2018b} (\textbf{left}) compared to our weakly-supervised PWarpC-NC-Net (\textbf{right}). Green and red line denotes correct and wrong predictions, respectively, with respect to the ground-truth. 
}\vspace{-4mm}
\label{fig:ncnet-pf}
\end{figure*}

\begin{figure*}[t]
\centering%
\vspace{-20mm}
\includegraphics[width=0.70\textwidth]{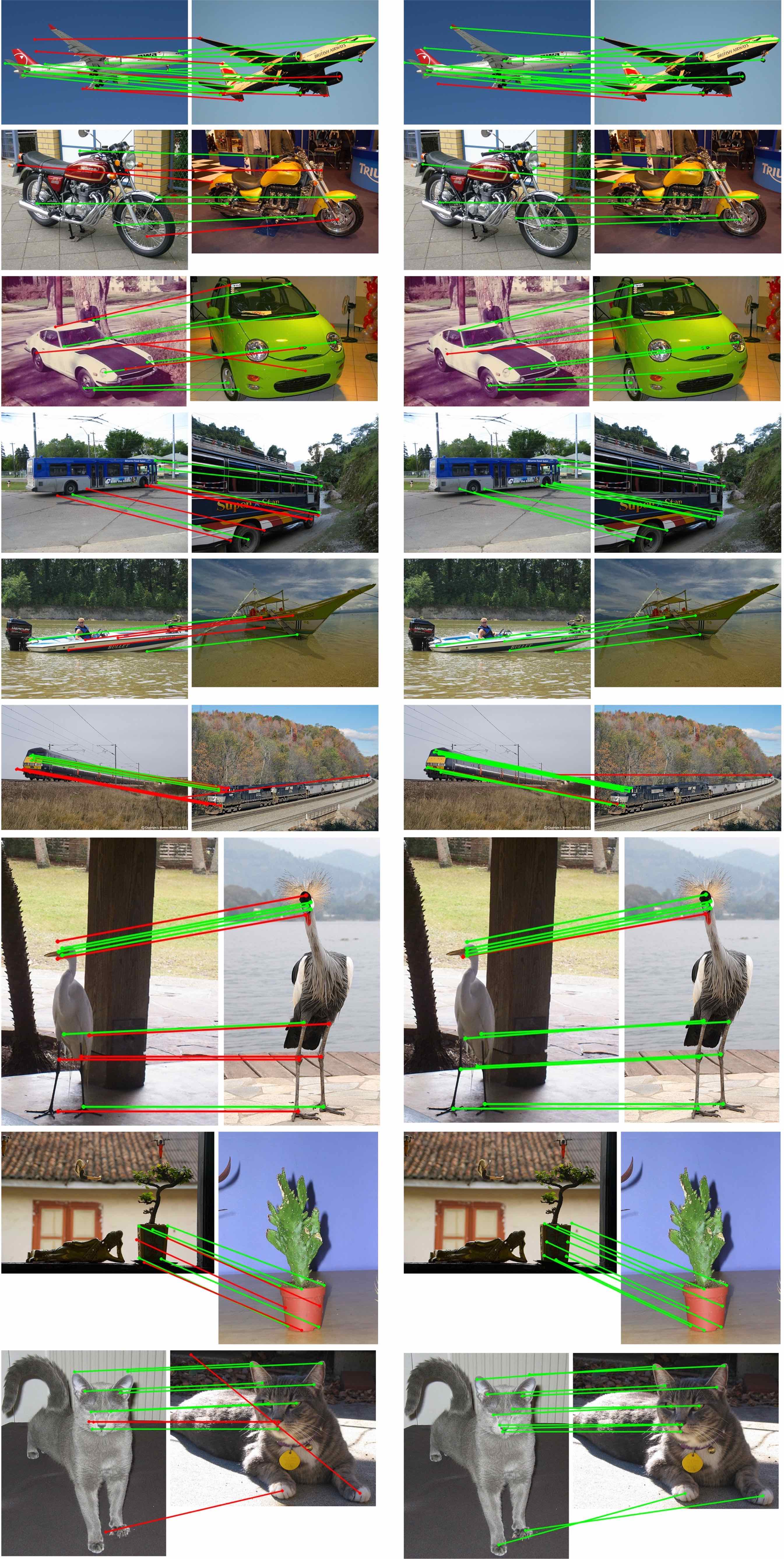}
\vspace{-3mm}
\caption{Example predictions on SPair-71K, of baseline NC-Net~\cite{Rocco2018b} (\textbf{left}) compared to our weakly-supervised PWarpC-NC-Net (\textbf{right}). Green and red line denotes correct and wrong predictions, respectively, with respect to the ground-truth. 
}\vspace{-4mm}
\label{fig:ncnet-spair}
\end{figure*}

In Fig.~\ref{fig:sfnet-pfpascal} and~\ref{fig:sfnet-spair}, we show example predictions of baseline SF-Net compared to our weakly-supervised approach PWarpC-SF-Net on the PF-Pascal, PF-Willow and SPair-71K datasets. Similarly, we present example predictions for NC-Net and PWarpC-NC-Net in Fig.~\ref{fig:ncnet-pf}-\ref{fig:ncnet-spair}. Our weakly-supervised Probabilistic Warp Consistency approach finds significantly more correct matches than the baseline in both cases, for a variety of object classes. 

